\pdfoutput=1
\documentclass[10pt,twocolumn,letterpaper]{article}
\usepackage[margin=0.75in]{geometry}
\usepackage{times}
\usepackage[dvipsnames]{xcolor}
\usepackage[numbers,sort&compress]{natbib}
\usepackage[hyphens]{url}
\expandafter\def\expandafter\UrlBreaks\expandafter{\UrlBreaks\do\a\do\b\do\c\do\d\do\e\do\f\do\g\do\h\do\i\do\j\do\k\do\l\do\m\do\n\do\o\do\p\do\q\do\r\do\s\do\t\do\u\do\v\do\w\do\x\do\y\do\z\do\A\do\B\do\C\do\D\do\E\do\F\do\G\do\H\do\I\do\J\do\K\do\L\do\M\do\N\do\O\do\P\do\Q\do\R\do\S\do\T\do\U\do\V\do\W\do\X\do\Y\do\Z\do\0\do\1\do\2\do\3\do\4\do\5\do\6\do\7\do\8\do\9}
\usepackage[font=small,labelfont=bf]{caption}
\usepackage{subcaption}
\usepackage[shortlabels,inline]{enumitem}
\usepackage{booktabs,amsmath,amssymb,graphicx,microtype,multirow,colortbl}
\usepackage{tikz}
\usepackage{balance}
\usepackage{placeins,array}
\usetikzlibrary{shapes.geometric}

\usetikzlibrary{positioning,arrows.meta,fit,calc,plotmarks}
\usepackage[pagebackref,breaklinks,colorlinks,allcolors=blue]{hyperref}
\newcommand{\ours}{\textup{HarnessMM}}
\newcommand{\oursfull}{HarnessMinusMinus (\ours{})}

\newcommand{\E}{\mathbb{E}}
\newcommand{\D}{\mathcal{D}}
\title{
  HarnessMinusMinus: A Technical Report\\
  on Efficient Agent Harnesses\\[0.3em]
  {\large\mdseries\itshape Minus Overhead, Minus Cost}
}
\author{%
Yuhua Jiang$^{1}$\quad Chang Xu$^{2,\dagger}$\quad Feifei Gao$^{1}$\quad Li Zhao$^{2,\dagger}$\\[3pt]
{\small $^{1}$Tsinghua University\quad $^{2}$Microsoft Research Asia (MSRA)}\\[2pt]
{\footnotesize \href{mailto:yh-jiang24@mails.tsinghua.edu.cn}{\textcolor{black}{\texttt{yh-jiang24@mails.tsinghua.edu.cn}}}\quad
\href{mailto:chanx@microsoft.com}{\textcolor{black}{\texttt{chanx@microsoft.com}}}}\\
{\footnotesize \href{mailto:feifeigao@tsinghua.edu.cn}{\textcolor{black}{\texttt{feifeigao@tsinghua.edu.cn}}}\quad
\href{mailto:lizo@microsoft.com}{\textcolor{black}{\texttt{lizo@microsoft.com}}}}%
}
\date{}

\begin{document}
\maketitle
\begingroup
\renewcommand{\thefootnote}{\fnsymbol{footnote}}
\footnotetext[2]{Co-corresponding authors.}
\endgroup

\begin{abstract}
Agent harnesses often invoke reasoning after each tool result,
even when earlier calls have already determined subsequent actions.
We introduce HarnessMinusMinus (HarnessMM), a framework centered
on Executable Decision Graphs (EDGs). An EDG binds resolved native
actions and literal arguments to causal dependencies and public
execution guards. This representation preserves decisions across
tool transitions and lets later feedback determine their readiness
without regenerating their content. Within an EDG, Guarded Decision
Reuse (GDR) preserves the graph and activates eligible nodes; Certified Callback
Substitution (CCS) executes them through the native tool interface
and records omission of the corresponding reasoning callbacks.
Recursive self-improvement (RSI) uses training feedback to refine
the rules that construct and execute these graphs, with validation
selecting the deployment configuration. Actor weights stay fixed,
and new semantic decisions return control to the original reasoning
model. This technical report consolidates the graph framework,
implementations, and studies across coding, tool-based, and
visual-agent settings. On HumanEval, HarnessMM preserves Codex
success at 90.82\% while reducing Tokens/task (k) from 66.72 to
47.02 (29.54\% saving). For DeepSeek Harness, success improves from
71.43\% to 85.71\% while Tokens/task (k) decrease from 96.28 to
11.62 (87.93\% saving). An MBPP ablation shows 26.91\% lower cost
than single-use graphs and 19.52\% lower cost than retained
callbacks. The report also summarizes the HarnessMM-V visual-agent
study and documents optimization overhead and execution time where
measured.
Code is available at \url{https://github.com/microsoft/HarnessMinusMinus}.
\end{abstract}

\section{Introduction}
\label{sec:intro}
Agent harnesses connect model reasoning to tool execution. ReAct alternates reasoning, actions and observations~\cite{yao2023react}; SWE-agent demonstrates that interface design shapes software-engineering performance~\cite{yang2024sweagent}, and Toolformer learns when to call APIs~\cite{schick2023toolformer}. After each tool result, the harness must decide whether the observation demands new reasoning or merely permits an already resolved action to execute. Repeated deliberation can regenerate decisions whose semantics have not changed.

Efficiency methods intervene at different points. Prompt compression and history management reduce the input replayed into a call~\cite{jiang2023llmlingua,jiang2024longllmlingua,pan2024llmlingua2,xiao2026agentdiet,lindenbauer2025complexitytrap}. Fast--slow architectures and routing change who reasons and how deeply~\cite{christakopoulou2024talkerreasoner,yang2026cogrouter,ong2025routellm}. Executable skills, observation-independent plans and compiled tool dependencies reuse or schedule prior computation~\cite{wang2023voyager,xu2023rewoo,kim2024llmcompiler}. ActionEngine further executes task programs against reusable application memory with reactive repair~\cite{zhong2026actionengine}. HarnessMM stores literal actions for the current episode and uses later feedback to decide when they can execute. Its execution record identifies the reasoning call replaced by each stored action.

Feedback can optimize instructions and agent programs through APE, STOP, ADAS, Self-Refine and Reflexion~\cite{zhou2023ape,zelikman2024stop,hu2025adas,madaan2023selfrefine,shinn2023reflexion}. The resulting harness still needs a rule for deciding which model calls to repeat during execution. Jev's typed probability interface~\cite{almeida2026jev} motivates compact candidate selection, but adding a selector does not remove the original callback automatically. We call this accounting mismatch the \emph{Substitution Paradox}: computation is saved only when eliminated reasoning exceeds added authorship, selection and recovery.

A concrete opportunity appears when slow reasoning fixes several actions before their execution feedback arrives.
After writing a program, the next compilation command can already be specified; after compilation, a successful return code can activate an already-specified test command.
We call a decision resolved when prior reasoning has fixed its action and literal arguments; the unknown tool receipt determines whether that action is ready to execute.
Here \emph{semantic uncertainty} concerns the action and its arguments, while \emph{execution uncertainty} concerns readiness. This distinction lets the harness encode a decision for future execution before its readiness is known: it stores the fixed arguments, dependencies and execution guard, then checks the guard against later feedback.

We introduce \oursfull{}, a framework built around Executable
Decision Graphs (EDGs) (Figure~\ref{fig:design}). The central
innovation is an executable representation of resolved reasoning:
each decision binds a native action and its literal arguments to
causal predecessors and public execution guards. This binding lets
a decision survive several tool transitions while later receipts
determine when it can execute. The graph thus carries prior
reasoning directly into future native execution. Actor weights,
tools and task interfaces remain fixed.

Two execution operators act on this representation. Guarded
Decision Reuse (GDR) retains the graph across feedback boundaries
and activates nodes whose dependencies and guards are satisfied.
Certified Callback Substitution (CCS) dispatches an eligible node
through the native interface in place of the registered reasoning
callback, recording both the executed action and the omitted call.
A typed selector chooses an eligible node or abstains; the original
actor handles new semantic decisions and can author a fresh graph.
Persistent graph execution avoids replaying context and regenerating
reasoning for actions already resolved.

EDGs also give harness improvement a concrete execution target.
Recursive self-improvement (RSI) revises the author prompts,
context settings and adapter rules that construct and execute
the graphs. Training task results and complete execution-token
measurements guide each revision; the retained harness supplies
feedback for the next iteration. Disjoint validation selects among
training-frozen configurations, and the selected configuration is
fixed before testing (Appendix~\ref{sec:self-evolution}). Complete-task
accounting includes graph authorship, selection, remaining reasoning
calls, fallback and unsuccessful executions.

We integrate HarnessMM into Codex and DeepSeek Harness and evaluate each against its original harness. On the same 98 HumanEval test tasks, HarnessMM preserves Codex success at 90.82\% while reducing Tokens/task (k) from 66.72 to 47.02, a 29.54\% saving. For DeepSeek Harness, success rises from 71.43\% to 85.71\% while Tokens/task (k) fall from 96.28 to 11.62, an 87.93\% saving (Section~\ref{sec:industrial-harness}). Both harnesses use fewer tokens while preserving or improving task success. An MBPP ablation isolates persistent decision reuse and callback removal, measuring 26.91\% lower cost than single-use graphs and 19.52\% lower cost than retained callbacks. Companion studies evaluate their benchmark-specific instruction and context configurations separately (Section~\ref{sec:broader}).
\paragraph{Scope of this report.}
This technical report provides a project-level account of
HarnessMM, covering its shared framework, coding and other
tool-based applications, and the visual-agent study presented
in HarnessMM-V~\cite{jiang2026harnessmmv}.
Section~\ref{hv:sec:visual-extension} consolidates the visual
methods and empirical findings from the companion study,
with shared results attributed to their original source. This report covers:

\begin{itemize}
    \item We introduce EDGs as an executable representation of
    resolved reasoning. Fixed actions and arguments, causal
    dependencies and public guards make earlier decisions available
    for execution across subsequent tool-feedback boundaries.

    \item We develop GDR and CCS as graph execution operators for
    persistent reuse and callback substitution. Integrations with
    Codex and DeepSeek Harness measure task success and token cost;
    ablations isolate graph persistence and physical callback removal.

    \item We study training-guided improvement of graph construction
    and execution rules, including the visual-agent setting developed
    in HarnessMM-V~\cite{jiang2026harnessmmv}. The report brings
    together these application studies, optimization cost accounting
    and reproducibility details.
\end{itemize}

\section{Related Work}
\label{sec:related}
\paragraph{Prompt, context and reasoning compression.}
LLMLingua and its successors compress prompt inputs~\cite{jiang2023llmlingua,jiang2024longllmlingua,pan2024llmlingua2}; AgentDiet and observation masking reduce replayed trajectory context~\cite{xiao2026agentdiet,lindenbauer2025complexitytrap}.
Thinking budgets control generated reasoning~\cite{jali2026tab}.
These methods reduce the size or depth of a call; HarnessMM targets whether that call needs to recur at all.

\paragraph{Fast--slow agents and model routing.}
Talker--Reasoner separates response and planning~\cite{christakopoulou2024talkerreasoner}; CogRouter adapts cognitive depth~\cite{yang2026cogrouter}; boundary-guided routing chooses the reasoning model~\cite{zhang2026boundaryrouting}.
Jev motivates typed compact decisions~\cite{almeida2026jev}.
Our Jev-motivated typed selector uses a candidate-ID/abstention interface, with single-token decoding when selection requires a model call.
HarnessMM studies whether such a compact decision actually removes a future reasoning callback.

\paragraph{Skills, memory and reusable programs.}
Chain-of-thought prompting exposes intermediate reasoning~\cite{wei2022chain}, Voyager retains executable skills~\cite{wang2023voyager}, and SkillLens retrieves procedural context~\cite{miao2026skilllens}.
A skill or macro can inform future reasoning or encapsulate a routine.
An EDG stores literal actions resolved for the current episode. Once their public execution conditions hold, these actions can execute without another reasoning call.

\paragraph{Agent and harness optimization.}
APE, STOP and ADAS improve prompts, programs and agent configurations from feedback~\cite{zhou2023ape,zelikman2024stop,hu2025adas}.
Self-modification methods extend the search to improvement operators~\cite{yin2025godelagent,zhang2026dgm,tan2026metarsi}.
HarnessMM uses EDGs to make resolved decisions executable at later tool boundaries. RSI refines the harness rules that construct and execute these graphs. The report also evaluates instruction optimization and context compaction in separate companion studies.
Their development protocols and costs appear in Appendix~\ref{sec:self-evolution}.

\begin{figure*}[t]
\centering
\input{figures/method_overview}
\caption{EDGs provide the execution representation for HarnessMM. The upper loop uses task feedback to improve the harness rules that construct and execute the graphs. The operation pool expresses graph changes as node and edge edits; the example composes Guarded Decision Reuse (GDR) with Certified Callback Substitution (CCS). GDR adds a guarded reuse route, and CCS replaces a reasoning subgraph with native execution of a stored action. Blue and green nodes invoke slow reasoning and fast decisions; the cyan callback executes without a model call. Dashed boxes mark the edits.}

\label{fig:design}
\end{figure*}

\section{Problem Formulation}
\label{sec:formulation}
\subsection{Reasoning calls and tool actions in a harness}
The actor parameters $\theta$ are fixed. An existing harness $h$ supplies
instructions, context and native execution rules. HarnessMM intervenes at its
reasoning callbacks: a stored action may replace a new model call. For task
$x$, unroll a finite execution under the benchmark limits into a computation
graph $\mathcal G_h(x)=(V,E)$. Nodes include reasoning, retrieval, routing,
tools, verification and recovery; edges encode execution dependencies.
For policy $\pi$, node $v$ has model-token cost $c_v$, giving
\begin{equation}
 C(\pi)=\E\!\left[\sum_{v\in V}
 \mathbf 1\{v\text{ invoked under }\pi\}\,c_v\right].
 \label{eq:graph-cost}
\end{equation}
The routing policy determines which reasoning calls repeat after public
feedback. Deterministic tools use no actor tokens but affect subsequent calls
and quality. Both node costs and invocation probabilities can change when a
continuation is substituted. We use $s_t$ for public state at transition
$t$: the request, observed tool feedback and completed action history.

\subsection{Capability-constrained computation substitution}
Let $Q_b(h)=\E[m_b(\operatorname{Run}_\theta(x;h))]$, where $m_b$ is
benchmark $b$'s official score. Relative to the original harness $h_0$,
define capability regret $R_b(\pi)=Q_b(h_0)-Q_b(h_\pi)$.
The objective is
\begin{equation}
 \min_\pi C(\pi)\quad\text{subject to}\quad
 R_b(\pi)\leq\epsilon_b
 \label{eq:objective}
\end{equation}
for every declared quality endpoint, equivalently
$Q_b(h_\pi)\geq Q_b(h_0)-\epsilon_b$. This is a policy-level constraint,
assessed by evaluating both harnesses on the same tasks.
Tolerances are fixed before evaluation; zero-regret protocols use $\epsilon_b=0$. Both harnesses are evaluated on the full task set to assess the constraint, including unsuccessful tasks. The
original tools, evaluator and limits are shared within each comparison.
We report score differences and uncertainty under the
evaluation protocol (Appendix~\ref{sec:acceptance}).

\subsection{The Substitution Paradox}
The paradox arises when fast--slow composition adds a decision without
removing its downstream deliberation. If a fast stage costs $C_f$ but
removes none of an original continuation's
$C_n$ tokens, its additive path costs $C_n+C_f$. Short output alone cannot
resolve this overhead: typed output and frequent fast-path activation
do not establish callback elimination. A beneficial substitution requires
\begin{equation}
 C_{\mathrm{removed}}>C_{\mathrm{introduced}},
 \label{eq:paradox}
\end{equation}
where introduced computation includes authorship, routing, additional
context and any fallback or recovery. This cost condition and the
capability constraint address separate requirements.

\section{HarnessMM: Graph Construction and Execution}
\label{sec:method}
HarnessMM represents reusable portions of harness execution as EDGs. A graph binds prior reasoning decisions to explicit conditions for native execution, making the route from tool feedback to the next action explicit. GDR retains the graph and activates eligible nodes; CCS replaces registered slow callbacks with execution of those nodes. The controller integrates with an existing harness through its public-state adapter, slow-call entry point and native action dispatcher.

\paragraph{Improving graph construction and execution.}
Figure~\ref{fig:design} places RSI around the execution graph. Training feedback guides revisions to author prompts, context settings and adapter rules, which determine how future graphs are constructed and executed. The operation pool illustrates the resulting node and edge changes, with GDR and CCS providing the reuse and substitution operations. Each study fixes its editable components and update procedure (Appendix~\ref{app:configuration-map}); validation selects a training-frozen configuration before testing.

At runtime, the selected rules instantiate fresh episode graphs. Each public tool receipt updates graph eligibility, and the selector chooses a stored node or returns control to the slow actor. The latter can resolve new decisions and author another graph. RSI therefore improves graph-producing and execution rules across training iterations, while GDR and CCS execute the resulting graphs within a task.

\subsection{Executable Decision Graphs (EDG)}
\label{sec:edg}
An EDG turns a resolved reasoning decision into a persistent executable node. Its action and arguments are fixed at authorship, while its readiness is determined by causal predecessors and public evidence. This binding allows readiness to change after a tool receipt without regenerating the decision. For example, a coding actor can fix a file-write action, a \texttt{make} command and a \texttt{pytest} command together. Their guards then admit the build and test actions as the required files and successful build receipts become available.

At slow decision $i$, actor $H_0$ emits the current tool action and
an EDG $G_i=(V_i,E_i)$, a bounded executable continuation.
Its future decision nodes are
\begin{equation}
 B_i=\{d_1,\ldots,d_m\},\qquad
 d_j=(a_j,\mathrm{arg}_j,\mathrm{pred}_j,g_j).
 \label{eq:edg-node}
\end{equation}
Here $a_j$ is the native action, $\mathrm{arg}_j$ its complete literal arguments, $\mathrm{pred}_j$ its predecessor and $g_j:\mathcal S_{\mathrm{public}}\rightarrow\{\mathrm{true},\mathrm{false}\}$ its public execution guard. The author fixes $g_j$ with the node; the adapter evaluates it against subsequently observed receipts. Each node has its own guard.

Figure~\ref{fig:design} depicts the harness execution graph, including reasoning, context compaction, fast decisions and tool use. The authored EDG supplies the bounded continuation of resolved actions used along its reuse routes. For the coding example, guards can check a missing file, available build inputs or a successful build. Dependencies must also be complete. Edges connect the root or an earlier node to a successor, forming an acyclic graph. Missing or rejecting receipts prevent admission; an empty graph returns control to the slow actor.

\subsection{Guarded Decision Reuse (GDR)}
\label{sec:eligibility}
GDR retains the graph across tool transitions. The green GDR panel shows how a condition guard enables a reuse route around repeated reasoning. Each observed receipt updates predecessor completion and guard truth. All action arguments must already be fixed; execution cannot require new reasoning. For node $d_j$,

\begin{equation}
\begin{split}
 \operatorname{eligible}(d_j,s_t)={}&
 \mathbf 1\{\mathrm{pred}_j\text{ completed}\}\\
 &\cdot\mathbf 1\{g_j(s_t)=\mathrm{true}\}
 \cdot\mathbf 1\{\mathrm{arg}_j\text{ closed}\}.
\end{split}
\label{eq:edg-eligible}
\end{equation}
The eligible set is $\mathcal E(s_t)=
\{d_j\in B_i:\operatorname{eligible}(d_j,s_t)=1\}$.
Guards consume observed receipts, never predicted results. Unsupported observations, unmet dependencies or arguments requiring new inference defer to $H_0$.

Implementation banks serialize the nodes and edges (Appendix~\ref{sec:slow-fast}). Literal closure is enforced by the schema and candidate-membership checks; whether prior reasoning correctly resolved an action remains a semantic assumption assessed through task scoring. A passing guard confirms the condition it checks. The task scorer determines whether the resulting action helps solve the task.

A Jev-motivated typed selector returns an eligible candidate ID or abstains, without revising arguments. A unique eligible node can execute deterministically. Empty sets and abstention invoke the slow actor. Each execution receipt updates the same graph.
Appendix~\ref{app:edg-routing-implementation} specifies the public-state adapter and native integration.

\subsection{Certified Callback Substitution (CCS)}
\label{sec:certification}
CCS converts an admitted GDR decision into an executable replacement for the next slow callback. It runs the exact authored action through the original tool interface and binds the resulting receipt to the omitted callback.
At a registered original execution boundary, the parent follows
$s_t\!\to\!H_0\!\to\!a_t$. An admitted EDG decision instead follows
$s_t\!\to\!d_j\!\to\!a_t$ through the same original tool interface.
A substitution certificate binds
\begin{equation}
\begin{split}
 \kappa_t=(&\mathrm{graph\_hash},\mathrm{decision\_id},
 \mathrm{predecessor\_receipt},\\
 &\mathrm{exact\_args},\mathrm{tool\_receipt},
 \mathrm{bypassed\_callback\_id}).
\end{split}
\label{eq:certificate}
\end{equation}
The fingerprint $\mathrm{graph\_hash}$ identifies the authored graph content, and $\mathrm{decision\_id}$ identifies its node. The remaining fields bind the predecessor's completion, exact executed arguments, actual tool result and omitted callback. The red CCS panel depicts this substitution: a certified callback replaces the redundant reasoning subgraph and retains its execution dependencies.

The audit matches this fingerprint and decision ID to literal
arguments, the completed predecessor's callback and receipt, and the
tool execution receipt. It checks the registered parent boundary and
requires
\begin{equation}
 N^{\mathrm{parent}}_{\mathrm{slow},t}-
 N^{\mathrm{HarnessMM}}_{\mathrm{slow},t}\geq1.
 \label{eq:slow-elimination}
\end{equation}
The parent count registers the original callback boundary. The certificate verifies exact native execution and absence of that request, without predicting its unobserved response or token count. Native scoring measures task correctness; complete-task totals measure cost.
Appendix~\ref{app:substitution-audit} records the receipt audit.

\subsection{Complete task cost}
\label{sec:evolution-cost}
The optimization target is complete episode cost:
\begin{equation}
 C_{\mathrm{episode}}(h;x)=\sum_{c\in\mathcal C(h,x)}
 (T_c^{\mathrm{in}}+T_c^{\mathrm{out}}).
 \label{eq:complete-tokens}
\end{equation}
The ledger charges authorship, selector inputs and outputs, added context, remaining slow calls, fallback, recovery, finalization and failed paid executions. Cached input remains counted. Deterministic checks add CPU work and no actor tokens. Historical and MiMo Codex costs use the scopes stated with those results. A complete-cost comparison includes every physical request.

Development cost $C_{\mathrm{dev}}$ is reported separately, including
training, proposals, protected validation and rejected attempts. For $N$
deployment tasks,
\begin{equation}
 \overline C_N=\frac{C_{\mathrm{dev}}}{N}
              +\E_x[C_{\mathrm{episode}}(h;x)].
 \label{eq:amortized-tokens}
\end{equation}
Shared authoring is charged exactly once to its declared development or
execution cohort. This deployment amortization is separate from reuse within an episode. Token cost does not establish latency or energy savings; those require separate measurements.

\paragraph{Selecting graph construction and execution rules.}
\label{sec:optimization}
Training develops the author prompts, context settings and adapter rules used to construct and execute EDGs. Disjoint validation selects among training-frozen configurations; the selected implementation is frozen before testing. These choices determine graph authorship and the execution controller's action contract. Each test episode instantiates its own graphs under the frozen rules and uses public feedback to choose between an eligible node and slow reasoning.
Appendix~\ref{sec:self-evolution} gives the development protocols. Appendix~\ref{app:execution-contract} specifies the execution algorithm, assumptions and structural invariants. 

\begin{figure*}[t]
\centering
\input{generated/completed_comparisons_figure}
\caption{\protect\input{generated/figure2_six_benchmark_caption}}
\label{fig:historical}
\end{figure*}

\section{Experiments}
\label{sec:experiments}

\subsection{Evaluation protocol}
\label{sec:evaluation-protocol}
All methods use the same tasks, backbone, original tools, task limits and scorer.
All compared methods, including baselines and native harnesses, receive token-efficiency prompts and an explicit objective to reduce complete-task input-plus-output token consumption while preserving task performance.

\emph{Baseline harness} is the original harness for each experiment, with the same model, native tools, limits and token-efficiency instruction as its comparison arm. In the companion studies, it omits the added benchmark-specific skill policy; in industrial harness pairs, it is Codex or DeepSeek Harness without HarnessMM.

Each benchmark scorer measures task success. Token cost sums input and output over all paid calls, including failures.
Development optimization cost is reported separately from complete-task execution.
Full-population coding comparisons use custom 20/20/60 splits: HumanEval 33/33/98 and MBPP 85/85/257, with deterministic rounding and fixed task IDs.
TRAIN supplies revisions, validation selects TRAIN-frozen candidates, and TEST measures the selected configuration.
The coding ablations evaluate the core execution method; native harness pairs evaluate its integration with fixed context and finalization settings. Figure~\ref{fig:historical} retains the companion studies. Appendix~\ref{app:configuration-map} identifies their differing mechanisms, populations and cost scope.

\subsection{Which reasoning calls does reuse replace?}
\label{sec:mechanism}
Table~\ref{tab:humaneval-mechanism} reports model-call and substitution statistics on 98 HumanEval and 257 MBPP-sanitized test tasks.
On HumanEval, the DeepSeek Harness falls from 954 to 415 paid model calls, with 147 certified callback bypasses (1.50/task); Codex falls from 524 to 499 calls, with 18 certified bypasses (0.18/task).
On MBPP-sanitized, DeepSeek Harness falls from 1,633 to 1,088 calls, with 472 certified bypasses (1.84/task), while success rises from 90.27\% to 91.83\%.
The MBPP-sanitized Codex pair records 91.83\% and 90.66\% success at 129.55 and 49.22 Tokens/task (k) (Appendix~\ref{app:pending-results}).
The original harnesses have zero certified bypasses. All rates use the full task set for each method.

\begin{table}[t]
\centering
\footnotesize
\setlength{\tabcolsep}{3pt}
\begin{tabular}{lrrrr}
\toprule
Harness & Pass (\%) & Tokens/task (k) & Calls/task & Bypasses\\
\midrule
\multicolumn{5}{l}{\textit{HumanEval ($n=98$)}}\\
Codex & 90.82\% & 66.72 & 5.35 & 0\\
\quad +HarnessMM & 90.82\% & 47.02 & 5.09 & 18\\
DeepSeek & 71.43\% & 96.28 & 9.73 & 0\\
\quad +HarnessMM & 85.71\% & 11.62 & 4.23 & 147\\
\midrule
\multicolumn{5}{l}{\textit{MBPP-sanitized ($n=257$)}}\\
Codex & 91.83\% & 129.55 & 6.93 & 0\\
\quad +HarnessMM & 90.66\% & 49.22 & 5.40 & 158\\
DeepSeek & 90.27\% & 53.98 & 6.35 & 0\\
\quad +HarnessMM & 91.83\% & 10.61 & 4.23 & 472\\
\bottomrule
\end{tabular}
\caption{Model calls and callback substitutions on HumanEval and MBPP-sanitized with the shared backbone. Calls/task counts model calls; Bypasses counts executed substitutions that skip a model call. Actor costs include input and output tokens.}
\label{tab:humaneval-mechanism}
\end{table}

The second ablation in Section~\ref{sec:core-mechanism-ablation} isolates persistent GDR reuse and CCS callback substitution with the same execution settings.

\begin{figure*}[t]
\centering
\input{figures/industrial_harness_phases}
\caption{Success and actor token costs for Codex, DeepSeek Harness and their HarnessMM variants. (a,b) use DeepSeek-flash; (e,f) use MiMo-v2.6-pro. Each harness pair shares its backbone, public task prompt, tools and scorer. Costs are reported as Tokens/task (k).}
\label{fig:industrial-harness}
\end{figure*}

\subsection{Industrial harness results}
\label{sec:industrial-harness}
We use the official Codex and DeepSeek Harness runtimes~\cite{openai2026codex,deepseek2026harness}, with the same pinned backbone, original interface and scorer within each pair.
HumanEval supplies a complete held-out comparison: Codex preserves 90.82\% success with 66.72 versus 47.02 Tokens/task (k) (29.54\% saving); DeepSeek Harness improves from 71.43\% to 85.71\% with 96.28 versus 11.62 Tokens/task (k) (87.93\% saving).
Paired outcomes show five improvements and five regressions for Codex; DeepSeek Harness has 22 improvements and eight regressions (two-sided exact McNemar $p=0.0161$).
These harnesses combine literal-bank reuse, context management and finalization, so the total reduction in calls includes all three. The MBPP ablations separately vary persistence and callback retention.
The MiMo-v2.6-pro DeepSeek Harness comparison completes the same 98-task population: baseline success is 96.94\% and HarnessMM success is 97.96\%, with 11.62 versus 4.40 Tokens/task (k) (62.12\% saving). Figure~\ref{fig:industrial-harness}(e) reports this MiMo pair separately from the Qwen comparisons; the MiMo Codex pair completes the same 98-task evaluation at 93.88\% versus 94.90\% success. Its costs are 62.60 versus 10.09 Tokens/task (k); MiMo MBPP Codex completes the same 257-task evaluation at 96.50\% versus 97.28\% success, with 47.56 versus 11.00 Tokens/task (k); the MiMo DeepSeek Harness pair completes the same 257-task population at 97.67\% versus 98.05\% success, with 6.20 versus 5.11 Tokens/task (k). Slow reasoning authors literal action banks, and CPU receipt guards execute 480 native substitutions (Appendix~\ref{app:pending-results}).

On MBPP, the complete Qwen-backed DeepSeek Harness comparison improves from 90.27\% to 91.83\%, while Tokens/task (k) fall from 53.98 to 10.61 (80.35\% saving). Slow reasoning authors executable action banks; fast guarded execution performs 472 certified substitutions. The CoT-only baseline reaches 93.39\% at 15.04 Tokens/task (k), giving a 29.46\% HarnessMM saving. Figure~\ref{fig:industrial-harness}(c) reports this held-out pair; the independent Codex comparison records 91.83\% and 90.66\% success, at 129.55 and 49.22 Tokens/task (k).

Figure~\ref{fig:industrial-harness} and Appendix~\ref{app:execution-contract} also report the earlier MBPP and GAIA2 development comparisons.

\begin{figure*}[t]
\centering
\begin{tikzpicture}[x=1cm,y=1cm,font=\scriptsize]
\definecolor{ablationFull}{RGB}{37,99,235}
\definecolor{ablationControl}{RGB}{100,116,139}
\draw[black!20,rounded corners=1mm] (0,1.05) rectangle (17.1,-2.62);
\node[anchor=west,font=\bfseries\footnotesize] at (.16,.80) {ALFWorld instruction ablations (2133 tasks)};
\node at (7.20,.40) {Success (\%) $\uparrow$};
\node at (13.80,.40) {Tokens/task (k) $\downarrow$};
\draw[black!25] (5.15,.15) -- (5.15,-2.20);
\node[anchor=north] at (5.15,-2.20) {0};
\node[anchor=north] at (9.25,-2.20) {100};
\draw[black!25] (11.65,.15) -- (11.65,-2.20);
\node[anchor=north] at (11.65,-2.20) {0};
\node[anchor=north] at (15.95,-2.20) {120};
\node[anchor=west] at (.16,-0.000) {HarnessMM};
\fill[black!4] (5.150,-0.085) rectangle (9.250,0.085);
\fill[ablationFull] (5.150,-0.085) rectangle (7.844,0.085);
\draw[black!75,line width=.45pt] (7.803,-0.000) -- (7.864,-0.000);
\draw[black!75,line width=.45pt] (7.803,-0.045) -- (7.803,0.045);
\draw[black!75,line width=.45pt] (7.864,-0.045) -- (7.864,0.045);
\node[anchor=west] at (7.914,-0.000) {65.70};
\fill[black!4] (11.650,-0.085) rectangle (15.950,0.085);
\fill[ablationFull] (11.650,-0.085) rectangle (12.776,0.085);
\draw[black!75,line width=.45pt] (12.765,-0.000) -- (12.797,-0.000);
\draw[black!75,line width=.45pt] (12.765,-0.045) -- (12.765,0.045);
\draw[black!75,line width=.45pt] (12.797,-0.045) -- (12.797,0.045);
\node[anchor=west] at (12.847,-0.000) {31.44};
\node[anchor=west] at (.16,-0.400) {w/o general guidance};
\fill[black!4] (5.150,-0.485) rectangle (9.250,-0.315);
\fill[ablationControl] (5.150,-0.485) rectangle (7.692,-0.315);
\draw[black!75,line width=.45pt] (7.676,-0.400) -- (7.716,-0.400);
\draw[black!75,line width=.45pt] (7.676,-0.445) -- (7.676,-0.355);
\draw[black!75,line width=.45pt] (7.716,-0.445) -- (7.716,-0.355);
\node[anchor=west] at (7.766,-0.400) {62.01};
\fill[black!4] (11.650,-0.485) rectangle (15.950,-0.315);
\fill[ablationControl] (11.650,-0.485) rectangle (12.678,-0.315);
\draw[black!75,line width=.45pt] (12.675,-0.400) -- (12.680,-0.400);
\draw[black!75,line width=.45pt] (12.675,-0.445) -- (12.675,-0.355);
\draw[black!75,line width=.45pt] (12.680,-0.445) -- (12.680,-0.355);
\node[anchor=west] at (12.730,-0.400) {28.69};
\node[anchor=west] at (.16,-0.800) {w/o task-family guidance};
\fill[black!4] (5.150,-0.885) rectangle (9.250,-0.715);
\fill[ablationControl] (5.150,-0.885) rectangle (7.168,-0.715);
\draw[black!75,line width=.45pt] (7.161,-0.800) -- (7.172,-0.800);
\draw[black!75,line width=.45pt] (7.161,-0.845) -- (7.161,-0.755);
\draw[black!75,line width=.45pt] (7.172,-0.845) -- (7.172,-0.755);
\node[anchor=west] at (7.222,-0.800) {49.21};
\fill[black!4] (11.650,-0.885) rectangle (15.950,-0.715);
\fill[ablationControl] (11.650,-0.885) rectangle (13.045,-0.715);
\draw[black!75,line width=.45pt] (13.036,-0.800) -- (13.051,-0.800);
\draw[black!75,line width=.45pt] (13.036,-0.845) -- (13.036,-0.755);
\draw[black!75,line width=.45pt] (13.051,-0.845) -- (13.051,-0.755);
\node[anchor=west] at (13.101,-0.800) {38.94};
\node[anchor=west] at (.16,-1.200) {w/o skills};
\fill[black!4] (5.150,-1.285) rectangle (9.250,-1.115);
\fill[ablationControl] (5.150,-1.285) rectangle (6.859,-1.115);
\draw[black!75,line width=.45pt] (6.834,-1.200) -- (6.884,-1.200);
\draw[black!75,line width=.45pt] (6.834,-1.245) -- (6.834,-1.155);
\draw[black!75,line width=.45pt] (6.884,-1.245) -- (6.884,-1.155);
\node[anchor=west] at (6.934,-1.200) {41.69};
\fill[black!4] (11.650,-1.285) rectangle (15.950,-1.115);
\fill[ablationControl] (11.650,-1.285) rectangle (12.872,-1.115);
\draw[black!75,line width=.45pt] (12.870,-1.200) -- (12.874,-1.200);
\draw[black!75,line width=.45pt] (12.870,-1.245) -- (12.870,-1.155);
\draw[black!75,line width=.45pt] (12.874,-1.245) -- (12.874,-1.155);
\node[anchor=west] at (12.924,-1.200) {34.10};
\node[anchor=west] at (.16,-1.600) {w/ another task family};
\fill[black!4] (5.150,-1.685) rectangle (9.250,-1.515);
\fill[ablationControl] (5.150,-1.685) rectangle (7.132,-1.515);
\draw[black!75,line width=.45pt] (7.103,-1.600) -- (7.155,-1.600);
\draw[black!75,line width=.45pt] (7.103,-1.645) -- (7.103,-1.555);
\draw[black!75,line width=.45pt] (7.155,-1.645) -- (7.155,-1.555);
\node[anchor=west] at (7.205,-1.600) {48.34};
\fill[black!4] (11.650,-1.685) rectangle (15.950,-1.515);
\fill[ablationControl] (11.650,-1.685) rectangle (13.201,-1.515);
\draw[black!75,line width=.45pt] (13.192,-1.600) -- (13.218,-1.600);
\draw[black!75,line width=.45pt] (13.192,-1.645) -- (13.192,-1.555);
\draw[black!75,line width=.45pt] (13.218,-1.645) -- (13.218,-1.555);
\node[anchor=west] at (13.268,-1.600) {43.28};
\node[anchor=west] at (.16,-2.000) {w/ neutral text};
\fill[black!4] (5.150,-2.085) rectangle (9.250,-1.915);
\fill[ablationControl] (5.150,-2.085) rectangle (6.846,-1.915);
\draw[black!75,line width=.45pt] (6.834,-2.000) -- (6.859,-2.000);
\draw[black!75,line width=.45pt] (6.834,-2.045) -- (6.834,-1.955);
\draw[black!75,line width=.45pt] (6.859,-2.045) -- (6.859,-1.955);
\node[anchor=west] at (6.909,-2.000) {41.37};
\fill[black!4] (11.650,-2.085) rectangle (15.950,-1.915);
\fill[ablationControl] (11.650,-2.085) rectangle (13.200,-1.915);
\draw[black!75,line width=.45pt] (13.188,-2.000) -- (13.209,-2.000);
\draw[black!75,line width=.45pt] (13.188,-2.045) -- (13.188,-1.955);
\draw[black!75,line width=.45pt] (13.209,-2.045) -- (13.209,-1.955);
\node[anchor=west] at (13.259,-2.000) {43.26};
\end{tikzpicture}
\par\vspace{2mm}
\begin{tikzpicture}[x=1cm,y=1cm,font=\scriptsize]
\definecolor{sarFull}{RGB}{37,99,235}
\definecolor{sarControl}{RGB}{100,116,139}
\draw[black!20,rounded corners=1mm] (0,1.05) rectangle (17.1,-1.87);
\node[anchor=west,font=\bfseries\footnotesize] at (.16,.80) {MBPP-sanitized GDR/CCS ablations (257 tasks)};
\node at (7.20,.40) {Success (\%) $\uparrow$};
\node at (13.80,.40) {Tokens/task (k) $\downarrow$};
\draw[black!25] (5.15,.15) -- (5.15,-1.45);
\node[anchor=north] at (5.15,-1.45) {0};
\node[anchor=north] at (9.25,-1.45) {100};
\draw[black!25] (11.65,.15) -- (11.65,-1.45);
\node[anchor=north] at (11.65,-1.45) {0};
\node[anchor=north] at (15.95,-1.45) {25};
\node[anchor=west] at (.16,-0.000) {Full HarnessMM};
\fill[black!4] (5.15,-0.085) rectangle (9.25,0.085);
\fill[sarFull] (5.15,-0.085) rectangle (8.995,0.085);
\node[anchor=west] at (9.045,-0.000) {93.77};
\fill[black!4] (11.65,-0.085) rectangle (15.95,0.085);
\fill[sarFull] (11.65,-0.085) rectangle (13.916,0.085);
\node[anchor=west] at (13.966,-0.000) {13.17};
\node[anchor=west] at (.16,-0.360) {W/o GDR (single-use EDG)};
\fill[black!4] (5.15,-0.445) rectangle (9.25,-0.275);
\fill[sarControl] (5.15,-0.445) rectangle (8.963,-0.275);
\node[anchor=west] at (9.013,-0.360) {93.00};
\fill[black!4] (11.65,-0.445) rectangle (15.95,-0.275);
\fill[sarControl] (11.65,-0.445) rectangle (14.751,-0.275);
\node[anchor=west] at (14.801,-0.360) {18.03};
\node[anchor=west] at (.16,-0.720) {W/o CCS (callbacks retained)};
\fill[black!4] (5.15,-0.805) rectangle (9.25,-0.635);
\fill[sarControl] (5.15,-0.805) rectangle (8.979,-0.635);
\node[anchor=west] at (9.029,-0.720) {93.39};
\fill[black!4] (11.65,-0.805) rectangle (15.95,-0.635);
\fill[sarControl] (11.65,-0.805) rectangle (14.466,-0.635);
\node[anchor=west] at (14.516,-0.720) {16.37};
\node[anchor=west] at (.16,-1.080) {W/o EDG execution};
\fill[black!4] (5.15,-1.165) rectangle (9.25,-0.995);
\fill[sarControl] (5.15,-1.165) rectangle (9.059,-0.995);
\node[anchor=west] at (9.109,-1.080) {95.33};
\fill[black!4] (11.65,-1.165) rectangle (15.95,-0.995);
\fill[sarControl] (11.65,-1.165) rectangle (14.434,-0.995);
\node[anchor=west] at (14.484,-1.080) {16.19};
\end{tikzpicture}
\caption{\protect\input{generated/figure4_instruction_and_sar_edg_caption}}
\label{fig:alfworld-ablation}
\end{figure*}

\subsection{Companion configurations: quality and cost}
\label{sec:broader}
Figure~\ref{fig:historical} presents ALFWorld, SearchQA, LiveMath, AIME, Terminal-Bench 2 and GAIA2 in their original order, with Baseline harness, SkillOpt, AutoSaddler and HarnessMM in each panel.
The HarnessMM label identifies each study's selected configuration. ALFWorld varies instructions; SearchQA uses input routing, a compact-rule draft and paid fallback; LiveMath changes the answer-first prompt and output cap; AIME uses a compiled decision policy with domain fallback; the archived Terminal panel uses an instruction policy. GAIA2 combines a literal bank with context handling. The MBPP ablation tests EDG persistence and callback removal directly (Appendix~\ref{app:configuration-map}).

On complete SearchQA validation, HarnessMM reaches 76.76\% EM
at 0.99 Tokens/task (k), saving 19.08\% against Baseline harness
(paired 95\% interval $[16.57,21.23]$\%), 38.57\% against SkillOpt
and 20.75\% against AutoSaddler
(Table~\ref{tab:searchqa-autosota-validation}).
On all 560 AIME test questions, HarnessMM reaches
87.14\% accuracy, compared with 86.96\% for Baseline harness,
87.86\% for SkillOpt and 88.04\% for AutoSaddler.
HarnessMM uses 5.90 Tokens/task (k), including shared authorship:
14.94\% below SkillOpt, 28.34\% above Baseline harness and 20.44\% above AutoSaddler.
The cost decomposition in Table~\ref{tab:aime-complete-cost-components}
shows 0.54 input, 4.28 output and 1.08 shared-author Tokens/task (k).
Baseline harness uses 0.22 input and 4.38 output Tokens/task (k): generated reasoning
dominates this mathematical workload. The small reduction in HarnessMM's
execution output is outweighed by its additional input and authorship.
The comparison uses one execution per question under the same frozen
model and native scoring protocol. SkillOpt uses three completed training
rounds; AutoSaddler retains its completed four-round policy.
The separate original 18-question AIME 2025 comparison remains in
Appendix~\ref{app:figure2-six-provenance}.
On all 480 GAIA2 test scenarios, HarnessMM reaches 43.96\% success
with 639.64 Tokens/task (k), compared with
42.71\% and 730.22 for Baseline harness,
46.88\% and 770.63 for SkillOpt, and
42.08\% and 772.22 for AutoSaddler.
HarnessMM reduces complete actor cost by 12.41\%, 17.00\% and 17.17\%, respectively.
The integrated harness performs 146 verified fast native executions
(1.40\% of all action attempts); input accounts for 99.47\% of its actor cost
(Table~\ref{tab:gaia2-complete-cost-components}).
It gains 1.25 percentage points over Baseline harness while
remaining 2.92 points below SkillOpt.
Each scenario runs once per method after frozen validation selection;
all unsuccessful executions remain in the quality and cost aggregates.
Appendix~\ref{app:quality-cost-followup} reports a separate quality--cost
study of development candidates and frozen GAIA2 validation results.

Under the common ALFWorld actor, HarnessMM's frozen instructions achieve
the highest observed success and lowest model-token cost among the four
instruction policies: 66.20\% at 31.12 Tokens/task (k).
SkillOpt reaches 63.29\% at 101.33 Tokens/task (k).
The paired success difference is 2.91 points (95\% floorplan-cluster
interval $[0.43,5.61]$), with 69.29\% fewer model tokens.
Table~\ref{tab:alfworld-fresh226} reports the four freshly executed
instruction-only conditions. The complete benchmark replay and accounting
are documented in Appendix~\ref{app:alfworld-fresh226}.

\paragraph{ALFWorld task-family instruction ablation.}
\label{sec:alfworld-components}
Figure~\ref{fig:alfworld-ablation} compares six frozen instruction
conditions on the same 2,133 tasks across seeds 9401, 9402 and 9403.
HarnessMM averages 65.70\% success at 31.44 Tokens/task (k). The task-family
paragraph alone reaches 62.01\% at 28.69 Tokens/task (k); general guidance alone reaches
49.21\% at 38.94 Tokens/task (k). The family-only condition is cheaper but loses 3.69
success points. Its per-seed paired 95\% floorplan-cluster intervals
are $[2.61,6.21]$, $[0.48,3.76]$ and $[2.68,6.48]$.

The remaining instruction conditions and per-seed paired intervals appear in Appendix~\ref{app:alfworld-instruction-ablation}. All use the common actor and 50-step limit. These ablations isolate instruction content, rather than EDG execution.
\subsection{GDR and CCS ablation on MBPP}
\label{sec:core-mechanism-ablation}
Our second ablation isolates GDR and CCS on MBPP-sanitized~\cite{austin2021mbpp} with DeepSeek Harness and a fixed backbone. Figure~\ref{fig:alfworld-ablation} compares four conditions on the same 257 tasks. Full HarnessMM reaches 93.77\% success at 13.17 Tokens/task (k), versus 93.00\% and 18.03 for W/o GDR (single-use EDG). GDR reduces complete cost by 26.91\%, with a task-paired adjusted interval of [4.01, 45.57]\%.

Full HarnessMM makes 407 certified substitutions, including repeated bypasses in 201 graphs. W/o CCS (callbacks retained) reaches 93.39\% at 16.37 Tokens/task (k), a measured 19.52\% cost reduction for full HarnessMM. Disabling EDG execution reaches 95.33\% at 16.19: full HarnessMM uses 18.62\% fewer tokens but solves four fewer tasks. Certificates verify removal; their independent benefit is not ablated. Appendix~\ref{app:core-mechanism-ablation} gives the shared execution settings, paired uncertainty and historical comparison.

\paragraph{What the ablations measure.}
The complete coding comparisons measure integrated harness quality and cost. The single-use and retained-callback ablations directly test persistent reuse and callback removal. Instruction-only ALFWorld measures instruction content. Certificates establish execution and omission, while native scoring establishes quality; no per-certificate token value is assumed.

\begin{figure}[tb]
\centering
\includegraphics[width=\columnwidth]{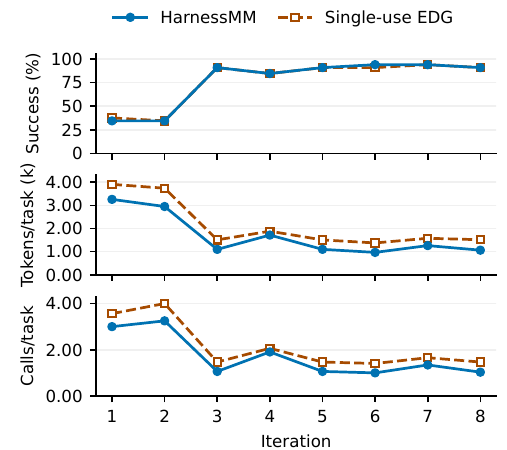}
\caption{Eight training-guided harness RSI iterations on MiniWoB++. Each comparison uses the same 32 tasks per arm, Qwen3.5-9B, full screenshots, native rewards, 15 actions and 180 seconds per episode. Calls/task counts physical model requests over all tasks. Tokens/task (k) includes complete episode authorship, control, fallback and unsuccessful calls; optimization and selection costs are separate.}
\label{fig:vigrace-evolution}
\end{figure}

\subsection{HarnessMM harness RSI}
\label{sec:vigrace-evolution}
Figure~\ref{fig:vigrace-evolution} shows eight training-guided harness RSI iterations on the same 32 MiniWoB++ training tasks. Each pair shares the actor, author prompt, full screenshots, coordinate compiler, guards, fallback, native scorer and limits; single-use EDGs permit one future execution per graph, while HarnessMM retains eligible decisions. Training feedback refines visual grounding and screenshot context. Disjoint validation selects a frozen configuration before testing.

Across the first eight iterations, HarnessMM success rises from 34.38\% to 90.63\%, while Tokens/task (k) fall from 3.26 to 1.06 (67.43\%). At the eighth iteration, both arms solve 29 of 32 tasks; persistence saves 29.83\% (1.51 versus 1.06 Tokens/task (k); paired task-bootstrap 95\% interval $[18.84,39.00]\%$). Calls/task fall from 1.47 to 1.03, with 16 second-and-later executions within retained graphs.

These are complete episode execution costs. Appendix~\ref{app:vigrace-evolution} gives the measurements and their separate development-cost scope.

\section{HarnessMM-V: Visual agent harness token efficiency}
\label{hv:sec:visual-extension}
HarnessMM-V extends EDGs to visual-agent harnesses by binding executable decisions to screenshot evidence. GDR and CCS operate on these image-bound graphs, while RSI improves the rules that construct and execute them. The separate paper, \emph{HarnessMinusMinus-V: Recursive Self-Improvement for Visual Agent Harness Token Efficiency}~\cite{jiang2026harnessmmv}, develops this setting in detail. This section describes the visual graph representation, its execution operators and their training-guided improvement.

\subsection{Method}
\label{hv:sec:method}
\subsubsection{What HarnessMM-V adds to HarnessMM}
HarnessMM supplies the EDG representation and its execution operators: slow reasoning authors graph nodes with resolved actions and public guards, GDR preserves the graph and activates eligible nodes, and CCS substitutes their native execution for later reasoning callbacks. HarnessMM-V specializes the graph for screenshot-based interaction. Its additions specify how a node is grounded in an image, how later images establish its eligibility, and how RSI improves the rules for visual graph construction and dispatch.

A coding guard can read a compiler return code, whereas a GUI action also depends on where its target remains visible. A button can move, become covered or change after an earlier action. HarnessMM-V therefore binds each stored action to author-time visual evidence, combines a current-image check with native predecessor receipts, and includes both images in the substitution record. The fast path preserves the resolved coordinates and text. A new target or changed visual evidence invokes fresh VLM reasoning.

The optimized object is the harness policy for authoring, retaining and dispatching these image-bound decisions. Each episode creates its own graph from its current screenshots. The rules can transfer to another task through the same native interface, while the target episode supplies its own coordinates, text and visual anchors. The shared EDG/GDR/CCS architecture is described in Section~\ref{sec:method}; the visual mechanisms are specified below.

\subsubsection{Image-bound executable decision graphs}
\label{hv:sec:problem}
At transition $t$, the actor receives task objective $q$, screenshot $I_t$ and public action/observation history $h_t$. A slow call at authorship time $\tau$ resolves an immediate root action and an acyclic continuation $G_\tau=(V_\tau,E_\tau)$. A future action node stores
\begin{equation}
\begin{split}
 v_j&=(\mathrm{id}_j,a_j,\theta_j,P_j,A_j,g_j),\\
 A_j&=(\tau,R_j,\phi_j,H(I_\tau)).
\end{split}
\label{hv:eq:visual-edg-node}
\end{equation}
Here $a_j$ is a native action, $\theta_j$ contains its complete literal arguments, and $P_j$ identifies its predecessors. The anchor $A_j$ records the author image, fixed region $R_j$ and reference pixel signature $\phi_j$; $H$ denotes a content hash. Guard $g_j$ specifies the configured comparison against a later screenshot. The actor resolves the action and its arguments before the node enters the graph.

The compiler checks unique node identifiers, existing predecessors, acyclic dependencies, bounds and anchor containment. It converts normalized coordinates into native pixels once, then retains the compiled coordinates, region and text. MiniWoB++ uses a model-authored region containing the action point; typing also requires a pixel-only closed-input-box predicate. VisualWebArena compiles a visible control footprint or a radius-24 patch before root execution. All arms within a comparison share the compiler. These are benchmark-specific visual bindings of the common decision interface, rather than target searches performed during reuse.

Graph capacity and repeated authorship follow each study's frozen configuration. The screenshot-only MiniWoB++ persistence study permits one authoring attempt and up to eight future nodes. The web configuration permits two ordered click/type successors and later graph authorship at slow boundaries. Other training-selected configurations retain their own declared limits (Appendices~\ref{hv:detail:sec:method} and~\ref{hv:app:longform}).

\subsubsection{GDR across screenshots and action receipts}
GDR retains the authored continuation after an action completes. It reads the resulting screenshot and native receipt to decide whether another stored node can execute. Let $\mathcal C_t$ contain nodes with successful native-interface receipts, and let $s_j^{\rm bind}$ indicate valid author-time target support. Eligibility is
\begin{equation}
\begin{split}
 \eta_j(t)={}&\mathbf1[s_j^{\rm bind}]\mathbf1[P_j\subseteq\mathcal C_t]\\
 &\cdot\mathbf1[j\notin\mathcal C_t]\\
 &\cdot\mathbf1[\operatorname{size}(I_t)=\operatorname{size}(I_\tau)]g_j(I_t).
\end{split}
\label{hv:eq:visual-sufficiency}
\end{equation}
This extends the generic public guard with current-image evidence, image-size agreement and a record of nodes already executed. MiniWoB++ checks equality of decoded pixels within $R_j$ through their signatures. The web guard additionally accepts its frozen achromatic edge-and-literal-ink signature and checks scene/input state. These checks use pixels and public receipts without another model request. Their implementation is fixed before evaluation (Appendix~\ref{hv:app:native-vwa}).

Eligible nodes execute in authored order, once per node instance. A receipt can activate a successor at the next observation boundary, so one graph can span several feedback steps. For example, the actor may resolve text entry and a later Search click together. The click waits for the text-entry receipt and a passing check of its stored button region. A passing visual patch alone cannot satisfy the predecessor requirement. A rejected guard returns control to the original actor and abandons the unresolved continuation; a later slow call may author a fresh graph if the frozen policy permits it.

Persistence extends the lifetime of a resolved continuation. Single-use EDG permits one future action and discards the remainder. Persistent GDR rechecks retained nodes after each native action until the graph is exhausted, its reuse limit is reached or no stored node is eligible. Reuse depth counts distinct future-node executions from the same graph, excluding its root.

\subsubsection{CCS with visual execution certificates}
\label{hv:sec:visual-contract}
CCS makes an eligible visual decision replace a physical VLM request. Before routing, the harness registers the callback that would request the next decision. After GDR admission, it executes the stored arguments through the original mouse/keyboard or web interface, records the native receipt and marks the node complete. The alternative branch invokes the original visual actor. The two branches share the same native action interface.

The substitution certificate extends the general record in Equation~\ref{eq:certificate} with the author and current images, anchor and visual-guard result. A compact representation is
\begin{equation}
\begin{split}
 \kappa_t=(&H(G_\tau),j,H(I_\tau),H(I_t),A_j,\\
          &g_j(I_t),r_{P_j},\theta_j,r_t,b_t),
\end{split}
\end{equation}
where $r_{P_j}$ binds predecessor completion, $r_t$ records the executed native action and $b_t$ identifies the omitted callback. Verification checks literal arguments against the native receipt, recomputes the visual evidence and confirms that the physical request ledger contains no request for that callback. The authoring request and every remaining actor, control and fallback request remain charged.

This record establishes execution under the declared visual checks and omission of the registered request. Native scoring independently determines task success: matching pixels and completed predecessors establish dispatch eligibility, rather than semantic correctness. Appendix~\ref{hv:app:visual-contract} gives the execution state machine and invariants.

\subsubsection{RSI of visual authoring, retention and dispatch}
The training loop improves the rules that instantiate and execute visual EDGs. Editable components include authoring instructions, graph mode, persistence/reuse limits and callback dispatch. The actor weights, native tools, visual-access contract, scorer and task limits remain fixed within each comparison. Visual grounding feedback can revise future authorship; it does not rewrite the literal arguments of an already authored runtime node.

Let $h_i$ denote the retained harness after iteration $i$, distinct from an episode graph $G_\tau$. With training feedback $F_i=(S_i,C_i^{\mathrm{tok}})$, the proposer $\mathcal P$ generates a candidate, evaluates it and retains a harness:
\begin{equation}
\begin{aligned}
 \widetilde h_{i+1}&=\mathcal P(h_i,F_i),\\
 \widetilde F_{i+1}&=\operatorname{Eval}_{\rm train}(\widetilde h_{i+1}),\\
 h_{i+1}&=\operatorname{Retain}(h_i,\widetilde h_{i+1};\widetilde F_{i+1}).
\end{aligned}
\label{hv:eq:rsi-harness}
\end{equation}
$S_i$ measures native success and $C_i^{\mathrm{tok}}$ includes every request in the complete training execution. Candidates share the same training population, model, tools, scorer, seeds and limits. Retention minimizes cost subject to $S\geq S_{\mathrm{parent}}-\epsilon$, with the parent, tolerance and required saving fixed in advance. Graph and receipt checks verify actual stored-action execution and callback omission. The retained harness and its measured training feedback supply the next proposal.

Protected validation selects among training-frozen candidates; the selected policy freezes before testing. Test episodes author fresh graphs under that policy, and held-out outcomes supply no edits. Thus RSI changes the visual harness across training iterations, while GDR and CCS execute decisions within each task. Optimization-inclusive comparisons charge all proposals, candidate assessments, selection and deployment costs in their stated scope; execution-only studies report task execution separately. Both use the complete-task accounting of Section~\ref{sec:evolution-cost}.

\FloatBarrier
\subsection{Experiments}
\label{hv:sec:visual-experiments}
The experiments test the added visual mechanisms in three ways: native task scoring measures whether guarded actions complete the task; persistence and retained-callback ablations separate graph lifetime from physical request removal; RSI and transfer studies evaluate learned harness rules on disjoint tasks. All methods receive token-efficiency prompts to reduce complete-task input-plus-output tokens while preserving performance. Execution costs include every request within the stated task population; optimization-inclusive comparisons additionally charge their declared training and selection procedure. Each study retains its frozen model, configuration and task population.

\subsubsection{Visual web task performance}
\label{hv:sec:native-vwa}
VisualWebArena~\cite{hv:visualwebarena} uses a custom partition of 182 training, 182 validation and 546 test tasks. Both harnesses share Qwen3.5-9B, full native screenshots and reference images, all 15 native action types and the 30-action limit. Released functional evaluators and a common blind GPT-5.6 Sol high semantic judge determine success. Validation selects training-frozen candidates before testing.

\begin{table}[t]
\caption{VisualWebArena with Qwen3.5-9B: 546 test tasks per arm and a common GPT-5.6 Sol high semantic scorer. Costs include all test actor/judge calls and method-specific learning, selected validation and setup. CCS counts certified callback omissions.}
\label{hv:tab:native-vwa}
\centering\small
\setlength{\tabcolsep}{4pt}
\begin{tabular}{lrrr}\toprule
Method & Success (\%) & Tokens/task (k) & CCS \\ \midrule
baseline harness & 14.84 & 74.31 & 0 \\
SkillOpt & 16.30 & 249.68 & 0 \\
AutoSaddler & 14.84 & 156.06 & 0 \\
Prompt Compaction & 13.74 & 95.17 & 0 \\
HarnessMM-V & 15.93 & 55.27 & 506 \\
\bottomrule\end{tabular}
\end{table}

HarnessMM-V raises success from 14.84\% to 15.93\% and reduces cost by 25.63\%, from 74.31 to 55.27 Tokens/task (k), with 506 certified omissions. Table~\ref{hv:tab:native-vwa} includes official learning and selected validation costs where applicable.

Appendix~\ref{hv:app:native-vwa} details the web configuration and scoring.

\subsubsection{Visual harness comparisons}
The Qwen3.5-9B comparisons use the same native screenshots, actions and scoring within each benchmark. Costs cover complete test execution; learning and verification are itemized separately in Tables~\ref{hv:tab:native-vwa} and~\ref{hv:tab:miniwob-cost-breakdown}.
On VisualWebArena, HarnessMM-V reaches 15.93\% success at 55.27 Tokens/task (k), compared with 5.31\% at 494.53 for DeepSeek Harness. On MiniWoB++, it reaches 62.50\% at 5.07 Tokens/task (k), using fewer execution tokens than the other visual harnesses; SkillOpt achieves slightly higher success. Codex Harness reaches 59.38\% at 44.77 Tokens/task (k).

HarnessPro uses the balanced Medium tier with normalized performance--cost selection weights $(1,1)$~\cite{hv:pan2026harnessprorecursiveharnessselfimprovement}. Its visual reproduction implements tiered evolution and generated-test selection with shared native tools and eight optimization rounds (Appendix~\ref{hv:app:rsi-common-start}).

\begin{figure}[t]
\centering
\resizebox{\columnwidth}{!}{\begin{tikzpicture}[x=1cm,y=1cm,font=\scriptsize]
\definecolor{actorNative}{RGB}{100,116,139}
\definecolor{actorOurs}{RGB}{0,121,107}
\tikzset{every node/.append style={inner sep=0pt}}
\node[anchor=west,font=\bfseries\footnotesize] at (0,.86) {GPT-5.6 Sol high actor};
\node[anchor=west] at (0,.61) {MiniWoB++; 480 tasks per condition};
\node at (3.70,.29) {Success (\%)};
\node at (6.40,.29) {Tokens/task (k)};
\node[anchor=west] at (0,-0.05) {Codex Harness};
\fill[black!5] (2.65,-0.10500000000000001) rectangle (4.0,0.0049999999999999975);
\fill[actorNative] (2.65,-0.10500000000000001) rectangle (3.83125,0.0049999999999999975);
\node[anchor=west] at (4.08,-0.05) {87.50};
\fill[black!5] (5.25,-0.10500000000000001) rectangle (6.6,0.0049999999999999975);
\fill[actorNative] (5.25,-0.10500000000000001) rectangle (6.45528,0.0049999999999999975);
\node[anchor=west] at (6.68,-0.05) {31.25};
\node[anchor=west] at (0,-0.35) {+HarnessMM-V};
\fill[black!5] (2.65,-0.40499999999999997) rectangle (4.0,-0.295);
\fill[actorOurs] (2.65,-0.40499999999999997) rectangle (3.83969,-0.295);
\node[anchor=west] at (4.08,-0.35) {88.13};
\fill[black!5] (5.25,-0.40499999999999997) rectangle (6.6,-0.295);
\fill[actorOurs] (5.25,-0.40499999999999997) rectangle (6.39142,-0.295);
\node[anchor=west] at (6.68,-0.35) {29.59};
\node[anchor=west] at (0,-0.78) {DeepSeek Harness};
\fill[black!5] (2.65,-0.8350000000000001) rectangle (4.0,-0.725);
\fill[actorNative] (2.65,-0.8350000000000001) rectangle (3.97750,-0.725);
\node[anchor=west] at (4.08,-0.78) {98.33};
\fill[black!5] (5.25,-0.8350000000000001) rectangle (6.6,-0.725);
\fill[actorNative] (5.25,-0.8350000000000001) rectangle (5.87928,-0.725);
\node[anchor=west] at (6.68,-0.78) {16.31};
\node[anchor=west] at (0,-1.08) {+HarnessMM-V};
\fill[black!5] (2.65,-1.135) rectangle (4.0,-1.0250000000000001);
\fill[actorOurs] (2.65,-1.135) rectangle (3.98031,-1.0250000000000001);
\node[anchor=west] at (4.08,-1.08) {98.54};
\fill[black!5] (5.25,-1.135) rectangle (6.6,-1.0250000000000001);
\fill[actorOurs] (5.25,-1.135) rectangle (5.84591,-1.0250000000000001);
\node[anchor=west] at (6.68,-1.08) {15.45};
\node[anchor=north] at (2.65,-1.30) {0};
\node[anchor=north] at (4.0,-1.30) {100};
\node[anchor=north] at (5.25,-1.30) {0.00};
\node[anchor=north] at (6.6,-1.30) {35.00};
\end{tikzpicture}}
\caption{GPT-5.6 Sol high visual actor on the same 480 MiniWoB++ test tasks. Costs include every call within scored executions; training and selection are separate.}
\label{hv:fig:gpt56-visual}
\end{figure}
On these 480 tasks, the Codex variant raises success from 87.50\% to 88.13\%; the DeepSeek variant raises it from 98.33\% to 98.54\%. Both save 5.30\% in scored execution cost. The DeepSeek variant uses standalone public-state encoding and issues every action through its native model loop. Appendix~\ref{hv:app:reho} gives the configurations and numerical comparisons.

\paragraph{ScreenSpot-Pro grounding.}
On all 932 ScreenSpot-Pro test tasks, HarnessMM-V raises accuracy from 47.53\% to 47.96\% and saves 20.87\% in the selected pipeline's complete cost, including training, optimization, selection, validation and testing. Literal grounding, menu alignment and integer-pixel decoding retain the label selector and native visual fallback. Appendices~\ref{hv:app:pro-full-test} and~\ref{hv:app:pro-training-validation} give the complete comparison.

\paragraph{LongForm-UI with GPT-5.6 Sol.}
The visual actor solves all 96 test tasks and saves 83.75\% in test-time tokens over the baseline. The native exact-field scorer, full-resolution screenshots and task membership remain fixed (Table~\ref{hv:tab:gpt-sol-longform-test}).

\begin{figure*}[!t]
\centering\includegraphics[width=\textwidth]{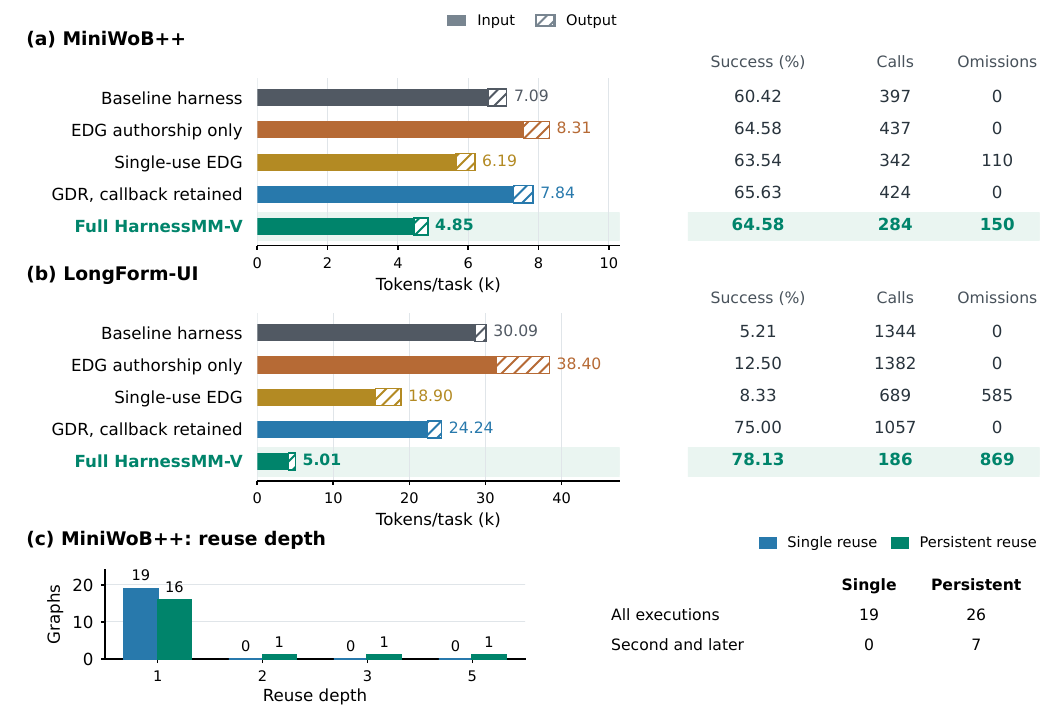}
\caption{Qwen3.5-9B ablations and reuse. (a,b) Five conditions on MiniWoB++ and LongForm-UI, with 96 test tasks per arm. Solid/hatched bars show input/output costs, including all episode calls; Calls and Omissions count model requests and certified substitutions. (c) MiniWoB++ reuse depths on 32 separate training tasks; bars count graphs, and totals count certified executions.}
\label{hv:fig:quality}
\end{figure*}

\subsubsection{Effectiveness and cost of harness RSI}
\begin{figure}[t]
\centering
\includegraphics[width=\columnwidth]{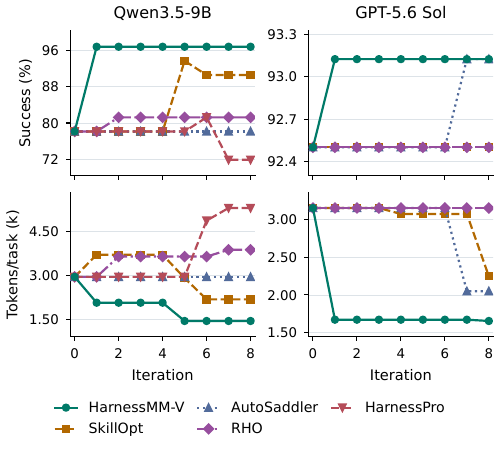}
\caption{Eight MiniWoB++ RSI rounds on 32 training tasks with Qwen3.5-9B and 160 with GPT-5.6 Sol. Each model group shares its initial harness and tasks. Curves show retained checkpoints and episode execution costs; optimization and selection are excluded.}
\label{hv:fig:matched-revisions}
\end{figure}

Figure~\ref{hv:fig:matched-revisions} compares eight RSI rounds from a common initial harness within each model group. On 32 MiniWoB++ training tasks, Qwen3.5-9B success rises from 78.13\% to 96.88\% as execution cost falls from 2.97 to 1.46 Tokens/task (k). On 160 training tasks with GPT-5.6 Sol, success rises from 92.50\% to 93.13\% and execution cost falls by 47.39\%, from 3.15 to 1.66 Tokens/task (k). Curves show retained checkpoints and exclude optimization overhead.

On 480 held-out tasks, HarnessMM-V solves 393 (81.88\%), versus 385 for SkillOpt and 382 for AutoSaddler, using 15.59\% and 6.36\% fewer measured execution tokens, respectively (Table~\ref{hv:tab:rsi-expanded-test}). RHO solves 395 tasks (82.29\%) at 9.72 Tokens/task (k). On LongForm-UI, bank-anchored RSI saves 87.18\% in complete first-use tokens and 12.87\% in elapsed time, recovering bank acquisition, optimization, candidate evaluation and validation costs within the 96-task test workload (Appendix~\ref{hv:app:rsi-first-use}).

\subsubsection{Cross-benchmark transfer}
\label{hv:sec:cross-benchmark}
We transfer frozen harnesses between MiniWoB++ and LongForm-UI through their shared mouse/keyboard interface, without target updates or selection. Source prompts, schemas, pixel preprocessing, compilation, guards and reuse policies remain fixed; only native entry and viewport adapt. Target screenshots supply fresh coordinates and values. Each target uses 32 training, 32 validation and 96 test tasks in disjoint groups. Costs cover all target test calls; source optimization and target training/validation are separate. Appendix~\ref{hv:app:transfer-settings} gives source configurations.

Qwen3.5-9B transfer to LongForm-UI improves success over single-use EDG and saves 73.49\% (schema-cluster 95\% interval: 70.04--77.34\%). With GPT-5.6 Sol high, MiniWoB++ to LongForm-UI transfer reaches 98.96\% success and saves 83.10\%; the reverse direction reaches 86.46\% at 2.13 Tokens/task (k) (Table~\ref{hv:tab:gpt-sol-training}). In the LongForm-UI direction, model calls fall from 1,056 to 159, with 729 of 888 certified omissions arising from later graph uses.

The transferred object is the policy for constructing and checking graphs. Each target episode authors fresh literal actions from its current screenshot; predecessor receipts establish when a following action can run. The source-trained policy determines which decisions persist, while target evidence controls each execution.

Of the 888 certified omissions in LongForm-UI transfer, 729 (82.09\%) occur after a graph's first use. The reduction from 1,056 to 159 physical requests therefore accompanies reuse across multiple native feedback steps. Target task scoring checks the final form independently of omission certificates.

\subsubsection{Ablation of visual decision reuse}
The five conditions compare fresh reasoning, graph authorship without execution, single-use EDG, retained callbacks and full HarnessMM-V. They share Qwen3.5-9B, screenshot-only inputs, native mouse/keyboard actions, a 15-action limit and a 180-second allowance. Graph conditions also share authorship, compilation, guards and fallback. Each benchmark uses 32 training, 32 validation and 96 test tasks from disjoint task-family or form-schema groups.

MiniWoB++ evaluates native browser rewards; LongForm-UI uses fully visible static forms requiring 9--13 operations and exact field-value scoring. Appendices~\ref{hv:app:miniwob-protocol} and~\ref{hv:app:longform} give screenshot settings, scoring and population construction.

\paragraph{Effects of persistence and callback removal.}
Figure~\ref{hv:fig:quality} compares all five conditions on 96 test tasks per arm. On MiniWoB++, persistence saves 21.62\% with slightly higher success than single-use EDG. On LongForm-UI, full HarnessMM-V raises success from 8.33\% to 78.13\% and saves 73.49\%. Every author request, control callback, fallback and unsuccessful execution is charged.

LongForm-UI requires 8, 10 or 12 literal field entries and native Submit. Full HarnessMM-V reduces slow calls from 7.18 to 1.94 per task. Appendix~\ref{hv:app:longform} gives scoring, the frozen schema-group split and length-specific results.

\paragraph{What each condition isolates.}
The authorship-only condition constructs the graph while every action still returns to the model. Single-use EDG enables one future native execution from an authored continuation. The retained-callback condition permits persistence but keeps a model request at the corresponding action boundary. Full HarnessMM-V combines persistent decisions with certified callback substitution. Comparing these conditions separates the existence of a graph, the lifetime of its decisions and the removal of physical requests. Their common compiler and guards keep the action interface fixed as these execution rules change.

\paragraph{Physical requests and task cost.}
On MiniWoB++, full HarnessMM-V uses 284 physical requests over 96 test tasks, compared with 342 for single-use EDG. Its execution cost is 4.85 rather than 6.19 Tokens/task (k), while success rises from 63.54\% to 64.58\%. On LongForm-UI, the corresponding request counts are 186 and 689, and costs are 5.01 and 18.90 Tokens/task (k). These totals charge graph authorship and subsequent calls in both conditions. The longer visible workflows provide more boundaries at which one authored continuation can replace another model request.

\paragraph{Execution depth within one graph.}
Figure~\ref{hv:fig:quality}(c) counts graph instances on separate MiniWoB++ training tasks. Single-use execution produces 19 depth-one graphs. Persistence produces 16 depth-one graphs and one each at depths two, three and five: 26 certified executions, including seven later uses. These later receipts bind to the same authored graph. Testing evaluates the selected rule on disjoint task groups.

On LongForm-UI, 869 certified substitutions include 698 later graph executions and up to eight actions per graph, identifying reuse beyond the single-use limit. Appendix~\ref{hv:app:visual-measurements} separately tests visual-guard responses to pixel interventions.

\section{Limitations}
Savings depend on provider input, cache and context accounting and on the availability of resolved actions. New semantic decisions require slow reasoning. Authorship, selection and recovery can outweigh avoided calls. Public-state adapters are task specific, and small development sets leave policy-level quality differences uncertain.

\FloatBarrier
\section{Conclusion}
HarnessMM introduces EDGs as an executable representation that carries resolved reasoning across tool-feedback boundaries. GDR retains and activates graph nodes, while CCS substitutes their native execution for repeated reasoning callbacks. RSI improves the rules that construct and execute these graphs through training feedback. Complete HumanEval comparisons on all 98 test tasks preserve Codex success at 90.82\% with 29.54\% fewer actor tokens and improve DeepSeek Harness from 71.43\% to 85.71\% with 87.93\% fewer tokens. The MBPP ablation isolates graph persistence and callback removal. HarnessMM-V extends EDGs to visual agents~\cite{jiang2026harnessmmv}, with screenshot-based interaction, grounding and transfer studies; the LongForm-UI first-use comparison also measures RSI savings after its optimization and selection costs.

\clearpage
\typeout{MAIN_BODY_NEXT_PAGE=\thepage}
{\small
\bibliographystyle{plainnat}
\bibliography{references,visual_experiments/references}
}

\clearpage
\onecolumn
\appendix
\section{HarnessMM visual harness RSI study}
\label{app:vigrace-evolution}
The eight MiniWoB++ harness RSI iterations in Figure~\ref{fig:vigrace-evolution} evaluate HarnessMM. Every RSI iteration evaluates all 32 training tasks once per arm, with identical task instances and actor seed. The actor is Qwen3.5-9B; inputs are full screenshots, the action limit is 15 and the episode allowance is 180 seconds. Single-use EDG permits one future execution per authored graph. HarnessMM retains eligible nodes across later tool-feedback boundaries. Both arms share the iteration-specific author prompt, public screenshot context, coordinate compiler, visual guards and fallback. Training feedback guides changes to these shared components across RSI iterations. The curve reports candidate execution on training tasks; the updates were not produced by an autonomous optimizer.

\begin{table}[ht]
\centering\small
\begin{tabular}{crrrrrr}
\toprule
& \multicolumn{2}{c}{Success (\%)} & \multicolumn{2}{c}{Tokens/task (k)} & \multicolumn{2}{c}{Calls/task} \\
Iteration & Single-use EDG & HarnessMM & Single-use EDG & HarnessMM & Single-use EDG & HarnessMM \\
\midrule
1 & 37.50 & 34.38 & 3.91 & 3.26 & 3.56 & 3.00 \\
2 & 34.38 & 34.38 & 3.74 & 2.95 & 4.00 & 3.25 \\
3 & 90.63 & 90.63 & 1.51 & 1.10 & 1.47 & 1.06 \\
4 & 84.38 & 84.38 & 1.88 & 1.71 & 2.06 & 1.91 \\
5 & 90.63 & 90.63 & 1.50 & 1.10 & 1.47 & 1.06 \\
6 & 90.63 & 93.75 & 1.37 & 0.96 & 1.41 & 1.00 \\
7 & 93.75 & 93.75 & 1.58 & 1.26 & 1.66 & 1.34 \\
8 & 90.63 & 90.63 & 1.51 & 1.06 & 1.47 & 1.03 \\
\bottomrule
\end{tabular}
\caption{Complete paired measurements for the eight RSI iterations on all 32 training tasks per arm. Success uses native rewards; tokens include all task executions, including unsuccessful ones.}
\label{tab:vigrace-evolution}
\end{table}

Tokens/task (k) equals complete actor input-plus-output tokens divided by 32 and 1,000. Calls/task counts every measured physical model request, including authorship, native reasoning, retained callbacks and paid fallbacks, divided by all 32 tasks. Every episode's authorship, native reasoning, control, fallback and unsuccessful requests is charged. Optimization, candidate selection, validation and historical research costs are outside these per-iteration execution measurements and remain in the separate complete ledgers. The curve therefore reports candidate execution cost. The paired interval at the eighth RSI iteration resamples tasks within the same fixed task set; there is one execution per task and arm.

\section{Implementation and selection}
\label{app:qa-development}
\subsection{QA profiles and execution}
\label{sec:instruction-routing}
\label{app:qa-profile-execution}
At round $k$, the QA harness stores profiles $h_{\mathrm D,k}$ (DELETE)
and $h_{\mathrm M,k}$ (MODIFY).
Initially, MODIFY adds five answer-normalization rules to DELETE;
the remaining four sections and answer contract are identical.
A frozen linear head selects MODIFY when $g(x)\geq\kappa$, and DELETE
otherwise. Its eight input-only features are context token count,
document count, question character and token counts, question--context
lexical coverage, quoted-span and relation-cue indicators, and named-token
count. The archived threshold maximizes discovery exact matches using
five-fold out-of-fold predictions, subject to at least 8\% token saving
and at most 5\% harm relative to SkillOpt. Deployment reads frozen
$h_{\mathrm D,K}$ and $h_{\mathrm M,K}$ without a router-model call,
generated answers or reference labels.
Interactive tasks instead use their task-family instructions.

The full-source QA adapter retrieves original rule sentences from the
selected profile, retaining normalization and answer-format rules.
It supplies these sentences and the complete context to a draft call
with a 512-token output cap. Format, source-location and explicit
answer-type checks either accept the draft unchanged or restore the
complete selected profile in a second actor call. Explicit
negative/comparison questions and multiple blanks bypass the draft.
All draft and fallback tokens are charged. Earlier compact-context and
model-selected variants have separate identities
(Appendix~\ref{app:historical-diagnostics}).

\subsection{Rule optimization and selection}
\label{sec:self-evolution}
\label{sec:acceptance}
\label{sec:efficiency-priority}
\label{sec:meta-rsi}
\label{app:metacost}
The historical textual pilot uses the original routed QA adapter and
32/32/32/128 source-group-disjoint tasks for training, inner validation,
meta validation and the final screen. This four-role protocol predates
the custom 20/20/60 comparisons. Training feedback contains responses,
official scores, token usage and successful cases to preserve.
A fixed compiler applies proposed edits to named rule slots after
checking the parent hash, schema, permitted fields and size constraints.
It rejects executable code, answer literals and case-specific facts.
Retained rules and answer contracts keep their source bytes; invalid
proposals retain the parent, and their cost is still charged.

A child passes the harness gate only if it changes the harness,
preserves the parent's EM and F1 on both validation splits, uses at most
90\% of Original's inference tokens on both, and improves at least one
quality or cost endpoint over the parent. Fixed-policy RSI ranks passing
children by meta-validation EM, then F1, then lower cost.
Rejection retains the parent; the next round obtains fresh training
feedback from the retained harness. Actor weights, routing, execution
code, task interfaces, scorers and output caps stay fixed.

The restricted MetaRSI branch~\cite{tan2026metarsi} also proposes a
strategy $\widetilde\pi_k$ from training feedback. Old and new strategies
receive the same parent and feedback and produce separate children.
Joint acceptance requires a changed strategy and a distinct child that
passes the harness gate, matches or exceeds the old-strategy child's EM
and F1 at no greater cost on each validation split, and improves at least
one endpoint strictly on meta validation. Rejection retains
$(h_k,\pi_k)$, including any previously accepted strategy change;
ordinary RSI keeps $\pi_k=\pi_0$. Protected validation examples and
answers never enter proposal prompts. Frozen final releases must improve
EM and F1 over both Original and fixed-policy RSI, save at least 10\%
of Original's inference tokens, and use no more tokens than fixed-policy
RSI. These final-screen requirements are separate from round acceptance.

Both branches run two rounds with two candidate proposals per round.
Neither accepts a harness update; MetaRSI also retains its original proposal strategy.
Six of eight proposals fail the literal-overlap guard; the two valid
three-rule edits add 0.06 and 0.06 input Tokens/task (k) on the 32 inner-validation tasks
without improving quality or output usage. Original, fixed-policy RSI,
restricted MetaRSI and Original A/A consequently retain identical final
request bytes: each solves 78.91\% of tasks (78.91 EM, 86.59 F1) using
3.02 Tokens/task (k). Detailed acquisition costs are retained in the external reproduction records. Since deployment cost is unchanged, neither branch recovers its optimization cost through token savings. The complete eight-arm screen and stage ledger
remain in \path{generated/meta_pilot_appendix.tex} and
\path{generated/meta_optimization_costs.tex}.

\section{Executable continuation implementation and historical adapters}
\label{app:edg-routing-implementation}
GDR reuses the authored graph across tool transitions, while CCS executes admitted actions in place of slow callbacks and certifies their removal.
The coding adapter serializes each EDG as an action bank.
A hash identifies the bank used by each recorded execution.
The native adapter checks predecessor completion, literal arguments and observed public guards before executing a node. Receipt errors, nonzero return codes, missing dependencies and unknown or false guards return control to slow reasoning. A unique eligible node takes a deterministic path with zero selector tokens. Frozen adapter settings and source identities remain in the external reproduction artifacts. The complete actor ledger includes authorship, selection, fallback and recovery.

The complete MBPP training set (85 tasks) SDK comparison records 153 certified substitutions
and 62 public-guard confirmations, with all physical actor costs reconciled.
Its policy is the historical structured literal-bank v23 candidate. The complete selected MBPP test set (257 tasks) comparison in Appendix~\ref{app:figure3-targeted} evaluates the separately frozen v66 candidate. The Codex compound-bank v10 candidate records
three certified substitutions but does not pass the broader adoption gate.
The training-time substitution analysis appears in Appendix~\ref{app:substitution-audit}.
The earlier configuration search uses the edit labels DELETE, BYPASS, SUBSTITUTE, COMPACT, CACHE/REUSE, TERMINATE and COLLAPSE. These labels describe changes to the surrounding harness settings; EDG, GDR and CCS define the execution method. Recovery and finalization rules retain their declared per-run settings.

\subsection{Execution variability and paired instruction effects}
For task $t$, let $x_t$ be its fixed input: the task request, supplied
context, and, for a visual task, an image $v_t$. Running the same task
more than once need not produce the same answer. We write the $r$-th
complete execution as
$y_{t,r}=\operatorname{Run}_\theta(x_t;h,z_{t,r})$,
where $\operatorname{Run}_\theta$ combines slow actor calls, fast
selection and tool execution operators. The input $x_t$, harness $h$, and actor
parameters $\theta$ are fixed across repeats; $z_{t,r}$ captures execution
conditions such as scheduling, co-batched requests, cache state, and
implementation-level variation. A client can prescribe a concurrency
and submission schedule $s$, but cannot thereby fix the server's
microbatch or numerical path. Below, $i$ indexes decisions within one
execution, and $k$ indexes offline rule-development rounds.

To measure an instruction edit, we compare the original and edited
harnesses under the same prescribed schedule. Let $m_b$ be benchmark
$b$'s official metric for an answer or episode, and let $y_t(h,s)$ denote
the response under harness $h$ and schedule $s$. The expected score
change from replacing $h$ with $h'$ is
\begin{equation}
 \tau_t(h',h;s)=
 \E\!\left[m_b(y_t(h',s))-m_b(y_t(h,s))\right].
 \label{eq:effect}
\end{equation}
The expectation averages over execution conditions within that schedule.
Repeated paired executions help separate the edit's effect from run-to-run
variation; one correct-to-wrong transition cannot do so. We report each
benchmark and quality endpoint separately.

\subsection{Inputs, outputs, and integration}
\label{sec:chisel}
\ours{} wraps an existing decision procedure $H_0$ and preserves its
original input and output contracts. The developer supplies a task adapter;
deployment then accepts the same task request $x$ as the original agent.
The adapter builds public state $s_i$ from the supplied context, current
observation, and completed interaction history. It exposes candidate
decisions $\mathcal D_i$ and a predicate $c_i(d)$ specifying which
candidates have local support.

A decision has a type and a payload. We write an action decision as
$\operatorname{Act}(a)$: it executes a tool command and, if the task
continues, receives feedback for the next decision. A terminal decision
$\operatorname{Finish}(y)$ returns an answer and ends the task.
Fixed-context QA is the terminal-decision case of this interface.
The adapter defines the domain semantics, while the shared controller
checks support and candidate membership, then executes the decision
or invokes the actor.

The original actor parameters, original tools, and evaluator remain fixed.
The integration requires a public-state adapter and does not assume that
arbitrary task semantics can be inferred automatically.

The offline RSI loop follows a fixed proposal strategy.
The current harness runs on training tasks; their execution feedback
guides candidate instruction edits. Disjoint validation tasks determine
which candidate to retain, and the retained harness supplies the next
round's training feedback. The selected configuration is frozen before
final evaluation and deployment. Appendix~\ref{app:qa-development}
records the development protocols.

\subsection{Executing a candidate or calling the actor}
\label{sec:shared-control}
At each decision, the adapter prepares a candidate $\widehat d_i$ and
its public support record. The controller applies
\begin{equation}
 d_i=\begin{cases}
 \widehat d_i, &
 \widehat d_i\in\mathcal D_i,\ c_i(\widehat d_i)=1,\\
 H_0(x,I_i,s_i), & \text{otherwise}.
 \end{cases}
 \label{eq:commit-defer}
\end{equation}
Here $I_i$ is the instruction text supplied to the original actor.
An empty candidate set, unsupported state, or rejected candidate takes
the actor path. Both routes return the same typed decision to the original interface, which commits it once. Nonterminal action feedback updates
$s_{i+1}$; a terminal answer ends execution without another state update.

Candidate production and support checks have separate roles. An action
adapter can propose a tool command without calling a model; a QA
adapter first pays for an actor-generated draft. The controller records
the selected route and rejection reason, and the cost ledger includes
all model calls made by either adapter. The benchmark scorer evaluates the selected decisions.

For action tasks, the initial actor call supplies the first transition.
Local execution then applies at every supported state. An explicit QA
bypass supplies no candidate and skips draft generation, so the
controller invokes the original answering procedure directly.

\subsection{Action adapter: CoT-authored conditional execution}
\label{sec:slow-fast}
The coding adapter separates constructing an action from deciding whether
the current feedback permits it. In a slow turn, the original CoT
actor~\cite{wei2022chain} emits a terminal command and a bounded
action bank, the serialized EDG. Each bank entry is
$b=(\ell,p,g,a)$: a label $\ell$, a predecessor $p$, a condition $g$
on public tool results, and literal tool arguments $a$. A predecessor
is the current root command or an earlier entry, so the bank defines an
acyclic execution graph. The actor declares future commands only when
their arguments are already determined; unseen data requires a new slow
turn.

After a command returns, the adapter exposes entries whose predecessor
was that command. The fast selector receives the original user request,
original tool schema, declared candidates and actual public results. It
returns one candidate label or an abstention symbol. A selected label
compiles into exactly that entry's tool command, which executes once.
The selector cannot invent commands or revise arguments. Empty banks,
unparsed results, exceeded bank limits and abstention take the original
slow path. Following each fast action, its new result determines the
next eligible entries.

The selector uses the same DeepSeek backbone as the slow actor, disables
reasoning for the selection request, and generates one control token
with model-reported log probabilities. Candidate membership and tool arguments
are checked before execution. Benchmark tests evaluate the typed selection interface. The CoT-only control uses identical instructions
and authors the same bank format with fast selection disabled.

The fast request retains public constraints, candidate IDs, complete arguments and the observations needed for their guards. It omits the full tool catalog, skill library, trajectory and slow rationale; unsupported states defer before selection. Candidate construction, typed selection and context compaction are recorded separately, with all their model costs included. Historical adapter settings and interface probes remain in the external reproduction artifacts.

Complete-task cost includes the slow actor's bank
authorship, every fast input and control token, failures and slow
fallbacks. Cached inputs remain counted. The single-encoding policy
supplies each decoded tool result once to the fast selector and binds
it to the original receipt hash; the full original history remains
available to slow reasoning. Noncanonical or unsupported observations
defer to CoT. Section~\ref{sec:industrial-harness} distinguishes the
implemented mechanism, its TRAIN comparison and final held-out evidence.

The historical ALFWorld adapter has a different candidate producer:
handwritten state tracking records pickup, transformations and placement,
then selects the first tool action in the highest-priority supported
group. It uses neither model selection nor probability scores.
Table~\ref{tab:alfworld-fresh226} and
Figure~\ref{fig:alfworld-ablation} disable that controller and compare
instructions under a common actor. These comparisons evaluate
instruction changes and the earlier handwritten controller separately.

\subsection{Terminal adapter: answer acceptance}
\label{sec:qa-fast}
An answer adapter supplies a terminal candidate to the same controller.
The recorded text-QA adapter first generates a draft from the complete
source context and retrieved original rule sentences, with fixed output
budget $L_{\mathrm{draft}}$. Its support predicate checks the answer
format, source location, and explicit answer type. Acceptance returns
the draft unchanged. Rejection calls the original actor with the full
selected instructions. An input-only applicability check bypasses draft
generation for explicit negative/comparison questions and multiple blanks.

The draft is a model call, and fallback is a second model call. The
original QA baseline has no separate verifier call to remove. This
adapter can change request and response lengths while adding calls on
rejected drafts; the action adapter instead avoids supported reasoning
calls. Both are measured with complete-task accounting. Stored
DELETE/MODIFY routing, source-rule retrieval, and the independently
evaluated offline rule-development loop are specified in
Appendix~\ref{app:qa-development}.

\section{Execution rules, algorithm and interpretation}
\label{app:execution-contract}
HarnessMM stores actions produced by the slow actor and executes them through the existing native harness. Each stored action has immutable arguments; public feedback changes only its readiness. The native request ledger identifies the reasoning callback omitted at execution. These checks let us verify reuse, while native task scores and complete costs determine whether it improves the harness.

\subsection{State and execution algorithm}
A graph instance contains an authored root action, a finite acyclic set of future nodes, literal native arguments, dependencies and public guards. Runtime state contains the completed native receipts, the remaining graph nodes and the public episode history. Graph content and runtime state are different objects: completing a predecessor changes the latter without rewriting the authored action.
\begin{enumerate}
\item At a native decision boundary, register the parent callback and ingest the preceding public tool receipt. Validate its association with the action that produced it.
\item Construct the eligible set from unresolved nodes whose dependencies are complete, public guards pass and arguments meet the adapter's closure checks.
\item If exactly one node is eligible, use the declared deterministic route. If several are eligible, the configured typed selector chooses a candidate ID or abstains. Charge any model-based selector input and output.
\item On admission, execute the selected literal arguments through the original native interface. Record the result and compare it with the authored node. Emit an omission certificate only after checking execution and absence of the registered slow request. Mark the node completed and retain the unresolved remainder of the graph.
\item On rejection, an empty set or abstention, invoke the original slow actor with its public task context. Record the author request and any fallback. A newly authored graph replaces the unresolved remainder; already executed receipts remain in the episode history.
\item Continue under the common benchmark limits. A terminal answer uses the same commitment point and stops. The native scorer evaluates the complete task; the request ledger charges every paid call.
\end{enumerate}
The steps above define the execution loop. The coding adapter uses literal terminal commands and observed tool results. Instruction-only and answer-acceptance studies use different candidate producers, stated in Appendix~\ref{app:edg-routing-implementation}; those studies evaluate instructions and answer acceptance, respectively, without testing persistent coding graphs.

\subsection{Closure, guards and structural invariants}
\paragraph{Literal closure and semantic closure.}
Literal closure is a syntactic restriction: all executable arguments have concrete values, with no placeholder to bind from unseen feedback. It is checked by the action schema and candidate-membership audit. Semantic closure is stronger: prior reasoning has determined the action's meaning and its future applicability is captured by the declared guard. A concrete command can still be semantically wrong. Parsing literal arguments cannot determine whether the action is appropriate for the task.

For a fixed authored decision $d$ and public state $s$, let $\operatorname{valid}(d,s)$ mean that the action has its intended task meaning at that state. Interpreting guard acceptance as semantic permission additionally assumes
\begin{equation}
 \operatorname{eligible}(d,s)=1\ \Longrightarrow\ \operatorname{valid}(d,s).
\end{equation}
This implication depends on the adapter and the authored guard. The implementation enforces the left-hand checks; complete native task evaluation measures their quality consequences. There is no statewise correctness guarantee for arbitrary tools or tasks.

\paragraph{Argument immutability.}
Assume the configured adapter uses the validated graph and native interface, the graph objects are not modified after authoring, and the recorded receipts are faithful. A reused node executes the literal arguments authored for that node. The typed selector returns only a node ID; it has no argument-generation interface. Native dispatch reads the selected stored arguments, and certificate verification compares those values with the tool receipt. This establishes an execution invariant under the stated assumptions, independently of task success.

\paragraph{Dependency-respecting execution.}
Initially, no future node is completed. At each transition, eligibility admits only a node whose predecessor receipts are in the completion map. A completed node is removed from the unresolved set. Induction over transitions therefore gives predecessor-respecting execution with at most one dispatch per node instance. The initial acyclicity check prevents a future node from requiring its own completion. A later author call can create an identical graph, but it is a separate paid authoring event.

\paragraph{Omission evidence.}
A valid certificate binds the authored graph instance, selected node, exact native arguments, predecessor receipt, tool receipt and registered callback. The physical request ledger independently establishes absence of the specified slow request at that boundary. This verifies which route executed. The certificate does not reveal what the omitted call would have returned or cost, or whether the task outcome improved. We measure quality and token savings by comparing complete runs on the same task sets.

\subsection{When reuse saves tokens}
Let $A$ denote additional author and selector actor tokens and $F$ additional fallback tokens. If $K$ eliminated callbacks would have costs $c_1,\ldots,c_K$, a same-trajectory token calculation gives net saving only when $\sum_j c_j>A+F$. Under the simplifying assumption of a common callback cost $c>0$, the condition is $K>(A+F)/c$. This relation describes the dependence on reuse coverage; it is not a prediction of measured saving, because the omitted calls are not executed and the substituted trajectory can differ.

We measure savings from the difference in total tokens across all episodes for each method. Every arm uses the same independent task membership and execution contract, and each arm's numerator includes its actual author, selector, slow, finalization and paid unsuccessful calls. Shared authoring is assigned once to its declared population. Cached provider input remains counted under the token protocol. CPU guards and native tools affect time and behavior, but their work cannot be added to a token total without an independently defined resource conversion.

\subsection{Relation to executable planning}
ReWOO separates planning from tool observations and supports dependency-bound evidence variables~\cite{xu2023rewoo}; LLMCompiler plans and dispatches tool dependencies for parallel execution~\cite{kim2024llmcompiler}. Voyager develops a library of executable skills reused across tasks~\cite{wang2023voyager}. ActionEngine constructs application-level state-machine memory, compiles task programs and repairs execution reactively~\cite{zhong2026actionengine}. HarnessMM shares executable planning and reuse with these approaches. It stores literal actions for the current episode, activates them from public receipts without generating new arguments, and records the callback omitted at native execution. We do not compare these planning systems experimentally, so the present evidence establishes the contract and benefits against the declared native harnesses rather than superiority over all executable planners.

\subsection{Decision reuse and benchmark-specific settings}
\label{app:configuration-map}
The core execution transformation has three operations: author literal future actions, retain and activate them using public feedback, and dispatch an eligible action while omitting its registered callback. Its controller selects an action route. A surrounding configuration fixes the author prompt, context serializer, native adapter and finalization rules before the run. Those settings determine what the controller receives; the controller does not search over them during an episode.

The reported studies use the following configurations. Some selected HarnessMM configurations change instructions or context without persistent execution; the table identifies these cases. The original run identities and exact configuration files remain in the external reproduction bundle.
\begin{center}\small
\begin{tabular}{>{\raggedright\arraybackslash}p{.17\linewidth}>{\raggedright\arraybackslash}p{.39\linewidth}>{\raggedright\arraybackslash}p{.34\linewidth}}
\toprule
Study & Fixed surrounding intervention & Role of executable reuse\\
\midrule
Coding harness pairs & Literal-bank author prompt, compact public context and harness-specific finalization & EDG/GDR/CCS enabled; complete pair measures the integrated configuration\\
MBPP mechanism ablation & Common author prompt, schema, public context, native tools and limits & Change graph execution, persistence or callback retention to test the core mechanism\\
ALFWorld & Selected task-family and general instructions under a common actor & Instruction-only comparison; local controller disabled\\
SearchQA & Frozen input-only profile routing, retrieved rule sentences, full-source draft and paid fallback & Terminal answer acceptance; no persistent multi-action graph\\
LiveMath & Answer-first prompt, disabled thinking and reduced output cap & One call per question; no callback substitution\\
AIME & Frozen compiled decision policy, shared authorship and domain fallback & Measures the selected policy; execution or abstention follows its fixed applicability rules\\
Terminal-Bench overview & Archived AutoSaddler-derived instruction policy & Instruction-policy snapshot; no certified EDG execution claim\\
GAIA2 & Frozen literal-bank authoring and public-context handling & Integrated native execution; verified fast actions and complete costs reported together\\
\bottomrule
\end{tabular}
\end{center}
These settings reduce cost in different ways. Shorter prompts or context change the tokens in a remaining call; EDG execution changes whether a registered call occurs. The MBPP ablations keep instruction, context and finalization settings fixed while varying graph reuse and callback execution. No experiment in this paper trains a general policy that dynamically composes these optimization techniques.

The token-efficiency prompt and objective apply to every comparison arm, including Baseline harness, SkillOpt, AutoSaddler, the native industrial harnesses and HarnessMM. Baseline harness retains the token-efficiency instruction while omitting the added skill policy. Every arm is instructed to reduce complete-task input-plus-output token consumption while preserving task performance. Method-specific mechanisms and native execution limits follow the configurations above.

\subsection{What each experiment measures}
\begin{center}
\begin{tabular}{p{.24\linewidth}p{.31\linewidth}p{.35\linewidth}}
\toprule
Evidence & Measured intervention & Supported interpretation\\
\midrule
HumanEval and MBPP harness pairs & Selected HarnessMM configuration versus its original harness & Complete task quality and actor cost for that implementation\\
MBPP single-use graphs & Limit each graph to one reused action & Effect of retaining graphs across tool calls\\
MBPP retained callbacks & Invoke the registered slow callback while retaining graph execution & Effect of callback removal with tasks and execution settings fixed\\
Native execution certificates & Match authored actions, public receipts and absent requests & Structural execution and callback omission\\
ALFWorld instruction comparison & Change instructions under a common actor & Effect of selected instruction content, not EDG substitution\\
SearchQA and AIME studies & Benchmark-specific selected policies and cost components & Their declared quality--cost trade-offs, with actual phases retained\\
\bottomrule
\end{tabular}
\end{center}
The integrated coding results also change context management and finalization. Total call-count reduction can therefore exceed certificate counts; these quantities are not expected to agree. The MBPP ablation supplies a more direct intervention test but does not isolate the intrinsic value of recording a certificate from the value of exclusive routing. Certificates are a measurement mechanism, and their logging overhead is included where it makes model calls.

\subsection{Historical development evidence}
Historical MBPP comparisons on 85 training tasks provide development mechanism evidence: Codex improves from 91.76\% to 94.12\% at 67.21 versus 59.78 Tokens/task (k); DeepSeek Harness improves from 85.88\% to 94.12\% at 15.46 Tokens/task (k) versus 11.91 Tokens/task (k).
The DeepSeek candidate records 153 certified substitutions and 62 public-guard confirmations.
Across 32 GAIA2 development scenarios, Codex improves from 37.50\% to 40.62\% at 907.06 versus 825.99 Tokens/task (k), and DeepSeek Harness improves from 40.62\% to 46.88\% at 891.75 Tokens/task (k) versus 750.53 Tokens/task (k).
Figure~\ref{fig:industrial-harness} preserves the earlier four panels and adds two separate MiMo panels; the appendix records their evidence phases.

The w/o skills condition averages 41.69\% at 34.10 Tokens/task (k); mismatched-family and
neutral instructions of equal token length reach 48.34\% at 43.28 Tokens/task (k) and 41.37\%
at 43.26 Tokens/task (k). HarnessMM gains 24.00 points over the w/o skills condition while using
7.81\% fewer tokens, and gains 24.33 points over neutral text with
27.33\% fewer tokens. Every action uses the common actor and the same
50-step cap. Whiskers show the observed seed range; task identities are
not independent across repetitions.
Appendix~\ref{app:alfworld-instruction-ablation} preserves all per-seed
counts, paired intervals and costs, including the two original seed-9401
reference conditions reused exactly once.
These instruction ablations measure content effects under a common actor.
\subsection{Complete quality--cost trade-offs}
\label{sec:negative}
Across 18 Terminal-Bench development tasks, Codex success falls from 77.78\% to 55.56\% while Tokens/task (k) fall from 3426.78 to 1089.95; DeepSeek Harness falls from 55.56\% to 50.00\% at 1586.15 versus 1144.32 Tokens/task (k).

A complete CoT/Jev TEST comparison executes verified fast actions in 94.44\% of episodes, yet costs 127.07\% more than the original harness and loses 7.29 success points.
It saves only 1.04\% against CoT-only, with input tokens accounting for 98.53\% of the actor ledger.
The coding control on 18 training tasks likewise records 10, 12 and 13 successes for the original harness, CoT-only and CoT/Jev, while CoT/Jev costs 30.37\% more than the original harness and 3.74\% more than CoT-only.
Appendix~\ref{app:industrial-harness} preserves the complete controls and their costs.

\subsection{Reproduction and tool permissions}
The executable adapter requires a declared native boundary, literal action serialization, public-state guards, callback registration and complete request logging. External reproduction artifacts retain frozen task memberships, model and software identities, integer costs, selection configuration and complete native outcomes. Shared optimizers use training evidence; validation selects among frozen candidates, and selected held-out configurations remain unchanged. Benchmarks with historical development populations retain their actual phase rather than being relabeled as final tests.

The controller preserves the parent's tool permissions and action space. It does not enlarge those permissions or establish a general safety guarantee. Incorrect authorship, incomplete guards and changed tool semantics can affect task quality even when the structural certificate is valid. These cases are handled by the native evaluator and the declared quality constraint, not by treating omission evidence as correctness evidence.

\section{GDR and CCS ablation}
\label{app:core-mechanism-ablation}
We evaluate four conditions on the complete MBPP-sanitized release using a custom 85/85/257 training/validation/test partition, seed 20261005. DeepSeek Harness dsh-v0.2.0-rc.1 uses the shared backbone. All arms share the author prompt and schema, public context, native buffered parser, tools, decoding settings, one attempt per task, two workers per arm, and 300-second actor/30-second verifier limits. The tool working directory is \texttt{/app}; the latest public-check call and its observed result remain in public context. Authored public assertions must match the original problem line by line. Rejected banks retain their authoring cost and return to paid native slow reasoning. Disabled-thinking requests use the same transport mapping across arms; task reasoning and output limits remain fixed.

TRAIN supplies method revisions. The only candidate meeting the training criteria is frozen before validation. The prespecified overall success criterion then selects it before TEST. Validation trajectories and test outcomes supply no method revisions. The final comparison contains all 1,028 task-arm executions. Tokens/task (k) include physical input and output from authorship, selectors, control, native recovery and unsuccessful executions, divided by 257 and 1,000.

Full HarnessMM combines GDR and CCS: GDR retains the executable graph across tool transitions; CCS replaces slow callbacks with authored actions and certifies their removal. W/o GDR (single-use EDG) discards the graph after its first certified execution, preserving its author and full authoring cost. W/o CCS (callbacks retained) retains a paid slow callback to regenerate each selected tool action. This intervention isolates callback substitution, not the independent benefit of certification. Without EDG execution preserves the author and its cost while disabling executable graph use.

\begin{table}[t]
\centering\small
\begin{tabular}{lrrr}
\toprule
Arm & Correct (\%) & Tokens/task (k) & Bypass/task\\
\midrule
Full HarnessMM & 93.77 & 13.17 & 1.58\\
W/o GDR & 93.00 & 18.03 & 0.82\\
W/o CCS & 93.39 & 16.37 & 0.00\\
W/o EDG execution & 95.33 & 16.19 & 0.00\\
\bottomrule
\end{tabular}
\caption{Results on all 257 test tasks with the same original public assertions in every condition. Every arm includes all 257 tasks and complete actor costs.}
\label{tab:core-mechanism-results-v5}
\end{table}

Full HarnessMM performs 407 certified bypasses; 201 graphs are reused twice. W/o GDR performs 212 bypasses and discards 212 graphs. The retained-callback condition performs 257 paid regenerations and zero certified bypasses; disabling EDG produces zero bypasses. Full HarnessMM solves two more tasks than W/o GDR while reducing measured complete cost by 26.91\%. Compared with retained callbacks it solves one more task at 19.52\% lower cost. Relative to disabled EDG, its 18.62\% cost saving accompanies four fewer correct tasks.

\begin{table}[t]
\centering\small
\begin{tabular}{lrr}
\toprule
Full versus control & $\Delta$ correct (pp) & Cost saving (\%)\\
\midrule
W/o GDR & +0.78 [-3.50, +5.06] & 26.91 [4.01, 45.57]\\
W/o CCS & +0.39 [-2.72, +3.50] & 19.52 [-2.07, 38.68]\\
W/o EDG execution & -1.56 [-5.45, +2.33] & 18.62 [-9.53, 39.78]\\
\bottomrule
\end{tabular}
\caption{Task-paired differences and full-population cost savings, with Bonferroni-adjusted $98\frac{1}{3}\%$ bootstrap intervals over three contrasts. All arms are resampled jointly by independent task for 10,000 draws, seed 20261009.}
\label{tab:core-mechanism-contrasts-v5}
\end{table}

The reuse cost interval excludes zero; all three correctness intervals include zero. Callback and EDG cost intervals include zero. One execution per task quantifies task variation, rather than between-run variation.

\subsection{Earlier execution ablation}
\label{app:core-mechanism-ablation-history}
Figure~\ref{fig:alfworld-ablation} complements the ALFWorld instruction ablation with execution interventions on MBPP-sanitized~\cite{austin2021mbpp}. Instructions, authoring format, public guards and dependency checks are fixed across the HarnessMM conditions.

\subsubsection{Benchmark and evaluation settings}
We use the complete 427-task MBPP-sanitized release with a custom 85/85/257 training/validation/test partition, seed 20261005. Deterministic rounding implements the required 20/20/60 ratio; task membership is disjoint and common to all arms. The parent is DeepSeek Harness dsh-v0.2.0-rc.1 with the shared backbone. All conditions share the public task interface, native tools, decoding settings, two workers per arm, one execution per task and 300-second actor/30-second verifier deadlines. Correctness evaluates every released assertion. Reference implementations are absent from the actor workspace.

Training supplies the structured author and frozen intervention definitions. The complete training comparison uses 85 tasks; disjoint validation selects frozen candidates without returning trajectories or failures to optimization. The historical evaluation uses the complete 257-task test population for every arm. Test outcomes supply no selection or revisions; the external archive retains every arm and execution.

\subsubsection{Execution interventions}
The original harness uses ordinary slow reasoning without EDG authoring or fast routing. Full HarnessMM combines the structured EDG, public eligibility checks and certified callback bypass. W/o GDR (the single-use GDR control) preserves the author and graph capacity, but discards unused nodes after its first certified execution. Unused authored nodes still incur cost.

The W/o CCS condition keeps fast selection but asks the same slow actor to regenerate exactly one tool action before execution. Changed arguments invalidate dependent nodes; invalid regeneration returns control to slow reasoning. A retained callback receives no substitution certificate. Graph authorship, selector input and control, regeneration, fallbacks and unsuccessful executions all contribute to Tokens/task (k), normalized by the full 257-task population. Development and selection costs remain separate from final execution.

\subsubsection{Complete results and paired uncertainty}
\begin{table}[t]
\centering\small
\begin{tabular}{lrrr}
\toprule
Arm & Correct (\%) & Tokens/task (k) & Bypass/task\\
\midrule
Original harness & 87.55 & 67.84 & 0.00\\
Full HarnessMM & 90.27 & 16.15 & 1.79\\
W/o GDR (the single-use GDR control) & 91.44 & 15.87 & 0.92\\
W/o CCS (callbacks retained) & 91.83 & 17.82 & 0.00\\
\bottomrule
\end{tabular}
\caption{Results on all 257 test tasks. A bypass is a certified eliminated slow callback. Cost includes all actor input and output over all tasks.}
\label{tab:core-mechanism-results}
\end{table}

Full HarnessMM reduces cost by 76.19\% relative to the original harness and increases observed correctness by 2.72 percentage points. The paired 95\% interval for the correctness difference is $[-1.95,7.39]$ points; it does not establish a significant accuracy gain. Full HarnessMM makes 459 certified substitutions, with at least two bypasses in 228 authored graphs. W/o GDR (the single-use GDR control) makes 237 substitutions and discards 237 graphs after one use. Recorded slow calls per task rise from 4.49 for full HarnessMM to 5.42 for the single-use GDR control and 6.41 when callbacks are retained. The retained-callback condition records 281 callbacks and 280 regenerated actions, with zero certified bypasses.

\begin{table}[t]
\centering\small
\begin{tabular}{lrr}
\toprule
Removal versus full & $\Delta$ correct (pp) & Cost ratio\\
\midrule
W/o GDR (the single-use GDR control) & +1.17 [-3.11, +5.45] & 0.98 [0.81, 1.22]\\
W/o CCS (callbacks retained) & +1.56 [-3.11, +6.23] & 1.10 [0.90, 1.39]\\
\bottomrule
\end{tabular}
\caption{Task-paired differences and cost ratios relative to full HarnessMM, with the original five-condition study's Bonferroni-adjusted $98\frac{1}{3}\%$ bootstrap intervals. We resample all 257 tasks jointly across conditions for 10,000 draws, seed 20261007. Cost ratios recompute cohort totals.}
\label{tab:core-mechanism-contrasts}
\end{table}

The additional EDG reuse is active, but the single-use GDR control has slightly higher correctness and lower aggregate cost in this comparison. Retaining slow callbacks increases measured cost by 10.34\% and correctness by 1.56 points; the paired intervals include equal cost and correctness. One execution per task measures task-level variation rather than run-to-run variability.

\section{HumanEval calls, substitutions and paired outcomes}
\label{app:causal-attribution}
\subsection{HumanEval results}
All four arms use the same 98 test tasks.
\begin{center}
\begin{tabular}{lrrrrr}
\toprule
Condition & Success (\%) & Input/task (k) & Output/task (k) & Tokens/task (k) & Paid calls\\
\midrule
Codex & 90.82\% & 64.47 & 2.26 & 66.72 & 524\\
Codex+HarnessMM & 90.82\% & 43.66 & 3.35 & 47.02 & 499\\
DeepSeek Harness & 71.43\% & 92.45 & 3.83 & 96.28 & 954\\
DeepSeek Harness+HarnessMM & 85.71\% & 7.09 & 4.53 & 11.62 & 415\\
\bottomrule
\end{tabular}
\end{center}
The Codex+HarnessMM and DeepSeek Harness+HarnessMM conditions contain 18 and 147 unique valid substitution certificates, respectively; both original harnesses have zero. Each binds literal arguments, an original tool, a receipt hash,
a predecessor and a bypassed callback ID; the audit checks no intervening
slow request and unchanged slow-sequence count at the boundary.
The certificate rates are 18/98=0.18 and 147/98=1.50 per task. Complete execution cost is 11.62 Tokens/task (k); it includes the whole cohort rather than only certified substitutions.

On the independent MBPP-sanitized test set (257 tasks) DeepSeek Harness and its HarnessMM condition achieve 232/257 and 236/257 successes, respectively. Paid model calls total 1,633 and 1,088, and valid certificates total zero and 472. Complete cost is 53.98 and 10.61 Tokens/task (k), respectively. These benchmark populations are reported separately.

The 25 and 539 fewer paid calls are differences between complete runs on the same tasks, not counts of locally certified callback elimination.
Input/task (k) declines by 20.80 for Codex and 85.35 for DeepSeek; output/task (k) increases by 1.10 and 0.70. The net Tokens/task (k) reductions are 19.71 and 84.65.

Paired task outcomes give 5 improvements, 5 regressions and 88 unchanged
outcomes for Codex; DeepSeek gives 22, 8 and 68. The two-sided exact
McNemar test conditions on the discordant tasks and yields $p=1$ and
$p=0.0161248$, respectively.

\subsection{Task adapters and evaluation populations}
Local decisions require a task-specific public-state adapter and support
rules. The interactive measurements use textual ALFWorld observations;
they do not establish visual-agent improvement. The ALFWorld instruction
comparison transfers archived optimized policies without rerunning
their optimizers. The custom test uses
one seed and shares some floorplans across partitions. The ALFWorld
comparisons and text-QA studies use previously exposed public data.
GQA grounding and generic instructions of equal token length both score
65\% in the small visual pilot. The separate rule-development pilot
accepts no updates.

\subsection{Training-time substitutions and model calls}
\label{app:substitution-audit}
This analysis uses complete comparisons on all 32 training tasks. Cached
inputs, failed episodes, fast requests and slow fallback remain charged.
The shared author is reused physically and charged once to each bank arm.
No validation or TEST trajectory supplies these diagnoses.
\begin{table}[htbp]
\centering\small
\begin{tabular}{lrrrr}
\toprule
Run/arm & SR (\%) & Tokens/task (k) & Slow/ep & Fast/ep \\
\midrule
Original harness & 40.62 & 891.75 & 15.69 & 0.00 \\
CoT-only v20 & 28.12 & 1351.62 & 21.66 & 0.00 \\
CoT/Jev v20 & 34.38 & 1602.29 & 19.31 & 8.78 \\
CoT/Jev v22 & 43.75 & 753.13 & 12.62 & 1.69 \\
CoT/Jev v23 & 37.50 & 1071.85 & 15.72 & 5.28 \\
\bottomrule
\end{tabular}
\caption{Complete TRAIN costs and scores. Slow and fast calls are counted separately. These are development results.}
\label{tab:sar-train-attribution}
\end{table}
In v20, 281 fast attempts produce 249 locally bypassed slow callbacks;
32 abstentions incur 212.17 slow-fallback Tokens/task (k). Fast requests cost
50.46 Tokens/task (k), including 18.91 input Tokens/task (k). Their removal alone would save
only 1.18\% of the complete bank-arm cost. The arm still makes 618 slow
calls against 502 baseline attempts. It uses 79.68\% more tokens and loses
6.25 success points despite fast actions in 93.75\% of episodes.

The passive benchmark audit verifies 227 local substitutions in 31
episodes; 22 compiled actions in one failed SDK episode remain unresolved.
Bank-bearing slow requests jointly perform reasoning and candidate
construction, so their full request cost is not an isolated authoring cost.

The v22 revision reduces slow calls from 15.69 to 12.63 per episode
and saves 15.55\% at 43.75\% versus 40.62\% baseline success. Renewing banks
at every slow step in v23 raises cost by 20.20\% and loses one success.
The external reproduction records contain the full training analysis and all earlier interface checks.

\subsection{Additional related-work and adapter context}
Model evaluation varies with task
sampling, prompt choice, random seeds, execution order, and numerical/runtime
behavior \cite{polo2024prompteval,yuan2025nondeterminism,fu2026tokenprob}.
Repeating one task characterizes its execution distribution, with the task
remaining the independent sampling unit.
We use paired source-group estimates, retain all comparison endpoints, and
separate exploratory inspection from frozen comparisons
\cite{dror2018significance,saravanan2020hierarchical,dror2017replicability}.
Crossed and nested dependence must be treated explicitly rather than
pooling all response cells as independent samples \cite{owen2007pigeonhole}.
Efficient serving mechanisms such as PagedAttention address execution
resources \cite{kwon2023vllm}; repeated-request controls characterize
variability under the chosen runtime.

Several evaluation methods study these sources of variation. FormatSpread quantifies spurious prompt-format sensitivity
\cite{sclar2024formatspread}; ProSA and POSIX examine prompt sensitivity
at finer granularity \cite{zhuo2024prosa,chatterjee2024posix}; PromptEval
estimates performance distributions over prompts efficiently
\cite{polo2024prompteval}. PARC analyzes language/image transformations
in vision-language models, with certainty derived from conformal answer
scores rather than repeated execution \cite{schmalfuss2025parc}.
Our comparisons combine instruction-content controls with repeated
execution and image interventions. They examine how these sources of
variation affect the interpretation of a quality--cost comparison.

Vision-language systems combine image representations with instruction
following, and visual-programming approaches expose intermediate operations
that can be inspected or recombined \cite{liu2023visualinstruction,suris2023vipergpt}.
These approaches motivate treating instructions as controls over visual
decisions. Balanced and compositional benchmarks, contrastive examples, and image
interventions examine complementary aspects of this problem
\cite{goyal2017vqav2,hudson2019gqa,johnson2017clevr,niu2021cfvqa}.
We compare original and shuffled images on the same tasks to test whether responses depend on the image.
The same commit-or-defer interface accepts terminal answers. A text-QA
adapter obtains an actor-generated draft, checks its source, type, and
format, and either returns it or calls the original answering procedure.
The draft and fallback are both charged. The action adapter can avoid
reasoning calls, whereas the answer adapter changes request and response
lengths. Both preserve the benchmark task interface, and complete-task
accounting covers every model call along either path.

\section{Benchmark results}
\label{app:benchmark-results}
\subsection{ALFWorld instruction comparison and skill removal}
\label{app:alfworld-fresh226}
Figures~\ref{fig:historical}(a) and~\ref{fig:alfworld-ablation} use the same
2,133 test tasks and seed 9401. The complete 3,553-task official TRAIN
pool is partitioned into 710 training, 710 validation and 2,133 test tasks.
Goal/scene trial variants stay together; the 240 known predecessor
development tasks remain in training. This is a custom partition of
previously exposed public data, with some floorplans shared across splits.
The selected instructions and common actor are frozen before fresh
execution. Training supplies six task-family operability checks per
condition. All 710 validation tasks per condition must pass benchmark replay
and cost verification before the complete test phase opens.
No validation or test outcome selects or changes a policy.

All four displayed conditions share Qwen3.5-9B, temperature zero,
disabled thinking, two-step history, a 2,048-token response cap and the
benchmark 50-step limit. The instruction comparison deploys archived
optimized artifacts without rerunning SkillOpt or AutoSaddler optimization.
The skill-removal pair compares the frozen HarnessMM instructions with empty
task-family instruction strings, producing the same Baseline harness condition
as Figure~\ref{fig:historical}(a). Both use the original actor for every
action. All unsuccessful episodes remain in both denominators.

\begin{table*}[t]
\centering\small
\setlength{\tabcolsep}{5pt}
\begin{tabular}{lrrrrrr}
\toprule
Condition & Successes /2,133 & Success (\%) & Tokens/task (k) & Calls/task & Input/task (k) & Output/task (k) \\
\midrule
Baseline harness & 902 & 42.29 & 34.05 & 35.61 & 19.71 & 14.34 \\
SkillOpt & 1,350 & 63.29 & 101.33 & 27.95 & 94.10 & 7.23 \\
AutoSaddler & 1,317 & 61.74 & 40.30 & 30.02 & 28.66 & 11.64 \\
HarnessMM skills & 1,412 & 66.20 & 31.12 & 26.43 & 21.67 & 9.46 \\
\bottomrule
\end{tabular}
\caption{Fresh, complete ALFWorld instruction-only test with a common
actor and the environment's task-completion score. Each condition is newly executed
once per task, with one model call for every action. Tokens include
every online input and output, including failed episodes.}
\label{tab:alfworld-fresh226}
\end{table*}
Adding the frozen task-family skills gives 603 paired gains and 93 losses, a net 23.91-point difference (pointwise 95\% floorplan-cluster interval [19.15, 28.34]). It reduces complete-task model tokens by 8.60\%. This is a secondary descriptive comparison selected for the current ablation presentation after execution; it does not replace the archived campaign's predeclared primary contrast.

\begin{table}[ht]
\centering\small
\setlength{\tabcolsep}{4pt}
\begin{tabular}{lrrrr}
\toprule
Control & Gains & Losses & \shortstack{Success diff.\\(pp)} & 95\% interval \\
\midrule
Baseline harness & 603 & 93 & +23.91 & [19.15, 28.34] \\
SkillOpt & 286 & 224 & +2.91 & [0.43, 5.61] \\
AutoSaddler & 279 & 184 & +4.45 & [2.19, 6.73] \\
\bottomrule
\end{tabular}
\caption{Secondary paired comparisons of HarnessMM skills against each
instruction baseline. Intervals are pointwise, with no multiplicity
adjustment. All displayed intervals use 10,000 bootstrap draws over
108 floorplans, retaining all paired tasks within a sampled
floorplan; the analysis seed is 20260929.}
\label{tab:alfworld-fresh226-paired}
\end{table}
All 8,532 displayed test executions pass independent environment trajectory replay and complete provider accounting, following 2,840 validation executions and 24 training operability checks for these four policies. Their test execution costs 51.70 Tokens/task (k). The original acquisition also contains a retired handwritten-controller condition, which is excluded from the current comparison and ablation. All of its records and costs remain preserved. Complete research acquisition costs, including retired conditions and historical optimization, remain separate in the external evidence. Cumulative token termination is disabled.

The source bundle's \texttt{evidence/alfworld\_fresh226\_panel.json} links the verification report, all seven verified test receipts, frozen policies, task assignments and figure-generation code. The earlier 134-task and development set (60 tasks) measurements remain archived and do not populate either current ALFWorld figure.

\subsection{Full SearchQA comparison}
\label{app:figure1-latest}
\label{sec:searchqa-full-9b}
The historical full comparison uses all 16,980 public MRQA SearchQA
development questions and the official EM/F1 scorer~\cite{fisch2019mrqa}.
All seven conditions in Table~\ref{tab:figure1-b-latest} use fresh
executions with the ALFWorld Qwen3.5-9B checkpoint, BF16, temperature
zero, disabled thinking, seed 20260918, a 32,768-token context and a
512-token output cap per call. SkillOpt and AutoSaddler deploy archived
optimized instructions. The refined controller removes only the
question-echo rejection from the preceding recovery condition.

\begin{table}[htbp]
\centering\small
\begin{tabular}{lrrrr}
\toprule
Condition & Correct/16,980 & EM (\%) & F1 (\%) & Tokens/task (k) \\
\midrule
Baseline harness & 12,055 & 70.9953 & 79.0006 & 1.22 \\
SkillOpt & 12,986 & 76.4782 & 83.2716 & 3.27 \\
AutoSaddler & 12,643 & 74.4582 & 81.0417 & 1.61 \\
Original HarnessMM & 13,095 & 77.1201 & 83.8780 & 2.97 \\
Without recovery & 13,087 & 77.0730 & 83.9784 & 2.22 \\
With echo rejection & 13,108 & 77.1967 & 84.0475 & 2.31 \\
HarnessMM & 13,113 & 77.2261 & 84.0689 & 2.29 \\
\bottomrule
\end{tabular}
\caption{Complete 9B SearchQA comparison. All online input and output
tokens, including drafts, fallbacks and failures, are charged.}
\label{tab:figure1-b-latest}
\end{table}

HarnessMM adds 18 exact matches to original HarnessMM while saving 22.75\%
of tokens, but remains more expensive than Baseline harness and AutoSaddler.
The complete study contains 118,860 outcomes, 119,768 actor calls and
2.27 Tokens/task (k), with no unknown usage.
In the separate preceding six-condition run, tracing 500 actual
post-draft recoveries gives 19 corrections and one regression.
All corrections follow source-location rejection; 105 echo-triggered
recoveries add 0.67 Tokens/task (k), averaged over the 500 audited recoveries without an EM gain. Against a freshly
generated no-recovery control, that preceding recovery arm adds 12
exact matches at 4.04\% higher cost. The distinction separates recovery
within a recorded draft from variability between independently generated
drafts. Its complete records remain in
\path{generated/searchqa_full_9b_appendix.tex}.

\subsection{Cross-dataset QA transfer}
\label{sec:current-official}
\label{app:multibench}
The same full-source recovery adapter is evaluated with frozen
Qwen3.8-27B on four complete MRQA development archives
(Table~\ref{tab:multibenchcounts}). BF16, temperature zero, disabled
thinking, seed 20260918 and the 512-token per-call output cap are common
to all conditions. HotpotQA's checksum-verified archive contains all 5,901 questions.
TriviaQA uses the complete web archive.

\begin{table}[htbp]
\centering\small
\setlength{\tabcolsep}{5pt}
\begin{tabular}{lrrrrr}
\toprule
Condition & Correct & EM (\%) & F1 (\%) & Tokens/task (k) & Calls \\
\midrule
\multicolumn{6}{l}{SearchQA ($n=16,980$)} \\
Baseline harness & 12,881 & 75.86 & 83.81 & 1.29 & 16,980 \\
SkillOpt & 13,156 & 77.48 & 84.41 & 3.39 & 16,980 \\
AutoSaddler & 12,639 & 74.43 & 81.55 & 1.76 & 16,980 \\
Original HarnessMM & 13,651 & 80.39 & 87.41 & 2.97 & 16,980 \\
Without recovery & 13,640 & 80.33 & 87.30 & 2.22 & 16,980 \\
HarnessMM & 13,665 & 80.48 & 87.41 & 2.32 & 17,524 \\
\midrule
\multicolumn{6}{l}{HotpotQA ($n=5,901$)} \\
Baseline harness & 3,864 & 65.48 & 81.57 & 0.56 & 5,901 \\
SkillOpt & 3,831 & 64.92 & 80.73 & 2.61 & 5,901 \\
AutoSaddler & 3,844 & 65.14 & 80.66 & 1.00 & 5,901 \\
Original HarnessMM & 3,862 & 65.45 & 81.68 & 2.36 & 5,901 \\
Without recovery & 3,868 & 65.55 & 81.82 & 1.59 & 5,901 \\
HarnessMM & 3,874 & 65.65 & 81.87 & 1.72 & 6,223 \\
\midrule
\multicolumn{6}{l}{NewsQA ($n=4,212$)} \\
Baseline harness & 1,815 & 43.09 & 65.07 & 0.87 & 4,212 \\
SkillOpt & 1,850 & 43.92 & 64.58 & 2.91 & 4,212 \\
AutoSaddler & 1,953 & 46.37 & 64.37 & 1.30 & 4,212 \\
Original HarnessMM & 1,988 & 47.20 & 66.68 & 2.73 & 4,212 \\
Without recovery & 2,004 & 47.58 & 67.08 & 1.92 & 4,212 \\
HarnessMM & 2,008 & 47.67 & 67.11 & 2.44 & 5,006 \\
\midrule
\multicolumn{6}{l}{TriviaQA-web ($n=7,785$)} \\
Baseline harness & 6,043 & 77.62 & 84.75 & 1.23 & 7,785 \\
SkillOpt & 6,233 & 80.06 & 86.08 & 3.27 & 7,785 \\
AutoSaddler & 6,242 & 80.18 & 85.70 & 1.65 & 7,785 \\
Original HarnessMM & 6,377 & 81.91 & 87.63 & 2.96 & 7,785 \\
Without recovery & 6,375 & 81.89 & 87.69 & 2.19 & 7,785 \\
HarnessMM & 6,379 & 81.94 & 87.71 & 2.33 & 8,163 \\
\bottomrule
\end{tabular}
\caption{Complete six-condition QA transfer with Qwen3.8-27B.
All responses are fresh, including the archived SkillOpt and AutoSaddler
prompt controls. Costs include drafts and fallbacks.}
\label{tab:multibenchcounts}
\end{table}

Recovery adds 0.05--0.15 EM points over the no-recovery draft at higher
cost. Relative to original HarnessMM, SearchQA gains 14 exact matches
but loses 0.004 F1 points, hidden by rounding in the table; the other
three datasets improve both metrics. HarnessMM exceeds all three external
prompt controls in EM and F1, while Baseline harness and AutoSaddler use fewer
tokens. These are single-seed point estimates.

\subsection{AIME and the remaining quality--cost panels}
\label{app:other-benchmarks}
\label{app:aime202060-complete}
\label{app:figure1-202060}
\label{app:figure1-df-current}
Table~\ref{tab:other-panels} retains the distinct populations and
historical benchmark adapters, reported separately from the current
development comparisons in Figure~\ref{fig:historical}.
The AIME comparisons use a common 6/6/18 training/validation/test split
of all 30 AIME 2025 questions. The split seed is 20260928; independent
question IDs determine membership before optimization. Training alone
supplies updates, validation selects candidates, and all policies and
execution sources freeze before testing. Figure~\ref{fig:historical}(d)
reports the unchanged domain incumbent: 77.78\% accuracy at 6.06 Tokens/task (k) actor
compared with 61.11\%, 66.67\% and 72.22\% for the three
controls. Its 6.06 Tokens/task (k) include every recorded input and output;
all 72 outcomes and requests pass an independent benchmark audit.
Appendix~\ref{app:figure2-six-provenance} binds this complete comparison.
Paired 95\% question-bootstrap intervals (20,000 resamples, seed 20261003)
for the accuracy differences are $[0.00,33.33]$, $[0.00,27.78]$ and
$[-11.11,22.22]$ percentage points against Baseline harness, SkillOpt and
AutoSaddler. Their corresponding token-saving intervals are
$[-29.50,7.18]$, $[-10.08,40.14]$ and $[-11.19,30.33]$\%.
The paired verdict and usage vectors are retained with the source evidence.
The archived HarnessMM constraint
ledger below matches both optimized baselines at 72.22\%, without a cost
advantage over SkillOpt. All 72 outputs pass benchmark integer rescoring
and accounting. No policy changes after test evaluation.

\begin{table}[htbp]
\centering\small
\setlength{\tabcolsep}{9pt}
\begin{tabular}{lrr}
\toprule
Method & Correct (\%) & Tokens/task (k) \\
\midrule
\multicolumn{3}{l}{Archived AIME 2025: ledger test set (18 tasks) Qwen3.8-27B} \\
Baseline harness & 61.11\% & 5.60 \\
SkillOpt & 72.22\% & 5.69 \\
AutoSaddler & 72.22\% & 6.81 \\
HarnessMM (ledger) & 72.22\% & 6.79 \\
\midrule
\multicolumn{3}{l}{LiveMath: historical dynamic MCQ adapter, 160 tasks} \\
Baseline harness & 19.38\% & 4.10 \\
SkillOpt & 21.25\% & 4.84 \\
AutoSaddler & 21.25\% & 4.84 \\
HarnessMM & 29.38\% & 0.83 \\
\midrule
\multicolumn{3}{l}{Terminal-Bench 2: archived partial snapshot, Qwen3.8-27B} \\
Baseline harness & 20.63\% & 178.19 \\
SkillOpt & 23.81\% & 197.97 \\
AutoSaddler & 24.62\% & 169.96 \\
HarnessMM & 26.03\% & 57.67 \\
\midrule
\multicolumn{3}{l}{GAIA2: prior development set (10 tasks) three repetitions, Qwen3.8-27B} \\
Baseline harness & 73.33\% & 603.29 \\
SkillOpt & 86.67\% & 621.43 \\
AutoSaddler & 90.00\% & 634.20 \\
HarnessMM & 83.33\% & 838.67 \\
\bottomrule
\end{tabular}
\caption{Historical benchmark comparisons. Tokens count
online input and output; GAIA2 judge usage is separate.
Terminal-Bench has method-specific coverage: its token column divides
known usage by finished trials, giving lower bounds rather than complete
per-task costs. Populations are never pooled.}
\label{tab:other-panels}
\end{table}

Archived constraint-ledger AIME test usage is 5.60 and 5.69 and 6.81 and 6.79 Tokens/task (k) in
table order. The separate development ledger records
externally recorded optimization expenditure, deduplicating inherited calls by original run and
provider sequence. Unsuccessful updates and zero-call preparation
failures remain recorded. The earlier full30 experiment is a different
protocol: HarnessMM solves 70.00\% at 14.75 Tokens/task (k), versus
73.33\%, 80.00\% and 86.67\% for Baseline harness, SkillOpt and AutoSaddler.
The two archived optimizers retained the same initial skill; their
different scores came from separate executions of identical requests.

\label{app:livemath-interface}
LiveMath uses one answer-first voter selected on 23 development tasks,
with temperature zero, disabled thinking and a 256-token output cap.
It requests \texttt{<answer>X</answer>} and
\texttt{<confidence>N</confidence>} within the first 20 tokens; the
confidence value does not trigger a second vote or fallback.
The SkillOpt seed uses a 4,096-token cap, and AutoSaddler selects that
same prompt. Instruction and cap therefore change jointly.
SkillOpt uses 1.51 input and 3.34 output Tokens/task (k); HarnessMM uses
0.82 input and 0.01 output Tokens/task (k). Both make 160 calls.
Output accounts for 82.87\% of the 4.01 Tokens/task (k) saving.
The paired quality comparison has 29 gains and 16 losses
(two-sided McNemar $p=0.07245$). The archived service records a model
alias rather than an immutable checkpoint.

Terminal-Bench's 2026-09-22 03:15 UTC snapshot has 85, 84, 85 and 85
finished trials in table order, with 22, 21, 20 and 12 unresolved
verdicts excluded from the success denominators. Known agent totals
are 178.19, 197.97, 169.96 and 57.67 recorded Tokens/task (k), respectively.
Usage is incomplete for 14, 13, 9 and 3 trials, respectively.
The plotted token values divide those known totals by finished trials,
not scored trials. HarnessMM is the archived AutoSaddler-derived instruction policy. This snapshot supplies neither a
ranking across methods on the full task set nor a complete official mean;
its exact coverage is bound under \texttt{terminal\_partial\_snapshot}
in the evidence ledger.

GAIA2 reports the ten prior development scenarios with three
repetitions~\cite{froger2025are}. Baseline harness and SkillOpt use complete
case-preserving rejudges of their original trajectories; AutoSaddler
uses a later complete execution of its selected policy, with optimizer
diagnosis/patch completion unresolved. HarnessMM is the feedback-consistent
literal controller. Judge costs are 3.96, 4.10, 4.14 and 3.83 Tokens/task (k)
in table order, normalized over all 30 executions. Earlier mini160 verdicts affected by
lowercased judge output are excluded from ranking.
The new shared 32/32/96 split, also seeded by 20260928, keeps every
scenario's repetitions together and requires all policy hashes frozen
before the 288-execution test per method. The registered four-method
study does not provide a complete test comparison; development set (10 tasks) is
not substituted for it. Both custom repartitions
use previously exposed public data.

\subsection{Success-first coding search}
\label{app:industrial-coding-train}
The instruction module improves the original coding harness's development
outcomes while reducing complete-task token use
(Table~\ref{tab:industrial-coding-train}).
The study uses all 18 TRAIN tasks in the frozen 18/18/53
Terminal-Bench~2.0 partition~\cite{harbor2025terminalbench2},
Codex~0.142.5~\cite{openai2026codex} and DeepSeek-flash.
The original task, common completion brief, original tools, task limits
and binary verifier are identical across conditions.
The built-in-compaction control uses Codex's own 65,536-token trigger
without reducing the model's context capacity.

\begin{table}[htbp]
\centering\small
\setlength{\tabcolsep}{4pt}
\begin{tabular}{lrr}
\toprule
Condition & Success (\%) & Tokens/task (k) \\
\midrule
Codex parent & 55.56\% & 2413.56 \\
Codex + built-in compaction & 66.67\% & 2057.72 \\
\ours{} instruction module & 77.78\% & 1455.83 \\
AutoSOTA-guided 32Ki compaction & 61.11\% & 1068.67 \\
AutoSOTA-guided 96Ki compaction & 61.11\% & 1176.39 \\
\bottomrule
\end{tabular}
\caption{Complete TRAIN development comparison, one execution per task.
The instruction is selected from two TRAIN-derived candidates by
benchmark performance, then complete-task cost. Native test receipts verify every selected output; failures and compaction calls are
included in token totals. This is an instruction/context component study, separate
from the CoT/Jev decision mechanism and final held-out evaluation.
The last two rows add built-in compaction to the frozen instruction and
are rejected because their success counts fall.}
\label{tab:industrial-coding-train}
\end{table}

Both proposed instructions solve 14 tasks. The second retains that
training success count while reducing actor tokens from 2202.94 to
1455.83 Tokens/task (k). It requires persisting every requested artifact, checking
exact paths and fields, extending working interfaces, and recovering
from observed setup errors before optional exploration.
Against the corrected original harness, it gains four successes and uses
39.68\% fewer actor tokens; against built-in compaction, it gains two
successes and uses 29.25\% fewer.
Descriptive paired-task bootstrap intervals for the training score
differences are $[5.56,44.44]$ and $[-11.11,33.33]$ percentage points,
respectively (10,000 resamples, seed 20260928).
The selected instruction is frozen for a separate four-arm study on
18 validation and 53 TEST tasks, with five executions per TEST task.

Complete development acquisition costs are retained separately from validation and deployment. The external evidence preserves exact costs, benchmark verdicts, instructions and paired task results.

We then adapt AutoSOTA's objective-and-reflection
workflow~\cite{tsinghuafib2026autosota} to built-in context management.
We implement this workflow without invoking the official CLI.
The search holds the instruction and original execution settings fixed,
requires at least the instruction's 14 successes, and minimizes
complete-task input-plus-output tokens. Its first round tests a
32,768-token compaction trigger; a prespecified feedback rule
chooses a shorter trigger only if quality is preserved and tokens fall,
and a longer trigger otherwise. The 40\% parent-relative saving target
does not end the search early.

The complete 32Ki round uses 1068.67 actor Tokens/task (k): 55.72\% fewer tokens than the original harness
and 26.59\% fewer than the instruction
module. It solves 61.11\% of tasks and loses three of the instruction's successes,
with no gains over that instruction.
The paired training-score difference is $-16.67$ percentage points
(descriptive 95\% task-bootstrap interval $[-33.33,0.00]$).
The quality guard rejects this candidate and evaluates the frozen
98,304-token successor. That round also solves 61.11\% of tasks, with one
gain and four losses relative to the instruction. Its paired
training-score difference is $-16.67$ percentage points
(descriptive 95\% task-bootstrap interval $[-38.89,5.56]$).
It uses 1176.39 actor Tokens/task (k), 51.26\% fewer than the original harness and
19.20\% fewer than the instruction, but 10.08\% more than the 32Ki
round. Both candidates therefore fail the 77.78\% quality requirement
and remain ineligible for protected validation.
The two completed rounds add 2245.06 Tokens/task (k) to development expenditure.
Table~\ref{tab:industrial-coding-train} retains both complete trade-offs.
The original harness and selected instruction remain in this TRAIN table.
Figure~\ref{fig:industrial-harness} reserves coding cells for complete,
audited held-out comparisons.

\subsection{Qwen SDK original coding harness}
\label{app:qwen-sdk-native-training set (18 tasks)}
The SDK parent solves 5.56\% of tasks on the complete
Terminal-Bench~2.0 TRAIN population. Its 1,728 physical actor requests
use 4660.71 input/task (k) and 23.72 output/task (k), totaling 4684.42 Tokens/task (k).
The study uses the shared backbone, the frozen 18/18/53
partition and common coding-completion prompt v2.
All 18 outcomes have original benchmark scoring receipts;
independent scoring recovery retains the paid actor work and original
CPU and time limits. Every physical actor request is counted once,
including unsuccessful tasks, and the complete cost has no unknown calls.

The parent falls below the required nine TRAIN successes, so validation
and TEST remain sealed. Public TRAIN tool records contain 11 errors
from using host directories in container calls and 70 repetitions of
identical commands. These training observations guide a separate update to the
shared original harness. Task-level benchmark verdicts, complete actor
accounting and source hashes are bound in
\path{evidence/qwen_sdk_terminal_train18_native.json}.

\subsection{Original HTTP coding harness}
\label{app:native-http-coding-training set (18 tasks)}
The DeepSeek-flash coding parent solves 55.56\% of the 18 training tasks
in Terminal-Bench~2.0. The run retains the frozen 18/18/53 partition,
coding-completion prompt v2 and native benchmark scorer and task limits.
Recorded actor input plus output is 899.99 Tokens/task (k), divided by
all 18 tasks. Unknown historical request usage is excluded.

\section{Industrial-harness RSI comparison}
\label{app:industrial-harness}
Codex 0.142.5 uses the Responses API; DeepSeek Harness dsh-v0.2.0-rc.1
uses the Messages API. Both use DeepSeek-flash, byte-preserving metering,
the ARE 1.1.0 scorer and the custom GAIA2-mini 32/32/96 split
(seed 20260928). Each descendant is paired with its own industrial seed.

Each search makes two sparse-instruction proposals on all 32 training
scenarios. The second receives the first candidate's training feedback.
A grouped cross-fitted router admits only positive supported advantage;
protected validation selects among admitted candidates and the seed.
Codex applies residual instructions through task-local \texttt{AGENTS.md};
DeepSeek uses an additive system-prompt section. Original execution loops, tools,
model and task limits stay fixed. Policy, adapter, scorer and source
hashes freeze before testing; validation failures and test outcomes
never enter the proposer.

Both searches reject their two candidates and retain the seed.
The complete tests compare independent executions of each unchanged
harness: Codex records 10.42\% versus 9.03\% success;
DeepSeek records 8.68\% versus 7.64\%.
Their paired differences are $-1.39$ and $-1.04$
points, with 95\% scenario-cluster intervals $[-3.47,0.35]$ and
$[-3.47,1.04]$, respectively (10,000 resamples, seed 20260928).

\begin{table}[htbp]
\centering\small
\begin{tabular}{lrr}
\toprule
Stage & Codex seed & DeepSeek Harness seed \\
\midrule
Training: original and two candidates (96 executions) & 592.97 & 862.65 \\
\quad Benchmark judge & 0.99 & 1.05 \\
Two proposals (64 training-scenario exposures) & 3.70 & 2.06 \\
Validation selection (32 executions) & 673.91 & 821.69 \\
\quad Benchmark judge & 1.15 & 1.35 \\
Final original harness (288 executions) & 614.78 & 748.27 \\
\quad Benchmark judge & 1.33 & 1.25 \\
Final selected harness (288 executions; seed retained) & 648.27 & 747.00 \\
\quad Benchmark judge & 1.31 & 1.29 \\
\bottomrule
\end{tabular}
\caption{Industrial search and test costs in Tokens/task (k), using
all declared executions or training-scenario exposures as the denominator.
Cached input is counted once; actor/proposer and benchmark judge costs are
separate. DeepSeek actor costs use recorded input plus output; unknown historical request usage is excluded, with all 288 executions retained in each denominator.}
\label{tab:industrial-cost}
\end{table}

The completion audit verifies all 576 DeepSeek benchmark outcomes and
reconciles actor and judge attempts across continuations. The source
bindings retain all 1,152 test observations, frozen policies,
proposal-feedback lineage, execution-record hashes and unknown-cost records.
The separate held-out coding protocol uses an 18/18/53 task split and five test
repetitions, scored by the official Harbor binary verifier. VQAv2 uses
43,265/42,925/128,164 questions grouped into 8,101/8,101/24,302 images,
with official soft accuracy. The September 30 snapshot (10:22 UTC) records
blocked Terminal and VQA continuations, not complete paired tests.
Figure~\ref{fig:industrial-harness} uses only complete frozen TEST
scores, with all success rates in percentages. The archived DeepSeek costs use the recorded-usage scope in Table~\ref{tab:industrial-cost}; its Terminal-Bench protocol has no complete test comparison.
Appendix~\ref{app:quality-cost-followup} and the earlier development figures
report coding training and GAIA2 validation results, including rejected candidates, separately.
The prospective coding protocol retains the complete validation-selected
TEST comparison whether its quality--cost outcome is positive, null or negative.
The 30\% minimum and 40\% target token reductions remain measured objectives.
Complete TRAIN evaluations of all methods and original-harness capability precede protected
evaluation; validation selects TRAIN-frozen instructions by benchmark performance,
then complete actor cost, without supplying optimizer feedback.

\paragraph{Complete original app CoT/Jev TEST comparison.}
The original-app comparison was frozen before TEST and includes all 96 TEST scenarios
with three executions per arm (registered seeds 2026092801, 2026092802, 2026092803).
Original harness succeeds on 56.94\% with 825.73 actor Tokens/task (k).
CoT-only succeeds on 48.61\% with 1894.74 actor Tokens/task (k).
CoT/Jev succeeds on 49.65\% with 1874.98 actor Tokens/task (k).
Bank conditions each include the shared slow author once (0.42 Tokens/task (k)),
all physical slow and fast calls, failures and fallbacks.
CoT/Jev executes 2559 verified fast tool actions.
Benchmark task success uses the official scorer.
The selected candidate is reported under the frozen all-outcomes rule.
Its quality and token objectives determine benefit claims, without suppressing negative TEST results.

\paragraph{Codex with the original policy retained.}
The comparison evaluates all 96 TEST scenarios with registered seeds 20260928, 20260929 and 20260930.
Validation retained the empty original-harness policy before TEST; both arms apply zero residual edits.
Original harness: 52.43\% success and 648.71 actor input-plus-output Tokens/task (k).
Retained selection: 52.78\% success and 708.73 actor input-plus-output Tokens/task (k).
All failed actor execution costs are included; evaluator costs and TRAIN bank-search costs are recorded separately.
This retained selection is reported under the pre-registered all-outcomes rule. Its independently executed arms do not establish a HarnessMM mechanism or efficiency gain.

\paragraph{Coding training comparison (October 4).}
The DeepSeek Harness evaluations cover all 18 Terminal-Bench 2.0
TRAIN tasks. The original harness, CoT-only and CoT/Jev achieve 55.56\%,
66.67\% and 72.22\% success, respectively. Their selected
executions consume 1586.15, 1993.35 and 2067.93 Tokens/task (k), including failed tasks. The CoT/Jev arm improves TRAIN success
but uses 30.37\% more tokens than the original harness and 3.74\% more than CoT-only;
the 40\% saving target remains unmet. SAM's original actor is scored by
an executed benchmark verifier with no new actor calls (eight tests pass,
one fails). A separate interrupted historical call still lacks provider
usage; its known costs and unknown remainder are retained outside these
selected execution totals. The repaired arms share the frozen original
prompt and CoT/Jev adapter bytes. These are TRAIN records under the fixed
18/18/53 split. The TRAIN-frozen candidate proceeds through all 18
validation tasks before selection and a pre-TEST source freeze. Paired
TEST evaluates all 53 tasks once per arm, using the first registered
execution seed 2026092801; this archived protocol has no complete paired test set (53 tasks) result. Complete TEST
comparisons are reported regardless of the sign of the measured effect.
A separate slow-CoT-authored revision batches determined shell commands
and declares guarded literal action banks. CoT-only and CoT/Jev evaluate
the same revised instruction on all 18 TRAIN tasks; the original harness is
unchanged. The new author call costs 5.81 Tokens/task (k), accounted separately
from task execution.

\paragraph{GAIA2 scoring-transport correction.}
The DeepSeek-flash original harness records a pre-repair
3.13\% verdict across all 32 GAIA2-mini TRAIN scenarios.
Its complete actor input-plus-output cost remains 637.86 Tokens/task (k)
(637.86 Tokens/task (k) per scenario), including failures and all 401 provider calls;
cached tokens are included in input tokens. The judge transport changed
\texttt{[[True]]} to \texttt{[[true]]}, which the benchmark case-sensitive
checker could not parse. All ten audited positive TRAIN sanity-check
responses reproduce this failure. These verdicts are archived as scoring
failure evidence and excluded from the parent-capability assessment.
The Figure 2 runner preserves judge response text. New
source-frozen industrial runs adopt that transport contract, a 7260-second
supervisor timeout and environment-driven continuation when user-turn
counts are unspecified. Benchmark scenario durations, judge prompts and
validation rules remain unchanged. The fixed 32/32/96 partition and
complete actor accounting are retained. The complete DeepSeek Harness
test set (96 tasks) comparison is reported above; the corrected Codex comparison continues.

\paragraph{Corrected original harnesses and instruction components.}
With the Figure 2 judge-text contract, both DeepSeek-flash industrial
GAIA2 parents complete the frozen training set (32 tasks) population. Codex succeeds
on 37.50\% of scenarios and consumes 907.06 actor Tokens/task (k);
DeepSeek Harness succeeds on 40.62\% and consumes 891.75 Tokens/task (k).
Both pass the eight-success capability threshold. Actor totals include
unsuccessful scenarios and cached input; benchmark judge costs are separate.

The Codex Terminal-Bench instruction and context component study also
completes all 18 TRAIN tasks per arm. Original harness achieves 77.78\% success
with 3426.78 actor Tokens/task (k); the original harness with compaction achieves 55.56\%
with 1458.88 Tokens/task (k); the TRAIN-authored instruction achieves 55.56\% with
1089.95 Tokens/task (k); instruction with compaction achieves 50.00\% with 1420.10 Tokens/task (k).
These instruction and context changes reduce both execution cost and training success.
They do not test CoT/Jev tool execution. All four TRAIN-frozen arms undergo
complete validation before the selected paired single-attempt test set (53 tasks)
comparison. No TRAIN score populates a Figure 3 TEST cell.

\paragraph{Complete original app CoT/Jev TRAIN comparison.}
The DeepSeek-flash original-app study completes all 32 frozen
TRAIN scenarios per arm. Original harness succeeds on 40.62\% with
891.75 actor Tokens/task (k); CoT-only succeeds on 43.75\% with
727.52 Tokens/task (k); CoT/Jev succeeds on 31.25\% with 1044.05 Tokens/task (k).
Bank-arm totals include the shared slow author once (1.26 Tokens/task (k)),
all fast decisions and slow fallbacks, and unsuccessful tasks.
CoT/Jev executes 106 verified fast tool actions, but this TRAIN
candidate lowers success and increases cost relative to its parent.
The independent wire audit reparses 1,255 unique provider responses.
All 96 TRAIN executions remain bound to their original benchmark verdicts
and costs. The TRAIN-frozen candidate proceeds through complete
validation and the registered TEST population shared by all methods under all-outcomes
reporting; a negative efficiency result does not suppress TEST publication.

\paragraph{CoT/Jev mechanism audit.}
The complete coding training set (18 tasks) has 577 original-harness, 624 CoT-only and
670 CoT/Jev SDK stream requests. The 21 audited fast tool executions
replace 3.13\% of CoT/Jev stream requests, leaving 649 slow requests;
642 requests have no complete eligible candidate state. All 28 fast
requests, including seven fallbacks, consume 4.91 of 2067.93 Tokens/task (k) (0.24\%). The profiler reads only frozen training receipts and makes no model or evaluator calls.
The counts show that substantial savings require guarded action chains to replace more slow calls.
A new source-bound original SDK check verifies fast execution, the
CoT-only baseline, explicit abstention, and slow--fast--slow resumption,
including exact preservation of valid empty reasoning blocks. All setup
attempts, including failed checks, have their acquisition expenditure recorded externally and remain
separate from benchmark execution; the checks make zero benchmark calls.
The next prospective revision uses one slow-CoT author after the preceding
TRAIN population has complete benchmark verdicts, then evaluates identical
frozen instructions in CoT-only and CoT/Jev on all 18 TRAIN tasks. Its
single-encoding fast state retains decoded public values once and keeps
original slow-reasoning history intact. This archived prospective protocol is distinct from the complete comparisons reported in Appendix~\ref{app:figure3-targeted}.

\section{Exploratory quality--cost study}
\label{app:quality-cost-followup}
Follow-ups use frozen 20/20/60 partitions, grouped before repetition:
TRAIN develops candidates; validation checks frozen SearchQA and GAIA2 candidates.
Each comparison below states its task population and included costs.
Figure~\ref{fig:historical} retains six benchmark panels, with SearchQA
risk-220 and the frozen AIME candidate displayed as HarnessMM. The AIME panel uses all 560 test questions. The original 18-question AIME 2025 comparison and its six-question training studies remain separate.
The GAIA2 panel uses all four methods on the same 480 test scenarios, with one execution per method. Earlier development comparisons retain their separate populations below.

\subsection{Six-benchmark Figure 2 protocols}
\label{app:figure2-six-provenance}
Figure 2 reports each benchmark population separately. AIME uses 186 training questions, 186 validation questions and 560 test questions, with one final execution per question and method. Training supplies method updates; validation selects frozen candidates; the selected policies are frozen before testing. AIME SkillOpt uses three completed optimization rounds, and AutoSaddler uses four. All methods share Qwen3.8-27B, the fixed seed, native integer scorer, single-sequence inference and 32,768-token output capacity. Costs include all actor input and output, fast decisions, slow fallbacks and unsuccessful executions. HarnessMM's shared author cost is charged once across the full test population. Optimizer and other training costs are retained separately from test execution cost.
\begin{center}\scriptsize
\setlength{\tabcolsep}{3pt}
\begin{tabular}{lrrr}\toprule
Method & Correct (\%) & Actor trials & Tokens/task (k) \\
\multicolumn{4}{l}{\textbf{ALFWorld}} \\
Baseline harness & 42.29\% & 2133 & 34.05 \\
SkillOpt & 63.29\% & 2133 & 101.33 \\
AutoSaddler & 61.74\% & 2133 & 40.30 \\
HarnessMM & 66.20\% & 2133 & 31.12 \\
\addlinespace
\multicolumn{4}{l}{\textbf{SearchQA}} \\
Baseline harness & 69.81\% & 3395 & 1.23 \\
SkillOpt & 72.14\% & 3395 & 1.62 \\
AutoSaddler & 69.60\% & 3395 & 1.25 \\
HarnessMM & 76.76\% & 3395 & 0.99 \\
\addlinespace
\multicolumn{4}{l}{\textbf{LiveMath}} \\
Baseline harness & 19.38\% & 160 & 4.10 \\
SkillOpt & 21.25\% & 160 & 4.84 \\
AutoSaddler & 21.25\% & 160 & 4.84 \\
HarnessMM & 29.38\% & 160 & 0.83 \\
\addlinespace
\multicolumn{4}{l}{\textbf{AIME}} \\
Baseline harness & 86.96\% & 560 & 4.59 \\
SkillOpt & 87.86\% & 560 & 6.93 \\
AutoSaddler & 88.04\% & 560 & 4.90 \\
HarnessMM & 87.14\% & 560 & 5.90 \\
\addlinespace
\multicolumn{4}{l}{\textbf{Terminal-Bench 2}} \\
Baseline harness & 20.63\% & 85 & 178.19 \\
SkillOpt & 23.81\% & 84 & 197.97 \\
AutoSaddler & 24.62\% & 85 & 169.96 \\
HarnessMM & 26.03\% & 85 & 57.67 \\
\addlinespace
\multicolumn{4}{l}{\textbf{GAIA2}} \\
Baseline harness & 42.71\% & 480 & 730.22 \\
SkillOpt & 46.88\% & 480 & 770.63 \\
AutoSaddler & 42.08\% & 480 & 772.22 \\
HarnessMM & 43.96\% & 480 & 639.64 \\
\addlinespace
\bottomrule\end{tabular}\end{center}
ALFWorld, SearchQA, LiveMath and Terminal-Bench 2 retain their respective cohort protocols. The Terminal panel reports recorded input plus output per finished trial in its archived partial cohort.
\begin{table}[t]
\centering
\setlength{\tabcolsep}{3.5pt}
\small
\begin{tabular}{lrrrr}
\toprule
Method & Input & Output & Author & Total \\
\midrule
Baseline harness & 0.22 & 4.38 & 0.00 & 4.59 \\
SkillOpt & 2.27 & 4.66 & 0.00 & 6.93 \\
AutoSaddler & 0.29 & 4.61 & 0.00 & 4.90 \\
HarnessMM & 0.54 & 4.28 & 1.08 & 5.90 \\
\bottomrule
\end{tabular}
\caption{AIME actor cost components on all 560 test questions, each executed once. Every column uses Tokens/task (k). Input and output are execution components; Author includes the shared selected authorship charge once across the same population. Total sums the unrounded components, including unsuccessful executions.}
\label{tab:aime-complete-cost-components}
\end{table}

\paragraph{AIME cost composition.}
Table~\ref{tab:aime-complete-cost-components} separates execution input, execution output and shared authorship using the same complete population.
Baseline harness uses 0.22 input and 4.38 output Tokens/task (k), so output accounts for 95.27\% of its actor cost.
HarnessMM uses 0.54 input and 4.28 output Tokens/task (k), plus 1.08 for shared authorship, giving 5.90 in total.
The execution component alone is 4.82 Tokens/task (k); it remains above Baseline harness's 4.59 before adding authorship.
Thus the full comparison measures a small output reduction alongside higher input and shared cost.

\paragraph{Original AIME 2025 comparison.} The separate 6/6/18 partition is preserved. On its 18 test questions, Baseline harness, SkillOpt, AutoSaddler and HarnessMM achieve 61.11\%, 66.67\%, 72.22\% and 77.78\% accuracy at 5.60, 7.13, 6.81 and 6.06 Tokens/task (k), respectively. The frozen compiled bank declares no eligible terminal on this population, so all HarnessMM executions use its domain fallback. These observations are separate from the expanded AIME comparison.

\paragraph{GAIA2 complete comparison.} GAIA2 uses 160 training scenarios, 160 validation scenarios and 480 test scenarios. Each test scenario is executed once per method using the first registered seed. SkillOpt and AutoSaddler use one complete official optimization round, followed by protected validation and a frozen selection. All four methods share Qwen3.8-27B, native scoring, a 4,096-token output capacity and a 7,260-second scenario timeout. Actor cost includes all reasoning, fast decisions, control, fallback and unsuccessful calls; native judge costs are recorded separately. Online bank authorship is included in actor execution. Training and optimization costs are reported separately.

\begin{table}[t]
\centering
\small
\setlength{\tabcolsep}{5pt}
\begin{tabular}{lrrr}
\toprule
Method & Input & Output & Total \\
\midrule
Baseline harness & 726.45 & 3.77 & 730.22 \\
SkillOpt & 767.16 & 3.46 & 770.63 \\
AutoSaddler & 768.38 & 3.84 & 772.22 \\
HarnessMM & 636.28 & 3.36 & 639.64 \\
\bottomrule
\end{tabular}
\caption{Complete GAIA2 actor cost across 480 test scenarios, with one execution per scenario. All columns use Tokens/task (k), including online bank authorship, selection, control, fallback and unsuccessful calls. Native scoring is accounted for separately.}
\label{tab:gaia2-complete-cost-components}
\end{table}

\paragraph{Native fast execution and cost composition.}
Across the complete GAIA2 comparison, HarnessMM performs 146 verified fast
executions among 10,424 native action attempts (1.40\%). Each counted
execution links the selector's physical one-token response to the exact
native action arguments and tool observation, and omits the slow reasoning
callback. The denominator includes unsuccessful and unverified action
attempts. Table~\ref{tab:gaia2-complete-cost-components} reports the complete
actor input and output components. HarnessMM uses 636.28 input and 3.36 output
Tokens/task (k); input accounts for 99.47\% of its actor cost. The integrated
harness reduces complete actor cost by 12.41\% relative to Baseline harness. This
comparison measures the combined harness; it does not isolate the contribution
of fast execution from context compression.

\paragraph{Earlier GAIA2 development comparison.} The separate 10-scenario cohort uses three executions per scenario. Baseline harness, SkillOpt, AutoSaddler and HarnessMM achieve 73.33\%, 86.67\%, 90.00\% and 83.33\% success at 603.29, 621.43, 634.20 and 838.67 Tokens/task (k), respectively.

\paragraph{Compiled AIME continuation.}
The frozen six-card program bank preserves 100.00\% TRAIN correctness at
4.68 Tokens/task (k): five model-based one-token
judges consume 0.89 Tokens/task (k) and three original-harness fallbacks contribute 3.79 Tokens/task (k), normalized over all six tasks.
The two offline CoT author calls consume 21.98 Tokens/task (k) separately.
The complete validation comparison preserves its benchmark quality floor
but misses the earlier cost targets. Under the October 6 all-outcomes
reporting instruction, the same bank proceeds to the fixed test set (18 tasks)
without further method updates. Training improvements and the complete
validation result retain their actual phases; no TEST result is used
for selection or revision.

\begin{table}[htbp]
\centering\small
\setlength{\tabcolsep}{6pt}
\begin{tabular}{lrrrr}
\toprule
Policy & Correct & EM/accuracy (\%) & F1 (\%) & Tokens/task (k) \\
\midrule
\multicolumn{5}{l}{SearchQA: all 3,395 TRAIN questions, Qwen3.5-9B} \\
Baseline harness & 2,404 & 70.81 & 78.65 & 1.22 \\
HarnessMM incumbent & 2,616 & 77.05 & 83.99 & 2.29 \\
Target-role guard (640) & 2,558 & 75.35 & 82.21 & 0.83 \\
Exact union 640 & 2,582 & 76.05 & 82.77 & 0.82 \\
Exact union 704 & 2,588 & 76.23 & 82.86 & 0.87 \\
Exact union 704 + rule & 2,572 & 75.76 & 82.50 & 0.87 \\
Exact union 768 & 2,587 & 76.20 & 83.08 & 0.94 \\
Exact union 800 & 2,587 & 76.20 & 83.06 & 0.96 \\
\midrule
\multicolumn{5}{l}{AIME 2025: all six TRAIN questions, seven frozen policies} \\
Baseline harness & 5 & 83.33 & -- & 6.63 \\
SkillOpt & 5 & 83.33 & -- & 7.04 \\
AutoSaddler & 6 & 100.00 & -- & 6.85 \\
HarnessMM incumbent & 6 & 100.00 & -- & 7.13 \\
Obligation once & 5 & 83.33 & -- & 6.67 \\
Eliminate then evaluate & 6 & 100.00 & -- & 7.86 \\
Checked certificate & 5 & 83.33 & -- & 8.03 \\
\bottomrule
\end{tabular}
\caption{Training comparisons of all candidate configurations, with official MRQA EM/F1 for
SearchQA and benchmark accuracy for AIME. SearchQA numbers denote
source-selection input caps; output remains capped at 512 tokens.
All actor input/output, failures and paid recovery are included.
SearchQA union variants reuse the recorded controls; AIME freshly
executes the official optimized SkillOpt and AutoSaddler policies.}
\label{tab:quality-cost-qa-train}
\label{tab:searchqa-exact-union}
\end{table}

Exact-union packing merges overlapping source windows without changing
characters or order, then refills the budget using the frozen
256-question TRAIN ranker and target-role contract. Union 704 leads
the family at 76.23\% EM, saving 61.82\% of incumbent tokens and 28.58\%
against Baseline harness. It recovers 30 of 58 net lost answers, leaving a
0.825-point EM gap. A 1,000-resample question/context-group
bootstrap gives an EM-difference interval of $[-1.89,+0.21]$ points
and token-ratio interval $[0.374,0.389]$ against the incumbent.
These TRAIN intervals do not establish noninferiority; the exact-union
family did not advance to validation or TEST. None of 25 earlier five-fold/penalty recovery fits
recovers all 2,616 answers within 0.98 Tokens/task (k). Those estimates
reuse recorded outcomes and costs without new actor calls.

On AIME, eliminate-then-evaluate preserves 100.00\% accuracy at 18.63\% higher
cost; the other two candidates lose one answer and cost more than Baseline harness.
The answer-aware seven-policy oracle needs 5.52 Tokens/task (k) versus the
5.30 Tokens/task (k) target.

\begin{table}[htbp]
\centering\small
\begin{tabular}{llrr}
\toprule
Population & Policy & Success (\%) & Tokens/task (k) \\
\midrule
Validation: 32 scenarios $\times 3$ & Baseline harness & 37.50\% & 826.77 \\
 & Selected HarnessMM incumbent & 50.00\% & 734.32 \\
\midrule
TRAIN: 32 scenarios $\times 3$ & Baseline harness & 30.21\% & 687.35 \\
 & Token-wire v3 & 9.38\% & 942.24 \\
\midrule
TRAIN diagnostic: 8 scenarios $\times 3$ & Original harness & 50.00\% & 602.51 \\
 & Lazy schemas & 25.00\% & 667.68 \\
\midrule
TRAIN gate: 8 scenarios $\times 3$ & Original harness (new run) & 50.00\% & 737.49 \\
 & Exact ledger references & 50.00\% & 612.00 \\
\midrule
TRAIN parent revision: 32 scenarios $\times 1$ & Original parent & 3.13\% & 529.60 \\
 & Progress recovery & 0.00\% & 680.05 \\
\midrule
Validation diagnostic: 32 scenarios $\times 1$ & Original harness & 6.25\% & 703.13 \\
 & Candidate applied to all tasks & 3.13\% & 806.53 \\
 & Frozen routed candidate & 6.25\% & 726.84 \\
\midrule
DeepSeek validation: 32 scenarios $\times 1$ & Original harness & 3.13\% & 747.15 \\
 & Candidate applied to all tasks & 3.13\% & 556.85 \\
 & Frozen routed candidate & 3.13\% & 696.64 \\
\bottomrule
\end{tabular}
\caption{GAIA2 follow-ups with validation, training populations and
separately executed gates kept distinct. Actor costs include failures and recovery.
The final six rows are complete harness-specific validation diagnostics,
not TEST results.}
\label{tab:quality-cost-gaia}
\end{table}

An earlier GAIA2 development comparison uses 32 training scenarios with three benchmark executions each.
Baseline harness solves 30.21\% at 687.35 Tokens/task (k); the frozen official
AutoSaddler selection solves 38.54\% at 809.18 Tokens/task (k).
Their respective native judge costs are 3.68 and 3.44 Tokens/task (k),
and optimizer expenditure remains separate from execution.

\begin{table}[htbp]
\centering\small
\begin{tabular}{lrrr}
\toprule
Policy & Success (\%) & Tokens/task (k) & Judge/task (k) \\
\midrule
No Harness & 28.13\% & 761.25 & 3.34 \\
Pi (unextended) & 20.83\% & $[7803.45, 7806.14]$ & 2.50 \\
Official SoL-Pi + original adapter & 33.33\% & $[2228.57, 2230.59]$ & 2.81 \\
HarnessMM receipt parent (historical) & 35.42\% & 927.03 & 3.62 \\
HarnessMM exact references (historical) & 43.75\% & 592.65 & 4.05 \\
\bottomrule
\end{tabular}
\caption{Complete GAIA2 training set (32 tasks) with three benchmark repetitions and common
Qwen3.8-27B. Actor input/output includes failures. Bounds reflect five missing
shutdown-call usage records, not sampling uncertainty or zero-imputed cost.
Official SoL-Pi and Pi share the ARE action bridge; file-edit-specific
extensions have no matching tools in this adapter.}
\label{tab:gaia2-official-train-complete}
\end{table}

The selected GAIA2 configuration improves validation and saves 11.18\%
of actor tokens. The respective 2.72 and 3.94 judge Tokens/task (k) give
11.00\% combined saving. Token-wire v3 loses quality and increases
actor cost by 37.08\%, with 2.15 separate judge Tokens/task (k) and no
unknown usage or context rejections. Lazy schemas lose quality at
10.82\% higher actor cost.
Exact ledger references preserve all 24 paired outcomes with 17.02\%
fewer actor tokens. References to unchanged tool actions/observations
replace duplicated receipts; task and tool contracts remain intact.
Parent/candidate use 472/474 paid actor calls and 124,328/120,968
judge tokens.

On all 32 TRAIN scenarios (Table~\ref{tab:gaia2-official-train-complete}),
official SoL-Pi~\cite{liu2026solpi} saves 71.42--71.45\% of Pi's actor
tokens with 33.33\% versus 20.83\% success. It gains five successes over
No Harness at 2.928--2.930 times the cost. Historical HarnessMM exact
references reach 43.75\%, saving 36.07\% against their parent, 22.15\%
against No Harness and 73.41--73.43\% against this SoL-Pi adapter.
This is separate from the CoT/Jev core: its earlier candidate scores
37.50\% versus 50.00\%, saves 5.29\%, and defers all 276 typed decisions.
The 50\% No Harness saving target remains unmet.

The prospective progress-recovery parent also fails its success-first
TRAIN gate. An explicit unmet-constraint checklist and tool-error recovery
solves 0.00\% of scenarios, compared with 3.13\% for the original harness,
while using 28.41\% more actor tokens. The 64 executions account for
604.83 actor Tokens/task (k) exactly once. Input/output usage is complete;
dollar cost and cache-write usage remain unknown. Validation and TEST
stay sealed, so this result rejects the revision without informing
selection on either protected split.

The complete 32-scenario validation diagnostic rejects both uses of the
TRAIN-derived candidate. Direct application loses one success and uses 14.71\%
more actor tokens than the original harness. Frozen routing applies the candidate
once, preserves 6.25\% success, and uses 3.37\% more tokens. The 96 executions
use 745.50 measured actor Tokens/task (k) with no missing token records. These
validation outcomes are used only to reject the candidate; they do not guide method changes, and
TEST remains sealed.

The DeepSeek validation study solves 3.13\% of scenarios under every
mode. Direct application uses 25.47\% fewer actor tokens on this observed
population, while frozen routing applies no residual and saves 6.76\%.
The 96 executions use 666.88 measured actor Tokens/task (k) with complete,
unique request and tool-receipt accounting. One repetition with one success
per mode does not establish a capability improvement or population-level
noninferiority; TEST remains sealed and dollar costs are unknown.

Exact acquisition and inherited development expenditure are retained separately in the external evidence. Benchmark outcome and complete execution-usage audits accompany the frozen policies.

\subsection{Additional training comparisons}
\label{app:latest-train-followups}
Figure~\ref{fig:historical}'s lower strip reports complete development cohorts.
SearchQA risk-220 combines the TRAIN-fitted answer-support ranker, greedy
source packing and a frozen recovery policy. On all 3,395 TRAIN questions
it records 77.53\% EM at 0.97 Tokens/task (k), versus the incumbent's
77.05\% at 2.29 Tokens/task (k): 57.65\% fewer tokens and a +0.471-point observed EM difference.
The paired 95\% interval is $[-0.471,1.443]$ points (1,000 question/context-group resamples).
The complete official-control comparison and subsequent validation rejection
appear in Table~\ref{tab:searchqa-autosota-validation}. The earlier exact-union
family remains in Table~\ref{tab:searchqa-exact-union}.

\begin{table}[htbp]
\centering\small
\setlength{\tabcolsep}{5pt}
\begin{tabular}{lrr}
\toprule
Condition & Correct (\%) & Tokens/task (k) \\
\midrule
\multicolumn{3}{l}{SearchQA: training set (3,395 tasks) Qwen3.5-9B} \\
Baseline harness & 70.81\% & 1.22 \\
Incumbent & 77.05\% & 2.29 \\
risk-220 & 77.53\% & 0.97 \\
\midrule
\multicolumn{3}{l}{AIME 2025: training set (6 tasks) Qwen3.8-27B} \\
Baseline harness & 83.33\% & 6.63 \\
SkillOpt & 83.33\% & 7.04 \\
AutoSaddler & 100.00\% & 6.85 \\
Incumbent & 100.00\% & 7.13 \\
Readout v4 & 100.00\% & 15.54 \\
\midrule
\multicolumn{3}{l}{GAIA2: training set (8 tasks) $\times$ 3, Qwen3.8-27B} \\
No Harness* & 50.00\% & 758.40 \\
Original gate & 37.50\% & 1071.51 \\
Rank gate & 37.50\% & 914.22 \\
\bottomrule
\end{tabular}
\caption{Complete baseline TRAIN results. Costs include every draft, typed
decision, failed call with measured usage and fallback.
The starred No Harness row is a historical same-population reference.}
\label{tab:latest-train-followups}
\end{table}

AIME readout v4 preserves 100.00\% accuracy but uses 15.54 Tokens/task (k);
the unchanged 5.30 Tokens/task (k) target is 80\% of the cheapest named control.
Both model-based one-token readouts reject their checkpoint; all six questions
use incumbent fallback. The independent audit reconstructs all 30 outcomes
and 38 paid calls (8.64 Tokens/task (k) across five conditions). The cost gate
rejects v4; its drafts and decisions remain charged.

The GAIA2 ranker that excludes unknown candidates executes 23 typed fast actions and solves
37.50\%, matching the fresh original gate's total success count while using
14.68\% fewer actor tokens. Three successes are gained and three lost.
It remains below the historical No Harness reference (50.00\% at 758.40
Tokens/task (k)).
The fresh original/rank gates incur 2.70 and 3.87 separate judge Tokens/task (k).
These screening results do not authorize the full four-method comparison or held-out evaluation.

Inherited development and instruction-search expenditure are retained separately in the external acquisition records.

\paragraph{SearchQA optimization with the AutoSOTA workflow.}
We adapt AutoSOTA's workflow of baseline evaluation, diagnosis, candidate updates and independent
evaluation~\cite{tsinghuafib2026autosota}; the official AutoSOTA CLI is not invoked.
TRAIN fits an answer-support ranker and a risk score from public source,
draft and model-reported token-probability features. Greedy source packing supplies
the short answer request. The frozen risk policy assigns incumbent recovery
to 220 of 3,395 questions; every initial draft and paid recovery is counted.
The 6.48\% recovery fraction and all model, scorer and policy hashes
are fixed before validation. Validation exposes aggregate selection statistics
only; no case, answer or failure trajectory enters training.
Official SkillOpt and AutoSaddler receive configured pools of
2,716 optimization and 679 internal-selection questions within TRAIN.
Actual distinct optimizer exposure is 3,395
questions for SkillOpt and 711 for AutoSaddler;
each selected-policy TRAIN and validation evaluation covers all 3,395 questions
in its split. The complete TRAIN comparison
meets the candidate-selection criteria against every method
(Table~\ref{tab:searchqa-autosota-validation}).

\begin{table}[htbp]
\centering\small
\setlength{\tabcolsep}{5pt}
\begin{tabular}{llrrr}
\toprule
Split & Method & EM (\%) & F1 (\%) & Tokens/task (k) \\
\midrule
TRAIN & Baseline harness & 70.81 & 78.65 & 1.22 \\
TRAIN & SkillOpt & 74.40 & 80.98 & 1.61 \\
TRAIN & AutoSaddler & 71.13 & 78.76 & 1.25 \\
TRAIN & HarnessMM incumbent & 77.05 & 83.99 & 2.29 \\
TRAIN & HarnessMM risk-220 & 77.53 & 84.19 & 0.97 \\
\midrule
Validation & Baseline harness & 69.81 & 77.51 & 1.23 \\
Validation & SkillOpt & 72.14 & 78.59 & 1.62 \\
Validation & AutoSaddler & 69.60 & 77.20 & 1.25 \\
Validation & HarnessMM incumbent & 77.20 & 83.78 & 2.30 \\
Validation & HarnessMM risk-220 & 76.76 & 83.36 & 0.99 \\
Validation & Risk-220 w/o recovery & 75.79 & 82.44 & 0.84 \\
\bottomrule
\end{tabular}
\caption{Complete SearchQA TRAIN and protected validation, 3,395 questions
per split under the frozen 3,395/3,395/10,190 partition and Qwen3.5-9B.
Every actor input/output, draft and recovery call is charged.
SkillOpt and AutoSaddler use their pinned official optimization engines
with the same full-context MRQA adapter. The final TEST split is sealed.}
\label{tab:searchqa-autosota-validation}
\end{table}

On complete validation, risk-220 uses 0.99 Tokens/task (k)
at 76.76\% EM and 83.36\% F1.
Baseline harness uses 1.23 Tokens/task (k)
at 69.81\% EM. The token reduction is
19.08\% (paired 95\% interval $[16.57,21.23]$\%),
with a 6.951-point EM gain (paired 95\% interval $[5.685,8.277]$ points).
Risk-220 saves 38.57\% against SkillOpt while improving
EM by 4.624 points, and saves 20.75\%
against AutoSaddler with 7.158 points higher EM.
Against the HarnessMM incumbent, the saving is 56.87\%
with a 0.442-point EM decrease
(paired 95\% interval $[-1.267,0.471]$ points).
Removing risk recovery gives 75.79\% EM at
0.84 Tokens/task (k); recovery adds
0.972 EM points (paired 95\% interval $[0.560,1.443]$).
Intervals use 1,000 question/context-group resamples with seed 20260928.

The frozen release rule requires no benchmark-score regression against
the strongest named control and incumbent, plus at least 20\% saving
against the cheapest named control. Validation misses the incumbent
quality threshold and saves 19.08\% against cheapest Baseline harness.
TEST remains sealed; this comparison is an exploratory quality--cost result.
All 16,980 public MRQA development questions have prior project exposure;
the custom partition is not the official hidden test.

Official optimization costs are separate: the completed SkillOpt loop uses
23,542,512 actor and 8.99 proposer Tokens/task (k); AutoSaddler uses
927,732 actor and 0.11 proposer Tokens/task (k). The selected TRAIN and
validation executions in the table are additional development costs.
Both current p8 optimizer loops have complete usage, with zero unknown-usage calls.
Inherited p7 operational attempts remain charged separately, including
2,028 SkillOpt requests with unknown usage. Administrative research-agent
usage is also unknown, precluding a complete amortized-cost or break-even estimate.

\subsection{Six-condition instruction ablation}
\label{app:alfworld-instruction-ablation}
The four paragraph-presence conditions form a $2\times2$ comparison. The first paragraph contains task-family guidance and the second contains inherited general guidance, which need not be identical across families. The mismatch control rotates the first paragraphs over the six sorted family names. The neutral control uses task-independent symbols and checks full-prompt token equality against the original instructions before every actor request. All variants retain the common actor, benchmark scorer and frozen 710/710/2,133 partition. Instruction transformations are fixed before new validation or test outputs; the underlying public task pool was previously exposed.
\begin{table*}[t]
\centering\small
\begin{tabular}{rlrrr}
\toprule
Seed & Condition & Success (\%) & Tokens/task (k) & Calls/task \\
\midrule
9401 & HarnessMM & 66.20 & 31.12 & 26.43 \\
9401 & w/o general guidance & 61.84 & 28.75 & 28.70 \\
9401 & w/o task-family guidance & 49.27 & 39.04 & 33.54 \\
9401 & w/o skills & 42.29 & 34.05 & 35.61 \\
9401 & w/ another task family & 48.48 & 43.03 & 34.11 \\
9401 & w/ neutral text & 41.07 & 43.34 & 36.02 \\
9402 & HarnessMM & 64.70 & 32.01 & 27.21 \\
9402 & w/o general guidance & 62.59 & 28.60 & 28.51 \\
9402 & w/o task-family guidance & 49.04 & 39.10 & 33.77 \\
9402 & w/o skills & 41.07 & 34.15 & 35.83 \\
9402 & w/ another task family & 47.63 & 43.77 & 34.55 \\
9402 & w/ neutral text & 41.35 & 43.51 & 36.01 \\
9403 & HarnessMM & 66.20 & 31.18 & 26.59 \\
9403 & w/o general guidance & 61.60 & 28.73 & 28.61 \\
9403 & w/o task-family guidance & 49.32 & 38.68 & 33.37 \\
9403 & w/o skills & 41.73 & 34.10 & 35.78 \\
9403 & w/ another task family & 48.90 & 43.03 & 33.98 \\
9403 & w/ neutral text & 41.68 & 42.93 & 35.76 \\
\bottomrule
\end{tabular}
\caption{Instruction ablations verified with the benchmark scorer. Each row contains all 2,133 tasks. Replication seeds reuse task identities and are not independent task populations. Only complete six-condition seeds are shown.}
\label{tab:alfworld-instruction-ablation}
\end{table*}
\begin{table*}[t]
\centering\small
\begin{tabular}{rlrrrr}
\toprule
Seed & Control & Full minus control (pp) & 95\% interval & Gains & Losses \\
\midrule
9401 & w/o general guidance & +4.36 & [2.61, 6.21] & 232 & 139 \\
9401 & w/o task-family guidance & +16.92 & [13.16, 20.28] & 470 & 109 \\
9401 & w/o skills & +23.91 & [19.15, 28.30] & 603 & 93 \\
9401 & w/ another task family & +17.72 & [14.15, 21.02] & 491 & 113 \\
9401 & w/ neutral text & +25.13 & [20.32, 29.44] & 621 & 85 \\
9402 & w/o general guidance & +2.11 & [0.48, 3.76] & 215 & 170 \\
9402 & w/o task-family guidance & +15.66 & [11.88, 19.20] & 460 & 126 \\
9402 & w/o skills & +23.63 & [18.82, 27.98] & 608 & 104 \\
9402 & w/ another task family & +17.07 & [13.36, 20.50] & 499 & 135 \\
9402 & w/ neutral text & +23.35 & [18.82, 27.43] & 602 & 104 \\
9403 & w/o general guidance & +4.59 & [2.68, 6.48] & 245 & 147 \\
9403 & w/o task-family guidance & +16.88 & [13.07, 20.23] & 474 & 114 \\
9403 & w/o skills & +24.47 & [20.09, 28.42] & 618 & 96 \\
9403 & w/ another task family & +17.30 & [14.07, 20.20] & 480 & 111 \\
9403 & w/ neutral text & +24.52 & [20.04, 28.52] & 603 & 80 \\
\bottomrule
\end{tabular}
\caption{Paired benchmark-success differences against the full skill. Pointwise 95\% intervals use 10,000 bootstrap draws over 108 floorplans, with analysis seed 20260930. All paired tasks in a sampled floorplan stay together. The exploratory intervals have no multiplicity adjustment.}
\label{tab:alfworld-instruction-ablation-paired}
\end{table*}
Seed 9401 reuses the two original fully audited reference conditions exactly once; its four new conditions are freshly executed. Additional seeds execute all six conditions. TRAIN operability and validation executions are separate from the test costs above. The retired ten-condition TRAIN pilot is not included as test evidence. Cumulative token termination remains disabled.
Seed 9401 contains 8,532 new test executions (154.16 Tokens/task (k)) and 4,266 reused reference executions (65.17 Tokens/task (k)).
Seed 9402 contains 12,798 new test executions (221.13 Tokens/task (k)) and 0 reused reference executions (0.00 Tokens/task (k)).
Seed 9403 contains 12,798 new test executions (218.64 Tokens/task (k)) and 0 reused reference executions (0.00 Tokens/task (k)).

\section{Historical diagnostics and negative results}
\label{app:historical-diagnostics}
\subsection{Bank authorship in the native Codex interface}
\label{app:codex-author-contract}
On the same 85 MBPP-sanitized training tasks, an author-contract condition
asks the model to write a solution, the original public-example checker
and an optional guarded continuation bank through the native Codex tools.
All conditions use Qwen3.5-4B, the common token-efficiency objective,
the original eight tools and 300-second actor/30-second verifier limits.
The condition solves 83 tasks versus 82 for the native control, with
two paired gains and one loss: a 1.18-point difference with a task-bootstrap
95\% interval of $[-2.35,5.88]$ (10,000 resamples, seed 20261103).
Every paid request includes the author
instruction, but the model authors no accepted bank and executes no
callback substitution. The success difference therefore does not
establish a decision-reuse effect.

\begin{table}[htbp]
\centering\small
\begin{tabular}{lrrr}
\toprule
Condition & Success (\%) & Tokens/task (k) & Bypasses \\
\midrule
Native Codex & 96.47 & 65.30 & 0 \\
Bank-authoring instruction & 97.65 & 115.52 & 0 \\
Mandatory bank authorship & 96.47 & 453.70 & 23 \\
\bottomrule
\end{tabular}
\caption{Complete native execution on all 85 MBPP-sanitized training
tasks per condition, one attempt per task. Input and output costs include
all authoring, control, fallback and unsuccessful calls. The optional
bank-authoring instruction produces no substitution; mandatory authorship
produces 23. These training conditions are separate from the selected
test results in Figure~\ref{fig:industrial-harness}.}
\label{tab:codex-author-contract}
\end{table}

The necessary preceding training and diagnostic calls add 305.16
Tokens/task (k), bringing this condition's complete training-program
cost to 420.68 Tokens/task (k), normalized over the same 85 tasks.
The native control's 65.30 Tokens/task (k) belongs to its issuing arm
and is counted once in joint accounting. These measurements establish
neither an execution saving nor an all-in RSI saving. Validation and
test remain unopened for this condition.

Requiring bank authorship activates reuse on 15 of the 85 tasks: the
model authors 27 banks on 18 tasks and executes 23 guarded
substitutions, of which seven actions succeed. Native task success
remains 82/85, with two paired gains and two losses; the task-bootstrap
95\% interval for the success difference is $[-4.71,4.71]$ percentage
points (10,000 resamples, seed 20261104). Execution cost rises to
453.70 Tokens/task (k), and the complete training program costs
874.38 Tokens/task (k), including preceding diagnosis and candidate
evaluations over the same 85-task denominator. The native control
remains charged once at 65.30 Tokens/task (k). Bank activation alone
therefore does not establish a token saving. Validation and test
remain unopened for this condition.

\subsection{Separating task rules from model calls}
\label{sec:alfworld-forensic}
The action adapter contains a handwritten task solver. Running its
unchanged public-state rules from the initial state, without skill
text, actor initialization or model fallback, solves 92.39\% of validation
tasks versus 92.82\% for rules plus actor. The three gains and zero
losses give a 0.42-point difference at 1.08 additional Tokens/task (k).
Fixed rules account for most success; this contrast does not isolate
learned instructions from initialization and fallback.

\begin{table}[htbp]
\centering\small
\begin{tabular}{lrrr}
\toprule
Configuration & Success (\%) & Tokens/task (k) & Calls/task \\
\midrule
Rules + actor & 92.82 & 1.08 & 1.04 \\
Rules only & 92.39 & 0.00 & 0.00 \\
\bottomrule
\end{tabular}
\caption{All 710 validation tasks, seed 9401, common 50-step limit,
public observations and tool actions. Failures remain in both
denominators. Zero model tokens do not imply zero CPU computation.}
\label{tab:alfworld-model-free}
\end{table}

Independent benchmark replay also reproduces all 2,010 archived outcomes
and provider-token totals for the full controller, skills-only actor
and three prompt controls. The historical 100\% result includes
handwritten task rules and does not measure an RSI optimization gain.

\subsection{Archived action-controller comparisons}
\label{sec:jev-mechanism-confirmation}
\label{app:alfworld-no-refresh-ablation}
The replication covers all 134 official \texttt{valid\_unseen}
tasks, seeds 9401--9403 and a common 50-step limit. The support-first
handwritten controller solves 100.00\% at each seed, versus 67.16\%,
67.91\% and 69.40\% for its skills-only actor. The stall-gated predecessor
solves 97.76\%, 99.25\% and 97.76\%; removing its progress filter gives
35.07\%, 32.84\% and 34.33\%. Removing its skill text gives 97.76\%,
97.01\% and 97.76\%. These interventions change the earlier controller;
the instruction-only ablation is reported separately in
Figure~\ref{fig:alfworld-ablation}. All ten configurations and their
per-seed costs remain in \path{generated/jev_mechanism_confirmation.tex}.
The 134 tasks, not the 402 repeated executions, are the independent units.

On the separate custom test set (2133 tasks) support-first execution solves
95.08\% of tasks at 1.00 Tokens/task (k), versus 92.55\% at 9.91 for
stall gating. The paired 80 gains and 26 losses give a 2.53-point
difference (95\% floorplan-cluster interval $[0.98,4.07]$) and 89.89\%
token saving. Both use the same instructions, selector and 50-step cap.
This compares historical controller configurations. The earlier
development set (60 tasks) selection used opened tasks and unequal historical limits;
it supplies no additional held-out evidence.

\subsection{Recorded execution examples}
\label{app:execution-examples}
The archived stall-gated traces illustrate both progress and failure.
In the heating example, the controller picks up \texttt{egg 3} at step
13, heats it at 15, and places it in the fridge at 17. Heating feedback
updates the tracked state, excluding the otherwise legal cooling action.
The episode succeeds in 18 steps with seven model calls and 8.17
Tokens/task (k) for that single episode. In the two-book example, the placement ledger distinguishes the
first completed book from the remaining target; the run succeeds in
22 steps with 11 calls and 9.96 Tokens/task (k).
In the heated-cup failure, supported exploration reaches the still-closed
fridge at the final allowed step, 48, without finding and heating the cup.
Local support does not guarantee timely task completion.
These are the lexicographically first recorded success/failure examples
in their categories, bound in \path{evidence/jev_execution_examples.json};
they are illustrative, not additional evaluation samples.

\subsection{Compression and transfer failures}
Earlier SearchQA studies use 5,000 opened training-source questions.
The 9B compact-context deterministic adapter reaches 70.54 EM at
1.90 Tokens/task (k) versus 75.46 at 3.00 for fresh original HarnessMM:
36.71\% fewer tokens with a 4.92-point loss. Model-selected recovery
reaches 73.42 at 3.02 Tokens/task (k). The later evaluations retain full source context.

The separate studies with the same backbone also fail their quality targets.
Model-based typed checks reach 72.12 EM at 4.08 Tokens/task (k) versus
73.32 at 3.00 Tokens/task (k) for original HarnessMM, falling back on 87.80\% of tasks.
Probability-only full/compact source selection reaches 44.16/45.06 EM.
Adding 96 CoT tokens gives 29.00 versus 27.06 for its fresh zero-CoT
control; restoring a short evidence sentence gives 35.10 versus
19.98 for its label-alias control. The initial evidence-generation
attempt was inactive and scored 20.10. Source-rule typed recovery
reaches 69.72 versus 72.08 for fresh original HarnessMM, saving only
10.50\% against its 30\% target. Each study retains its own model,
request template, complete controls and accounting in the source bundle;
none replaces the later full-source result.

Spreadsheet's opened LOGO267 replay gives 4.12\% at 48.36 Tokens/task (k)
for archived Baseline harness, 3.00\% at 51.58 Tokens/task (k) for SkillOpt, 2.25\% at 47.77 Tokens/task (k)
for AutoSaddler, and 3.00\% at 11.78 Tokens/task (k) for HarnessMM.
The last three use fresh executions; HarnessMM retains 170 repair/abstention
routes and 97 Baseline harness routes chosen post hoc by workbook group.
Its separate seven-condition test on 200 opened tasks gives
deterministic recovery 2.50\%, matching original HarnessMM at 25.06\% lower
cost, but below Baseline harness and AutoSaddler at 3.50\%.
Its 97-task state-controller study solves 0.00\% versus the parent's
1.03\% while increasing cost. The earlier Terminal-Bench subset gives
HarnessMM 8.33\% versus 25.00\% for each baseline; the visual-web subset gives
25.00\% versus 33.33\% for Baseline harness and SkillOpt.

The AIME draft-isolation diagnostic scores 70.00\% versus 80.00\% for both
visible-draft verification and the shared drafts. GAIA2's
feedback-consistent controller reduces repeated restorations from 64
to 3 and Tokens/task (k) from 929.89 to 838.67, but scores 83.33\%
versus 90.00\% for its original harness evaluated on the same development tasks in a new run.
The older budget-limited GAIA2 run and email-pagination activation
diagnostic have incomplete outcomes across methods; neither supports a
quality or cost comparison.

\subsection{Instruction-content and visual diagnostics}
The five-module response study evaluates all 32 masks, ten
equal-length body controls and Original A/A on 256 discovery tasks
and 257 held-out-stage tasks in 256 source groups.
The full harness solves 84.82\%; the discovery-selected
normalization/answer-type mask solves 84.44\% at 23.91\% lower cost.
No module passes the primary familywise criterion, and no average
main effect or pair interaction passes the Holm-corrected secondary
tests. Nineteen tasks vary in EM across masks; the outcome-selected
oracle reaches 86.77\% and is not a deployable selector.

The targeted replication retains those 19 tasks, two F1-only cases
and 32 stable controls, with six arms in eight blocks (four serial,
four 16-concurrent). Eighteen of the 19 EM effects retain their mean
direction, and 17 retain a nonzero sign in every block; both F1-only
effects persist. Stable controls have zero EM edge effects, while
Original A/A changes EM on two variable tasks.
These are conditional, outcome-selected effects; 13 of the 19
nominated modules concern normalization.
All retained module bytes, ordering and answer contracts are unchanged,
and generic/unrelated controls exactly match complete model input
token counts.

The GQA pilot evaluates 40 balanced questions in 19 image groups,
plus 88 entailment-support questions, with three repeats.
True-image accuracy is 62.50\% for the base and 65.00\% for both
grounding and generic instructions of equal token length; shuffled-image
accuracy is 35.00\%, 33.33\% and 35.00\%, respectively.
Grounding and generic correctness coincide on every balanced task.
Their true-minus-shuffled interaction is 1.67 points, with a
descriptive image-cluster interval of $[-2.44,7.69]$.
Images use 448-by-448 padded renderings with 196 model image tokens.
The 2,304 benchmark requests consume 0.27 Tokens/task (k); admission adds
500. Model image tokens and the unchanged publisher scorer were
replayed independently~\cite{hudson2019gqa}.

Two later feasibility studies also stop before final evaluation.
Text calibration evaluates 512 groups over three repeats after
training on 2,048 groups; all four adaptive thresholds fail the joint
gate. The least restrictive saves 4.3734\% while lowering EM and F1;
the quality-preserving three-short-route policy saves only 0.1483\%.
For the MMMU instruction library, 270 calls on 90 training questions
consume 0.65 Tokens/task (k), including 0.51 input-image/task (k). Even zero
output permits only 0.366\% saving, below the 10\% target.
This is a cost-feasibility bound; task accuracy was not measured.
The completed gate records remain in
\path{generated/token_efficiency_protocol.tex}; unexecuted plans
are omitted from the paper.

\section{Reproducibility}
\label{app:evidence}
\path{generated/CLAIM_EVIDENCE.json} binds reported values to source
artifacts, model identities and evaluation scope. Raw remote-run records
retain frozen policies and partitions, model/tokenizer revisions,
physical GPU assignments, seeds, packages, requests, outputs and usage,
including failures. Audits reconstruct original API requests, scores and
tokens. Historical tables retain their original cohorts, method
identities and costs in \path{generated/}.

In 17 audited GAIA2 worker launches, startup \texttt{PYTHONHASHSEED}
was not recorded. ARE 1.1.0 uses Python's salted hash for some receipt
ordering and app-RNG seeds; a fixed model-server seed does not establish
identical tool trajectories. OS-generated UUIDs and measured
simulation timing remain variable. The source audit is bound in
\path{evidence/gaia2_native_seed_audit_20261002.json}.

The primary source bindings are
\path{evidence/alfworld_fresh226_panel.json},
\path{evidence/figure1_bdf.json},
\path{evidence/aime202060_complete.json},
\path{evidence/industrial_harness_panel.json},
\path{evidence/industrial_harness_deepseek_test.json} and
\path{evidence/figure1_df_current.json}.
Appendix~\ref{app:quality-cost-followup} binds the quality--cost follow-up.
Exact split IDs accompany the 20/20/60 protocol; benchmark licenses
and release conditions apply.

Token totals include all input/output, truncations and fallback.
Search, rejected proposals, validation, judges and shared acquisition
are separate; inherited calls are charged once. Finite break-even
requires measured development cost and positive per-task savings.
Unknown historical costs stay unknown. Runs have no cumulative token
stop; model capacities and benchmark task limits remain.
In the paper directory, \texttt{bash build.sh} renders and compiles saved
results without model or benchmark calls; \path{template-dependency.json}
pins the CVPR and bibliography styles.

\section{Figure 3 evaluation settings}
\label{app:figure3-targeted}
Figure~\ref{fig:industrial-harness}(a) uses the complete frozen 32-scenario GAIA2 development population from the custom 32/32/96 partition, seed 20260928. Panel (b) uses all 18 frozen Terminal-Bench development tasks from the custom 18/18/53 partition. These panels report development measurements. Each harness pair shares its backbone, public task prompt, original tools, task limits and benchmark scorer.

Panel (c) uses MBPP-sanitized's complete 427-task release with the custom 85/85/257 partition, seed 20261005. The DeepSeek Harness pair reports all 257 TEST tasks; the independent Codex pair completes all 257 TEST tasks at 236/257 and 233/257 successes (91.83\% and 90.66\%). Both harness pairs use the same fixed backbone. Actors receive the released public assertions and common instructions, with a 300-second actor deadline and 30-second verifier deadline. Reference implementations are excluded from actor workspaces.

\begin{center}\small
\begin{tabular}{lrrr}
\toprule
Harness & Correct / 85 & Success (\%) & Tokens/task (k)\\
\midrule
Codex & 78 & 91.76 & 67.21\\
Codex+HarnessMM & 80 & 94.12 & 59.78\\
DeepSeek Harness & 73 & 85.88 & 15.46\\
DeepSeek Harness+HarnessMM & 80 & 94.12 & 11.91\\
\bottomrule
\end{tabular}
\end{center}
These earlier comparisons on 85 training tasks evaluate the compound and structured literal action banks. The DeepSeek condition records 153 certified substitutions and 62 public-guard confirmations. TRAIN develops candidates, validation selects only TRAIN-frozen configurations, and TEST evaluates the selected configuration without further revision.

Panel (d) uses the complete 164-task HumanEval release, custom 33/33/98 partition, seed 20261006, and one attempt per task. All four methods use the same fixed backbone. Actors receive the function signature, specification and docstring examples; reference solutions and verifier tests are excluded from their context. The Harbor adapter evaluates released functional tests with the common 300-second actor and 30-second verifier deadlines. The original harness must meet the training success criterion before its candidate proceeds to validation and selection.

The complete HumanEval test set (98 tasks) Codex pair preserves 90.82\% success at 66.72 versus 47.02 Tokens/task (k), a 29.54\% reduction. The DeepSeek Harness pair improves from 71.43\% to 85.71\% at 96.28 versus 11.62 Tokens/task (k), an 87.93\% reduction.

Panel (e) uses MiMo-v2.6-pro through the official DeepSeek Harness SDK Anthropic-compatible interface. The same 33/33/98 HumanEval partition, prompts, tools, scorer and one-attempt policy apply to both arms. The original harness reaches 96.97\%, its frozen candidate reaches 100.00\%, and both validation arms reach 96.97\% before selection. Complete test set (98 tasks) success is 96.94\% versus 97.96\%, at 11.62 versus 4.40 Tokens/task (k), a 62.12\% reduction. The MiMo Codex pair completes test set (98 tasks) at 93.88\% versus 94.90\% success; its costs are 62.60 versus 10.09 Tokens/task (k). The MiMo Codex MBPP comparison completes test set (257 tasks) at 248/257 baseline and 250/257 HarnessMM successes (96.50\% and 97.28\%), with 47.56 versus 11.00 Tokens/task (k). Panel (f) also reports the complete MiMo DeepSeek Harness test set (257 tasks) comparison: 251/257 versus 252/257 successes (97.67\% and 98.05\%), at 6.20 versus 5.11 Tokens/task (k). The author uses a concise literal-source prompt with the same structured action-bank schema and native tool interface; CPU receipt guards execute 480 certified substitutions. The custom 85/85/257 partition and seed 20261005 remain fixed. TRAIN develops the candidate, protected validation selects it, and source and selection freeze before TEST.

Actor costs are reported as input-plus-output tokens per task, divided by 1,000. Input and disjoint cache-input buckets are counted once.

\subsection{Industrial harness comparisons}
\label{app:pending-results}
\begin{center}\small
\begin{tabular}{lrrl}
\toprule
Pair & Baseline (\%) & HarnessMM (\%) & Tasks\\
\midrule
Codex MBPP & 91.83 & 90.66 & 257 / 257\\
MiMo Codex HumanEval & 93.88 & 94.90 & 98 / 98\\
MiMo Codex MBPP & 96.50 & 97.28 & 257 / 257\\
MiMo DeepSeek MBPP & 97.67 & 98.05 & 257 / 257\\
\bottomrule
\end{tabular}
\end{center}

Costs are normalized over all tasks in each comparison and include unsuccessful executions. Exact measurements and cost scopes remain in the reproduction artifacts.

The earlier MiMo Codex candidate completes the same 257 test tasks at 94.55\% success and 10.45 Tokens/task (k); the selected successor reaches 97.28\% with 11.00 Tokens/task (k). Both use the same native task population and limits.

\subsection{Qwen DeepSeek Harness on 257 MBPP test tasks}
The 427 MBPP-sanitized tasks use 85 TRAIN/85 validation/257 TEST, seed 20261005, and one attempt per task. Both actors receive all released assertions; reference solutions are excluded. Selection and source freeze before TEST.
The official dsh-v0.2.0-rc.1 and Harbor 0.18.0 MBPP adapter share the backbone, prompts, tools and 300-second actor/30-second verifier limits. Success is 90.27\% for the baseline and 91.83\% for HarnessMM. Input/output Tokens/task (k) are 50.40/3.58 (53.98 total) and 6.84/3.77 (10.61 total), respectively. Complete costs cover slow authors, fast decisions, finalizers, fallbacks and failed executions; development costs are separate.
HarnessMM makes 472 certified substitutions. CoT-only reaches 93.39\% success at 15.04 Tokens/task (k): 11.21 input and 3.84 output. Savings are 80.35\% against the original harness and 29.46\% against CoT-only. The 40\% target against both baselines with quality preserved remains open. TEST never guides revisions.

\clearpage
\section{Visual implementation details}
\label{hv:app:implementation}
\subsection{Execution rules and accounting}
\label{hv:detail:sec:problem}
\label{hv:detail:sec:method}
Section~\ref{hv:sec:method} defines image-bound decisions, runtime eligibility and harness-to-harness RSI; Section~\ref{sec:evolution-cost} defines complete-task cost. Here we specify the implementation choices behind those definitions.

\paragraph{Harness optimization and runtime execution.}
Equation~\ref{hv:eq:rsi-harness} separates proposal, training evaluation and retention. The proposer changes graph-authoring, persistence or callback-dispatch rules under the declared schema, visual-access and native-tool contracts. GDR adds stored decisions, predecessor dependencies, visual guards and reuse routes while retaining the ordinary slow path. CCS replaces an eligible repeated reasoning/tool-use subgraph with a certified native operation, preserving the action and its outgoing dependencies. Complete training executions determine native quality and actor cost; only candidates satisfying the experiment's fixed criteria are retained. Validation selects among training-frozen harnesses before testing. Within the selected harness, each episode instantiates its own continuation $G_\tau$ from current observations. Screenshots and receipts update runtime eligibility while the selected harness and authored literal decisions remain fixed.

\paragraph{Generalization across tasks.}
Within-benchmark studies evaluate unseen tasks that share the native interface and interaction structure. Training develops graph authorship and reuse; each new episode instantiates its own graph from current observations. The cross-benchmark study in Section~\ref{hv:sec:cross-benchmark} transfers a geometry-based MiniWoB++ harness unchanged to LongForm-UI, with only the environment entry point and viewport adapted. One fixed source configuration passes through target training and validation without parameter updates before testing. Benchmark-specific configurations used elsewhere retain their own training and selection protocols.

\paragraph{Authoring and target binding.}
Each continuation is acyclic, with unique node identifiers, existing predecessors and resolved arguments. MiniWoB++ and LongForm-UI continuation graphs contain up to eight future nodes. The frozen authoring policy determines whether a later slow call can create another graph; the transfer studies retain each source policy. VisualWebArena permits two ordered click/type successors and repeated authorship at later slow boundaries. The Qwen3.5-9B and MiMo-V2.6-Pro studies retain separate frozen configurations and populations.

An anchor binds a node to its author image, fixed region and reference signature. MiniWoB++ uses a model-authored ROI containing the action point; typing additionally requires a pixel-only closed-input-box predicate. VisualWebArena compiles a visible control footprint or radius-24 patch before root execution. Both configurations keep the compiled region and coordinates fixed throughout reuse. Coordinates compile once from image fractions into native pixels, with raw and compiled arguments retained. All interactive arms share the compiler, normalized interface and constrained JSON decoder.

A type operation clicks its fixed point and replaces the field text through the common native wrapper. Future focus-dependent key operations require slow inference or a separately frozen extension. A form-filling continuation can therefore order a click, literal entry and submission through native predecessor receipts without searching for a target at execution time.

\paragraph{GDR growth, eligibility and persistence.}
GDR supplies guarded paths that reuse resolved decisions. At runtime it checks author-time target support, successful predecessors, unexecuted status, image dimensions and the fixed visual guard. MiniWoB++ compares decoded ROI pixels exactly; the web guard also accepts its frozen achromatic edge-and-literal-ink signature and checks scene/input state (Appendix~\ref{hv:app:native-vwa}). Evidence records signatures, predecessor binding, verdict, runtime and model-call count. Guard CPU work is measured separately from actor tokens.

Eligible nodes execute in authored order. Each receipt can activate a descendant at the next observation boundary; reuse depth counts actual executions within that graph instance. A rejected guard returns to slow inference and abandons the unresolved remainder. A later slow call may author a fresh graph when the frozen configuration permits it, including the source policies used in the GPT transfer study. Invalid outputs, additional authorship and recovery remain charged. The ablations change one setting at a time: disable graph execution, allow one execution per graph or retain the slow callback.

\paragraph{CCS compression and execution certificates.}
\label{hv:detail:sec:visual-contract}
\label{hv:detail:eq:visual-sufficiency}
CCS replaces redundant callback paths with certified native execution. Following GDR admission, it dispatches stored arguments through the original native interface. Its certificate binds the graph, decision, author/current images, anchor, guard, predecessor receipt, arguments, native receipt and registered callback. Independent audit recomputes image signatures and checks that before/after slow-call counts match the physical ledger, with no request bearing that episode's omitted callback key. The original author request remains charged.

Certificates record native acceptance and callback omission; the task scorer independently measures success. Equation~\ref{hv:eq:visual-sufficiency} specifies dispatch eligibility from current visual evidence and native predecessor receipts. Appendix~\ref{hv:app:visual-contract} specifies the state machine and its implementation invariants.

\paragraph{Accounting.}
Every physical author, actor, selector and fallback request contributes input plus output usage, including unsuccessful executions. Once-only shared requests are allocated once within the declared evaluation scope; the MiniWoB++ capability check is allocated to final testing. Displayed Tokens/task (k) divides the complete arm total by all declared executions and by 1,000. Multimodal prompt usage is provider measured; image tokens are separated only when measured. Image count, resolution, model latency, episode duration and verifier CPU time are reported separately. Savings compare complete token totals over the same tasks.

\section{MiniWoB++ execution contract and algorithm}
\label{hv:app:visual-contract}
The screenshot-only implementation stores the authored graph separately from runtime state. Each future node contains an identifier, one predecessor, a serialized click/type action, reference image size, fixed rectangle and pixel signature. Each node in this implementation has one predecessor, represented as a singleton in the general predecessor set. The root can additionally use key, wait or stop. Focus-dependent key actions use the slow actor; future nodes use click/type actions with fixed targets. Each region is at least twelve pixels wide and high and contains its fixed action point; a call may author eight future nodes.

Signatures hash decoded pixels, with equality subject to the hash's collision resistance. Coordinates and text compile before persistence. The shared type wrapper clicks the declared point, selects existing text and enters the literal string without target search or structured page access. Appendix~\ref{hv:app:native-vwa} specifies the web configuration's repeated authorship and compiler-bound regions.

\paragraph{Exclusive execution loop.}
GDR supplies guarded reuse paths; CCS replaces redundant callback paths with certified native execution. Within the frozen harness, GDR establishes eligibility before CCS executes the stored action:
\begin{enumerate}
\item Obtain the screenshot and preceding public receipt; register the next callback identifier before routing.
\item Check unresolved nodes against completed predecessors, image dimensions and fixed ROI equality. Select the first eligible node in authored order.
\item If eligible, record the slow-call counter and execute its immutable native action once. Store the receipt and mark completion. Certify omission only when arguments match, the predecessor exists and the slow-call counter is unchanged.
\item Otherwise invoke the original visual actor with the screenshot and public history. At the first decision, validate and compile the root and continuation. At later slow boundaries, execute the fresh action and abandon the unresolved graph. Charge authoring and every fallback.
\item Repeat until native completion or the common action/time limit, then score through the independent reward interface.
\end{enumerate}
The retained-callback control invokes and charges the callback before executing the selected graph action and emits no omission certificate. Single-use discards the remaining graph after its first substitution; authorship-only always calls the slow actor. Each EDG arm permits one authoring attempt per episode.

\paragraph{Implementation invariants.}
Graph parsing checks schema, acyclic dependencies, bounds and anchor containment; malformed output triggers a measured model fallback. Serialized arguments and a Boolean-only guard preserve authored action identity. Certificate construction checks arguments against the native receipt. Completion-map membership excludes a second execution of the same node instance. Identical content in a later graph remains a distinct authoring event with separately charged requests.

Every physical request has a persistent intent before submission and raw response before interpretation. Omission audit uses run, arm, task and callback label together: the frozen Qwen3.5-9B runner stores episode-local labels, while the API runner serializes their full scope. A valid native receipt, unchanged counters and the complete scoped request ledger establish callback absence. Native scoring independently establishes task effect.

\section{Visual verification and action validity}
The guard checks fixed ROI pixels and predecessor receipts, while the native scorer evaluates task success. Equation~\ref{hv:eq:visual-sufficiency} specifies dispatch eligibility from matching evidence and predecessor completion; native scoring establishes task success. The actor resolves target coordinates and text before execution; runtime checks preserve these arguments. Shared actor-side screenshot hints in later configurations are specified separately.

\section{Complete cost and causal interpretation}
\label{hv:app:visual-cost}
The following measurements describe the earlier fixed-coordinate MiniWoB++ comparison, retained separately from the common-geometry comparison in Figure~\ref{hv:fig:quality}. For each arm, divide physical prompt-plus-completion tokens over all 96 test tasks by 96 and one thousand. All author outputs, retained callbacks, malformed outputs, paid fallbacks and unsuccessful episodes remain included. Full-method input and output components are 3.34 and 0.47 Tokens/task (k); the baseline harness components are 3.82 and 0.54. Components and combined cost are rounded independently.

We report verifier CPU time separately from model-token usage. Episode duration includes model calls, native actions, observations and scheduling. Prompt usage covers the complete multimodal input, with image-request counts reported alongside it.

Authorship-only measures graph-generation overhead without graph execution. Single-use limits persistence. The retained-callback condition invokes and charges the slow decision before executing the stored action. The baseline harness versus Full HarnessMM-V measures the integrated policy, including author prompts and resulting trajectories. Total call-count reduction includes both certified substitutions and subsequent changes in reasoning and recovery. If extra authorship and recovery cost $A$ and avoided callbacks would cost $S$, token benefit requires $S>A$. Because $S$ is counterfactual for omitted requests, measured savings compare complete episode totals on the same tasks.

\section{Evaluation protocol and reproduction}
The earlier development comparison uses six predetermined instances from four families; smoke and remaining instances complete eight per family. Each arm executes an instance once, using 480 by 630 pixel screenshots, fifteen native actions and a 180-second allowance. The later protected split assigns twenty groups of eight instances to four training, four validation and twelve test groups before outcomes. These populations remain distinct; instance bootstrap intervals condition on the observed families.

The reproduction bundle retains integer input/output usage, source/model identities, task membership, images, requests and settlements. Actors receive pixels and public action history; setup and reward APIs remain outside their interface. Appendix~\ref{hv:app:miniwob-protocol} gives the shared validation and test settings.

\subsection{MiniWoB++ environment and ablation settings}
\label{hv:app:miniwob-protocol}
\subsubsection{Visual environment and access boundary}
The MiniWoB++ controls with Qwen3.5-9B run on remote compute using the same Qwen3.5-9B multimodal endpoint. The browser renders the official MiniWoB++ task page at a fixed viewport and supplies a 480 by 630 pixel screenshot. The user instruction remains visible in the screenshot. Actor inputs contain the image and public action history; they contain no DOM, accessibility tree, HTML source, expected answer, or evaluator state. The controller accesses the released episode initialization and final reward APIs for task setup and independent evaluation, outside the actor and guard interfaces.

The agent acts through native browser mouse and keyboard operations. A normalized coordinate is converted to an original-image pixel at authorship and then to the browser's fixed coordinate system. Every arm uses the same conversion. A type operation clicks the declared coordinate and replaces the field's text; neither the native wrapper nor the guard searches for an input element. Invalid author output triggers an independently logged model fallback and retains its cost.

We report exact task success when the official raw reward equals one after native completion. The original raw reward and time-adjusted reward are retained alongside success. Episodes that exhaust the declared action budget without native completion count as unsuccessful. Evaluator and infrastructure errors have a separate invalid status. Every scheduled episode in a complete comparison has a valid native score and verified model-request costs.

\subsubsection{Population and freeze policy}
The prospective population contains twenty disjoint official task-family groups, with eight instances per group. Four groups supply 32 training tasks, four supply 32 validation tasks, and twelve supply 96 test tasks. Numbered and popup variants are excluded prospectively to avoid splitting near-duplicate templates across phases. The assignment seed and instance seeds are frozen before outcomes, and all compared methods use the same instances. No rounding is needed because the group sizes are equal.

Training observations supply prompt, schema, anchor and control-policy revisions. Protected validation selects among candidates frozen after training; its answers and trajectories remain outside optimization. The selected implementation, model identity, prompt, schema, admission rules, action budget and scorer are frozen before test execution. The Qwen3.5-9B evaluation completes all 96 held-out instances and reports every outcome. This custom screenshot-only MiniWoB++ subset tests transfer across disjoint task families.

\subsubsection{Five-condition mechanism ablation}
The token-efficiency prompt and objective are common to all compared methods, including the baseline harness, native harnesses and HarnessMM-V. Each is instructed to reduce complete-task input-plus-output token consumption while preserving task performance. Method-specific action and graph instructions retain the configurations specified below.

The five conditions separately test graph execution, persistence and callback removal. The baseline harness generates a fresh action at every screenshot. EDG authorship only produces the shared graph representation but disables its execution. Single-use EDG permits one substitution per graph. GDR with the callback retained selects and executes the graph action while invoking and charging the registered slow callback. Full HarnessMM-V retains the graph for later execution and certifies each omitted callback.

All arms share backbone, screenshots, native actions, tasks, seeds, output capacity and execution budget. The graph arms share author prompts, coordinate compilation, action and anchor schema, guard rule and fallback. Authorship-only disables EDG execution; single-use limits GDR; retained callbacks disable CCS omission. Full execution enables both. These ablations test GDR reuse and CCS callback substitution (Figure~\ref{fig:design}). Persistence can change outcomes only when the graph supports multiple executions.

The earlier fixed-coordinate study used development targets of a token saving of at least 20\%, a task-success loss of no more than five percentage points, and at least one audited callback substitution. Those targets determined admission to the next protected stage. Every scheduled task, including unsuccessful executions, contributes to the reported quality and cost.

\subsubsection{Shared screenshot inputs and geometry}
The current common-geometry study shares image-only control centers and author-time literal projection across all arms; its graph arms may author a new continuation at a required slow boundary. A common read-only screenshot OCR prefix also supplies visible text and row alignment. The current compact-graph candidate requires at least 30\% complete saving in both training and validation, no paired native-quality loss and actual certified second-and-later reuse; Full must solve all 32 training tasks. Selection freezes before testing. The selected test comparison reports every task and its complete cost against the 30\% objective.

DeepSeek Harness preserves its native reasoning, tool arguments and receipts. It sends the latest complete public observation and two most recent full screenshots, omitting superseded observations from later model calls. Typed native GUI proposals retain the fifteen-attempt limit.

\subsubsection{SoL-Pi native runtime}
SoL-Pi uses the released Pi v0.85.1 runtime with action fusion, observation packing, evidence-preserving reduction and online context compaction enabled. The primary actor and reducer both use Qwen3.5-9B. The native visual interface supplies the same full screenshots, public OCR and geometry, task instances, 15-action limit and 180-second episode allowance as the other MiniWoB++ methods. The fixed configuration completes all 32 training tasks and 32 validation tasks before testing on all 96 test tasks. It solves 58 test tasks (60.42\%) at 41.99 Tokens/task (k). Every primary, reduction, compaction and unsuccessful execution call contributes to the test total; native training and validation verification are itemized separately in Table~\ref{hv:tab:miniwob-cost-breakdown}.

\subsubsection{Codex native runtime}
Codex uses the official CLI v0.142.5 with the same Qwen3.5-9B actor, complete screenshots, public OCR and geometry, native GUI actions and task instances. Its adapter applies the shared limit of 15 action attempts, including rejected arguments, and the 180-second native episode allowance. Click and type operations explicitly require both coordinates; type also requires text. Native completion is checked throughout the session. Every model response already in flight is drained and charged. The configuration is frozen after all 32 training tasks, before 32 validation tasks and 96 test tasks. It solves 57 test tasks (59.38\%) at 44.77 Tokens/task (k), including every auxiliary and unsuccessful call. Table~\ref{hv:tab:miniwob-cost-breakdown} separates native verification from test execution.

\subsection{Persistence with shared screenshot OCR and geometry}
\label{hv:app:persistence-train}
This separate complete training comparison uses all 32 original MiniWoB++ training tasks once per arm, under the same disjoint 20/20/60 task-family split. The common Qwen3.5-9B sees the full screenshot, complete public action history, image-derived control centers and read-only CPU OCR lines with boxes and confidence. Read-only OCR supplies observed text to the actor; the actor selects actions, and runtime guards use pixels and native receipts. Both arms share the same prompt, schema, seed, model, native tools, scorer and limits. Every physical request is charged.

The common author-time compiler projects compatible points within 144 pixels for clicking and 72 pixels for typing to image-detected control centers and binds future targets to fixed 18-by-18-pixel patches. Arguments remain literal after graph construction. Runtime checks require exact authored-patch equality and predecessor receipts. The arms differ only in whether each graph can execute one future action or retain its remaining actions for later execution.

Single-use and Full solve 32 and 32 of 32 tasks (100.00\% and 100.00\%). Complete cost falls from 1.69 to 1.16 Tokens/task (k), saving 31.35\%. Input components are 1.50 and 1.01 Tokens/task (k), and output components are 0.19 and 0.15 Tokens/task (k); components round independently. Calls fall from 45 to 32 and certified omissions rise from 36 to 47. Full executes 16 nodes beyond the first reuse across 12 graph instances.

Independent checks reconstruct raw provider usage, screenshots, OCR inputs, literal compilation, native receipts and pixel certificates. All initial image/request pairs match; independent model decoding may produce different authored outputs. Task-pair resampling with 10,000 draws gives a 95\% saving interval of [20.34, 39.90]\% and a success-difference interval of [0.00, 0.00] percentage points. These are training measurements, separate from both main-text test comparisons.

The slow author returns an ordered array of literal future actions. A deterministic author-time compiler assigns graph identifiers, sequential dependencies and fixed control patches; an unsupported future target and its dependent tail abstain. Common OCR also reads visible text inside pixel-detected control rectangles. Certified fast boundaries skip unused actor-perception construction; runtime guards still use only the current screenshot and native receipts.

\subsection{Persistence with a shared screenshot-geometry parent}
\label{hv:app:persistence-geometry-earlier}
The screenshot-geometry comparison accompanying the earlier five-arm study uses all 32 original training tasks, once per arm, under the original family-group partition with 32 training, 32 validation and 96 test tasks. The shared Qwen3.5-9B, screenshots, author prompt, constrained schema, request seed, native mouse/keyboard interface, official raw-reward scorer, 15-action limit, 180-second episode allowance and 2,048-token response capacity are frozen before execution. Four immutable groups use the same admitted service for each task pair, with arm order counterbalanced. Every physical author, recovery and unsuccessful request is charged to its arm.

A common image-only geometry stage detects closed control rectangles from screenshot pixels. It supplies normalized center coordinates without DOM, accessibility, text-label or evaluator information. The frozen author-time compiler maps an emitted click or type point within 48 pixels of a compatible rectangle to its center; typing requires a light editable interior. It binds an admitted future target to a fixed 18-by-18-pixel patch. These compiled arguments remain literal after graph construction. Runtime guards require exact equality of the authored patch. Both graph arms may author a new continuation at a later required slow boundary. The arms differ only in whether the graph is discarded after one reuse or retained for later actions.

All 32 initial screenshot and author request/response pairs match exactly. Native outcomes agree task by task: both solve 29 tasks (90.63\%). Single-use EDG and full HarnessMM-V cost 1.51 and 1.10 Tokens/task (k), respectively, giving 27.30\% lower complete cost. Input costs are 1.24 and 0.88 Tokens/task (k); output costs are 0.27 and 0.21 Tokens/task (k). Components and totals are rounded independently. Slow calls decrease from 47 to 34. Certificate counts are 36 and 48; full HarnessMM-V executes 16 nodes beyond the first reuse across 12 persistent graph instances. Its maximum observed reuse depth is four.

The independent check reconstructs every physical request, provider usage, screenshot-derived coordinate hint, author-time action/region projection, native execution receipt and pixel certificate. No model request supplies geometry or verifies reuse. Paired resampling over all 32 task pairs uses 10,000 draws and seed 20261019; the 95\% interval for saving is [16.87, 36.58]\%, and the success-difference interval is [0.00, 0.00] percentage points. This training mechanism study complements the five-arm test comparison in Figure~\ref{hv:fig:quality}. The next subsection reports the fixed-coordinate training pair.

\subsection{Fixed-coordinate persistence ablation}
\label{hv:app:persistence-train-earlier}
An earlier persistence comparison executes every one of the 32 original training tasks once per arm. The four task families are login-user, click-checkboxes, enter-text and click-button-sequence, with eight frozen instances each. All instances remain in the success and cost denominators. Membership follows the original family-group partition of 32 training tasks, 32 validation tasks and 96 test tasks; no protected data supply revisions.

Before execution, we freeze the same Qwen3.5-9B, author prompt, constrained response schema, literal-coordinate compiler, pixel guard, native mouse/keyboard interface, raw-reward scorer, 15-action limit, 180-second episode allowance and 2,048-token response capacity for both arms. A new request seed, 20261019, is fixed for this comparison. Four immutable groups distribute independent task pairs across identical existing model services on both H200 nodes, with arm order counterbalanced across groups. Each pair uses the same service. The only difference is the single-reuse cap versus repeated graph execution. Both arms still pay for their own author request and all later native-reasoning and fallback requests. No additional model request is used for setup or checking.

The independently checked initial screenshots and author request/response pairs match exactly for all 32 tasks. Native task-success outcomes also match task by task: 15/32 for each arm. Input cost is 2.86 Tokens/task (k) for single reuse and 2.20 for persistence; output cost is 0.48 and 0.39 Tokens/task (k), respectively. Components and totals are rounded independently. Combined costs are 3.34 and 2.59 Tokens/task (k), respectively, a 22.49\% saving over the entire training population. This is a distinct training comparison, retained alongside the original five-arm held-out study.

Figure~\ref{hv:fig:quality}(c) reports the reuse-depth distribution for this 32-task training comparison. Depth counts certified native executions after the root within one authored graph. Single reuse has 19 graphs at depth one; persistence has 16, 1, 1 and 1 graphs at depths one, two, three and five, respectively. These give 19 versus 26 certified executions, including zero versus seven second-and-later executions.

For every certificate, the audit reconstructs the author and observed screenshots, exact ROI pixels, stored action arguments, predecessor order, native execution receipt and omitted callback. Reuse depth is counted within one authored graph; a later graph with identical content is counted separately. All 45 certificates pass these checks. The 29-call full-episode reduction includes downstream differences in reasoning and recovery after reuse; seven additional certificates identify direct substitutions beyond the first node.

Paired task bootstrap resampling uses all 32 task pairs, 10,000 draws and seed 20261019. The conditional 95\% interval for complete token saving is $[0.00,45.04]\%$; the success-difference interval is $[0.00,0.00]$ percentage points. Three graphs supply the additional reuse opportunities. This training comparison measures the additional reuse supplied by persistence; the original held-out comparison is reported separately. Original requests, screenshots, native rewards, fixed memberships and exact costs remain in the external evidence package.

\subsection{Common initial harness across RSI methods}
\label{hv:app:rsi-common-start}
The original 32-task comparison evaluates HarnessMM-V, SkillOpt, AutoSaddler, RHO and HarnessPro across eight complete RSI rounds with Qwen3.5-9B and GPT-5.6 Sol. Figure~\ref{hv:fig:matched-revisions} retains its Qwen curves and uses the expanded 160-task native Responses study for GPT-5.6 Sol (Appendix~\ref{hv:app:rsi-expanded-training}). Within each original model group, the five methods start from the same instruction and reuse configuration and use the same 32 MiniWoB++ training tasks, screenshots, native scorer, 15-action limit and 180-second episode allowance. The outer partition contains 32 training tasks, 32 validation tasks and 96 test tasks; validation and test outcomes do not enter these curves. All proposers use GPT-5.6 Sol with high reasoning effort and the shared objective of preserving native success while reducing complete task input-plus-output tokens.

HarnessPro uses the same initial harness and all 32 training tasks in each model group. Its paper-based visual reproduction uses tiered optimization and a generated-test DSL. This optimization uses disjoint internal update and selection sets of 16 tasks each, label-free test synthesis, and native assessments of the retained medium-tier harness. Eight complete rounds precede protected validation and testing. HarnessMM-V, SkillOpt and AutoSaddler edit the same typed instruction and reuse configuration, compiled into a fixed native execution kernel. RHO retains its released optimization core: representative task selection, three candidate rollouts, diagnosis and label-free self-preference selection. Its configured adapter edits general instructions, public action history, profiles, classifiers and convergence hooks while preserving the native actor, actions, scorer and episode limits. Native training assessments supply the plotted measurements and do not replace RHO's label-free optimization feedback. SkillOpt and AutoSaddler retain their released core implementations, with configured native-quality-first selection and token cost as the tie-breaker. Their internal update and selection sets each contain 16 disjoint tasks from the training population; HarnessMM-V uses the complete training population for candidate retention. The AutoSaddler provider uses the original workspace editor and a schema-bound continuation for fully measured outputs that omit the required session file. Each distinct compiled checkpoint is assessed once on all 32 training tasks, and later evaluations of the same checkpoint reuse that complete assessment. Curves report the retained checkpoint after each method's update, including unchanged checkpoints. Every full candidate assessment and proposer call remains in the reproduction record.

The MiniWoB++ comparison starts RHO from the original native baseline harness. HarnessPro starts from the same native baseline and completes eight tiered optimization rounds with a paper-based visual runtime and generated-test DSL, followed by the same 32 validation tasks and 96 test tasks. Its disjoint internal update and selection sets contain 16 training tasks each. It solves 59 test tasks (61.46\%) at 10.03 Tokens/task (k). Its eight optimization rounds use the same 32 training tasks, followed by 32 validation tasks and 96 test tasks. It retains the native screenshots, action interface, scorer and episode limits. The MiniWoB++ comparison reports complete test execution cost for every method. Table~\ref{hv:tab:miniwob-cost-breakdown} separately retains optimizer proposals, candidate rollouts, shared setup, native training-checkpoint assessment and validation. Each physical call contributes once to the total; initial-trajectory acquisition is shared equally between the two learning methods that use it. These components describe the registered procedure and selected configuration, while independent earlier research campaigns retain their own costs.

\begin{table}[t]
\centering\scriptsize
\caption{MiniWoB++ cost breakdown for the visual harness comparisons. Every entry is Tokens/task (k), using all 96 test tasks as the denominator. Learning includes optimizer proposals, native candidate rollouts and shared setup. Verification includes native training-checkpoint assessments and validation. The total adds each component once; it covers the registered learning procedure and selected configuration, with independent earlier research campaigns kept separately. $\dagger$ Learning and total are measured subtotals; test execution and verification are fully measured.}
\label{hv:tab:miniwob-cost-breakdown}
\setlength{\tabcolsep}{3pt}
\begin{tabular}{lrrrr}
\toprule
Method & Execution & Learning & Verification & Total \\
\midrule
baseline harness & 7.36 & 0.00 & 2.06 & 9.42 \\
SkillOpt & 7.97 & 9.79 & 2.74 & 20.51 \\
AutoSaddler & 7.87 & 12.21 & 2.58 & 22.65 \\
SoL-Pi & 41.99 & 0.00 & 16.29 & 58.28 \\
Prompt Compaction & 7.39 & 0.00 & 2.21 & 9.61 \\
DeepSeek Harness & 20.03 & 12.38 & 9.42 & 41.84 \\
Codex Harness & 44.77 & 0.00 & 11.47 & 56.24 \\
RHO & 9.10 & 59.33 & 8.78 & 77.20 \\
HarnessPro & 10.03 & $130.91^{\dagger}$ & 6.62 & $147.56^{\dagger}$ \\
HarnessMM-V & 5.07 & 0.00 & 0.99 & 6.06 \\
\bottomrule
\end{tabular}
\end{table}

\begin{table}[t]
\centering\small
\setlength{\tabcolsep}{4pt}
\begin{tabular}{llrr}
\toprule
Model & Method & Success (\%) & Tokens/task (k) \\
\midrule
Qwen3.5-9B & HarnessMM-V & 96.88 & 1.46 \\
 & SkillOpt & 90.63 & 2.20 \\
 & AutoSaddler & 78.13 & 2.97 \\
 & RHO & 81.25 & 3.89 \\
 & HarnessPro & 71.88 & 5.32 \\
\midrule
GPT-5.6 Sol & HarnessMM-V & 100.00 & 4.53 \\
 & SkillOpt & 100.00 & 4.92 \\
 & AutoSaddler & 100.00 & 11.11 \\
 & RHO & 100.00 & 10.96 \\
 & HarnessPro & 100.00 & 11.11 \\
\bottomrule
\end{tabular}
\caption{Retained harnesses after eight RSI rounds on 32 training tasks per model. Costs cover all calls within each task execution.}
\label{hv:tab:rsi-common-start}
\end{table}

With Qwen3.5-9B, HarnessMM-V solves 31 of 32 tasks, compared with 29 for SkillOpt, while using 33.52\% fewer execution tokens. With GPT-5.6 Sol, all five methods solve all 32 tasks; HarnessMM-V reduces execution tokens by 59.20\% relative to the common initial harness and by 7.81\% relative to SkillOpt. These curves and endpoint costs exclude proposal generation, candidate assessment and selection overhead. The complete first-use comparison in Appendix~\ref{hv:app:rsi-first-use} separately measures whether RSI recovers its overhead.

After eight RSI rounds and validation on 32 separate tasks, the original GPT-5.6 Sol Copilot CLI configuration evaluates all 96 test tasks once. RHO solves 76 tasks (79.17\%) at 19.06 Tokens/task (k), including every test-time actor call. These execution costs exclude optimization and validation.

\subsection{RSI on an expanded training population}
\label{hv:app:rsi-expanded-training}
We also run four RSI methods on 160 MiniWoB++ training tasks with GPT-5.6 Sol at high reasoning effort through the native Responses interface. The GPT-5.6 Sol curves in Figure~\ref{hv:fig:matched-revisions} use this expanded population. The original 32-task GPT comparison above uses the official Copilot CLI interface. The population follows the frozen extension of the original partition: 160 training tasks, 160 validation tasks and 480 test tasks, with disjoint task families. Each method starts from exactly the same initial harness and original task executions. All methods receive the same objective of preserving native success while reducing complete task input-plus-output tokens, and run eight complete RSI rounds. Model identity, full screenshots, the native action interface, scorer and episode limits remain fixed.

\begin{table}[t]
\centering\small
\setlength{\tabcolsep}{5pt}
\begin{tabular}{lrr}
\toprule
Method & Success (\%) & Tokens/task (k) \\
\midrule
HarnessMM-V & 93.13 & 1.66 \\
SkillOpt & 92.50 & 2.25 \\
AutoSaddler & 93.13 & 2.05 \\
RHO & 92.50 & 3.15 \\
\bottomrule
\end{tabular}
\caption{Retained harnesses after eight RSI rounds on the complete expanded population of 160 training tasks. Costs include all calls within each native task execution.}
\label{hv:tab:rsi-expanded-training}
\end{table}

HarnessMM-V raises native success from 92.50\% to 93.13\% and reduces execution cost from 3.15 to 1.66 Tokens/task (k), a 47.39\% reduction. At the final retained checkpoint, SkillOpt reaches 92.50\% at 2.25 Tokens/task (k), AutoSaddler reaches 93.13\% at 2.05 Tokens/task (k), and RHO reaches 92.50\% at 3.15 Tokens/task (k). These are full-population training measurements. Proposal, candidate-assessment and selection costs remain separate from these execution costs; the complete first-use comparison is reported in Appendix~\ref{hv:app:rsi-first-use}.

\paragraph{Held-out evaluation.}
After eight RSI rounds and validation on 160 separate tasks, the retained HarnessMM-V, SkillOpt, AutoSaddler and RHO harnesses each execute all 480 test tasks once. The task families, original screenshots, actor, native action interface, scoring and episode limits follow the same frozen protocol.
\begin{table}[t]
\centering\small
\setlength{\tabcolsep}{5pt}
\begin{tabular}{lrr}
\toprule
Method & Success (\%) & Tokens/task (k) \\
\midrule
HarnessMM-V & 81.88 & 3.91 \\
SkillOpt & 80.21 & 4.63 \\
AutoSaddler & 79.58 & 4.18 \\
RHO & 82.29 & 9.72 \\
\bottomrule
\end{tabular}
\caption{Complete evaluation on 480 held-out MiniWoB++ tasks. Costs are measured input-plus-output subtotals over all 480 tasks, including unsuccessful requests. Unmeasured components and RSI overhead are excluded.}
\label{hv:tab:rsi-expanded-test}
\end{table}

HarnessMM-V solves 393 test tasks, compared with 385 for SkillOpt and 382 for AutoSaddler. RHO solves 395 tasks at 9.72 Tokens/task (k). HarnessMM-V uses 15.59\% fewer measured execution tokens than SkillOpt and 6.36\% fewer than AutoSaddler (Table~\ref{hv:tab:rsi-expanded-test}).

\subsection{Persistence comparisons across harness RSI iterations}
\label{hv:app:matched-revisions}
Table~\ref{hv:tab:matched-revisions} compares HarnessMM-V with single-use EDGs across eight consecutive training-guided harness RSI iterations on the same complete population of 32 MiniWoB++ training tasks. Each arm executes every task once per RSI iteration using Qwen3.5-9B, full screenshots, the same task instances and actor seed, native rewards, a 15-action limit and a 180-second episode allowance. Within each RSI iteration, both arms share the author prompt, visual context, coordinate compiler, guards and fallback. Single-use EDG permits one future execution per graph; HarnessMM-V retains eligible nodes across later feedback boundaries. Training feedback guides updates to author-time grounding and public screenshot context across RSI iterations. These harness RSI iterations follow chronological training feedback, with validation and test reserved for subsequent selection and evaluation.

\begin{table*}[t]
\centering\small
\setlength{\tabcolsep}{5pt}
\begin{tabular}{crrrrrrr}
\toprule
& \multicolumn{2}{c}{Success (\%)} & \multicolumn{2}{c}{Tokens/task (k)} & & \multicolumn{2}{c}{Calls} \\
Iteration & Single-use & HarnessMM-V & Single-use & HarnessMM-V & Saving (\%) & Single-use & HarnessMM-V \\
\midrule
1 & 37.50 & 34.38 & 3.91 & 3.26 & 16.63 & 114 & 96 \\
2 & 34.38 & 34.38 & 3.74 & 2.95 & 21.15 & 128 & 104 \\
3 & 90.63 & 90.63 & 1.51 & 1.10 & 27.30 & 47 & 34 \\
4 & 84.38 & 84.38 & 1.88 & 1.71 & 8.94 & 66 & 61 \\
5 & 90.63 & 90.63 & 1.50 & 1.10 & 27.08 & 47 & 34 \\
6 & 90.63 & 93.75 & 1.37 & 0.96 & 29.76 & 45 & 32 \\
7 & 93.75 & 93.75 & 1.58 & 1.26 & 20.12 & 53 & 43 \\
8 & 90.63 & 90.63 & 1.51 & 1.06 & 29.83 & 47 & 33 \\
\bottomrule
\end{tabular}
\caption{Eight RSI iterations on 32 MiniWoB++ training tasks per arm. Savings use unrounded episode totals; Calls counts physical model requests.}
\label{hv:tab:matched-revisions}
\end{table*}

At the eighth RSI iteration, both methods solve 29 of 32 tasks (90.63\%). HarnessMM-V reduces complete execution cost from 1.51 to 1.06 Tokens/task (k), a 29.83\% saving. Input components decrease from 1.24 to 0.85 Tokens/task (k), and output components from 0.27 to 0.22 Tokens/task (k); components round independently. Physical model calls fall from 47 to 33, while certified omissions increase from 35 to 47. HarnessMM-V executes 16 second-and-later nodes within persistent graphs, directly identifying reuse beyond the single-use cap. The paired task-bootstrap 95\% interval for token saving is [18.84, 39.00]\%. This descriptive training comparison complements the five-arm held-out controls in Figure~\ref{hv:fig:quality}, where full HarnessMM-V saves 21.62\% on MiniWoB++ and 73.49\% on LongForm-UI relative to single-use EDGs.

\subsection{Training-guided harness RSI}
\label{hv:app:rsi-evolution}
The initial and final source-frozen harnesses are evaluated on the same complete population of 32 MiniWoB++ training tasks with Qwen3.5-9B, full screenshots, native rewards, a 15-action limit and a 180-second episode allowance. Training develops author-time visual grounding, public screenshot context and persistent execution. Validation selects the frozen candidate before held-out testing.

\begin{table}[t]
\centering\small
\begin{tabular}{lrr}
\toprule
Harness & Success (\%) & Tokens/task (k) \\
\midrule
Initial & 46.88 & 2.59 \\
Training-derived & 100.00 & 1.16 \\
\bottomrule
\end{tabular}
\caption{Initial and training-derived harnesses on 32 tasks. The 55.17\% saving includes all candidate execution calls.}
\label{hv:fig:rsi-evolution}
\end{table}

The final harness solves all 32 tasks, compared with 15 initially, and reduces candidate execution cost by 55.17\%. This endpoint comparison describes the improvement obtained through training-guided harness RSI. The external research record retains every evaluated candidate and its cost.

\subsection{RSI author-prompt compression}
\label{hv:app:rsi-compression}
A separate RSI proposal replaces repeated author-prompt instructions with a shorter complete prompt. The optimizer uses only the 32 training tasks. AST checks preserve the native actor, coordinate compiler, graph executor and scorer; the model, screenshots, action limits and task partition remain fixed. The candidate solves all 32 training tasks at 1.08 Tokens/task (k). Protected validation then selects this frozen candidate: both arms solve 6 of 32 tasks, at 4.28 Tokens/task (k) for the direct harness and 1.37 for the candidate.

On all 96 test tasks, executed once per arm after selection, the direct harness solves 58 tasks (60.42\%) and the compressed HarnessMM-V harness solves 60 (62.50\%). Complete execution cost falls from 8.14 to 4.06 Tokens/task (k), a 50.14\% saving. Every author, control, fallback and unsuccessful execution is included. These measurements establish the selected harness's deployment benefit. These deployment savings exclude optimization, candidate evaluation, selection, validation and historical research costs. A net RSI benefit requires adding those measured costs and comparing the same complete workload.

\subsection{Complete first-use efficiency}
\label{hv:app:rsi-first-use}
This experiment measures bank-anchored RSI on LongForm-UI, the controlled static-form benchmark. HarnessMM-V first acquires model-authored decision banks on all 32 training tasks. The structural proposer uses this bank's task-quality and cost feedback to select guarded reuse within the fixed execution kernel. It evaluates the proposed policy on the complete training population and stops when all demonstrated tasks succeed with certified substitutions. The native actor, model, full screenshots, literal actions, visual guards and scorer remain fixed. A training cache adopts only byte-identical public model requests; their original acquisition fees are charged once. Protected evaluations disable this cache.

One actual proposal selects persistent reuse with a maximum of 12 guarded actions. All 32 training tasks and all 32 validation tasks succeed. The source and policy are frozen before the 96 held-out tasks, with one execution per task and method. Complete first-use cost includes bank acquisition, the proposer, every candidate evaluation, validation, all test executions and all author, control and fallback calls. Time is measured continuously from fresh bank acquisition through the last test result, including preparation, native scoring, independent verification and scheduling. Both methods use the same two pinned Qwen3.5-9B service endpoints and eight workers; the model services are already loaded for both.

\begin{table}[t]
\centering\small\setlength{\tabcolsep}{5pt}
\begin{tabular}{lrrr}
\toprule
Method & Success (\%) & Tokens/task (k) & Time (min) \\
\midrule
Direct harness & 12.50 & 30.86 & 3.34 \\
HarnessMM-V & 100.00 & 3.96 & 2.91 \\
\bottomrule
\end{tabular}
\caption{LongForm-UI first-use efficiency on 96 test tasks. HarnessMM-V costs include bank acquisition, RSI, training, validation and testing.}
\label{hv:tab:rsi-first-use}
\end{table}

HarnessMM-V solves 96 of 96 tasks versus 12, while reducing complete first-use token consumption by 87.18\% and elapsed time by 12.87\%. These measurements show that the selected RSI workflow recovers its optimization and selection overhead within this 96-task workload.

\subsection{Case study: depth and breadth at the same search budget}
\label{hv:app:rsi-graph-case}
We prospectively compare two HarnessMM-V search schedules on the same complete population of 32 MiniWoB++ training tasks. Both use GPT-5.6 Sol at high reasoning effort through the native metered Copilot adapter, two rounds, three proposed candidates per round and one full training assessment per distinct candidate. Each schedule evaluates six candidates, covering 192 candidate--task pairs. The initial assessment and first-round candidates are shared, giving both schedules identical first-round evidence. In the second round, Depth proposes children of the accepted first-round harness; Breadth proposes further children of the fixed initial harness. Both receive exactly the same complete first-round feedback. Only the assigned parent differs.

The schedules, task membership, candidate budget, acceptance rule and example task are fixed in advance. Optimization and acceptance use all 32 training tasks; the disjoint 32-task validation and 96-task test populations remain unopened. The editable interface is the same for both schedules: graph mode, reuse limit and task-independent author instructions. Each batch contains at least two structural alternatives and may include an instruction refinement. The model, full screenshots, public history, screenshot-derived OCR/geometry, coordinate compiler, literal-action and pixel guards, native tools, native reward and 15-action limit remain fixed. Candidate assessments use the same six browser workers. Both schedules optimize native success first and then complete input-plus-output tokens. A new candidate is accepted only if it strictly improves this ordering; otherwise the incumbent is retained.

\begin{figure*}[t]
\centering
\resizebox{\textwidth}{!}{%
\begin{tikzpicture}[x=1cm,y=1cm,>=Latex,font=\sffamily\scriptsize,
 kept/.style={draw=teal!70!black,fill=teal!8,rounded corners=2pt,align=center,text width=1.48cm,minimum height=.65cm,line width=.8pt},
 candidate/.style={draw=gray!65,fill=gray!5,rounded corners=2pt,align=center,text width=1.48cm,minimum height=.6cm},
 keptedge/.style={->,draw=teal!75!black,line width=.8pt},
 proposededge/.style={->,draw=gray!60,dashed},
 action/.style={draw=black!60,fill=gray!6,rounded corners=1pt,minimum height=.42cm,minimum width=.8cm,font=\sffamily\scriptsize},
 model/.style={circle,draw=blue!65!black,fill=blue!7,inner sep=2.3pt,font=\sffamily\scriptsize},
 guard/.style={diamond,draw=teal!70!black,fill=teal!10,inner sep=1pt,font=\sffamily\scriptsize}]
\node[anchor=west,font=\sffamily\bfseries\small] at (0,3.55) {(a) Propose three, accept one};
\node[kept] (root) at (.9,0) {Initial\\Native};
\node[candidate] (s1) at (3.4,1.1) {1a\\Single use 1};
\draw[proposededge] (root)--(s1);
\node[kept] (s2) at (3.4,0) {1b\\Reuse 3\\Accept 1};
\draw[keptedge] (root)--(s2);
\node[candidate] (s3) at (3.4,-1.1) {1c\\Native};
\draw[proposededge] (root)--(s3);
\node[anchor=west,font=\sffamily\small] at (5.7,2.9) {Depth, round 2};
\node[kept] (dp) at (6.7,1.5) {Accepted 1\\Reuse 3};
\draw[keptedge] (s2.east)--++(.5,0)|-(dp.west);
\node[anchor=west,font=\sffamily\small] at (5.7,-.65) {Breadth, round 2};
\node[kept] (bp) at (6.7,-2.2) {Initial\\Native};
\draw[proposededge] (root.south)--++(0,-3.85)-|(bp.south);
\node[kept] (depth1) at (10.4,2.6) {D2a\\Reuse 4\\Accept 2};
\draw[keptedge] (dp)--(depth1);
\node[candidate] (depth2) at (10.4,1.5) {D2b\\Reuse 2};
\draw[proposededge] (dp)--(depth2);
\node[candidate] (depth3) at (10.4,0.4) {D2c\\Reuse 3};
\draw[proposededge] (dp)--(depth3);
\node[kept] (breadth1) at (10.4,-1.1) {B2a\\Reuse 4\\Accept 2};
\draw[keptedge] (bp)--(breadth1);
\node[candidate] (breadth2) at (10.4,-2.2) {B2b\\Reuse 2};
\draw[proposededge] (bp)--(breadth2);
\node[candidate] (breadth3) at (10.4,-3.3) {B2c\\Reuse 3};
\draw[proposededge] (bp)--(breadth3);
\node[anchor=west,font=\sffamily\small] at (.05,2.9) {Shared round 1};
\node[anchor=west,font=\sffamily\bfseries\small] at (0,-4.55) {(b) Accepted execution changes};
\node[anchor=north west,inner sep=0pt] at (.05,-5.2) {\includegraphics[width=2.2cm]{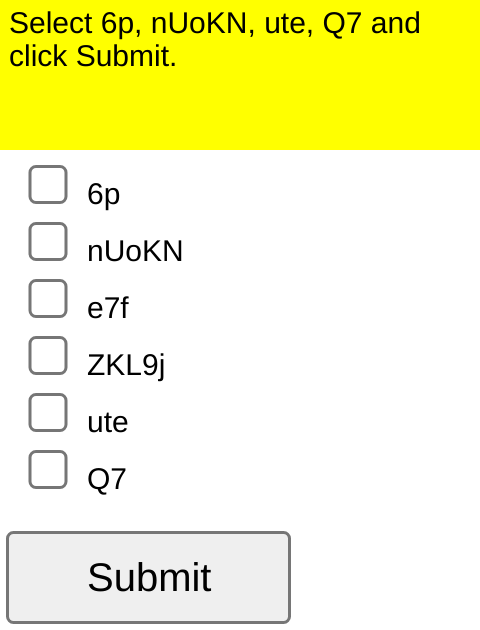}};
\node[anchor=west,font=\sffamily\scriptsize] at (2.9,-5.14) {Initial};
\node[model] (r0n0m) at (3.1,-5.6) {M};
\node[action] (r0n0a) at (4.1,-5.6) {6p};
\draw[->] (r0n0m)--(r0n0a);
\node[model] (r0n1m) at (5.550000000000001,-5.6) {M};
\node[action] (r0n1a) at (6.550000000000001,-5.6) {nUoKN};
\draw[->] (r0n1m)--(r0n1a);
\draw[->] (r0n0a)--(r0n1m);
\node[model] (r0n2m) at (8.0,-5.6) {M};
\node[action] (r0n2a) at (9.0,-5.6) {ute};
\draw[->] (r0n2m)--(r0n2a);
\draw[->] (r0n1a)--(r0n2m);
\node[model] (r0n3m) at (10.450000000000001,-5.6) {M};
\node[action] (r0n3a) at (11.450000000000001,-5.6) {Q7};
\draw[->] (r0n3m)--(r0n3a);
\draw[->] (r0n2a)--(r0n3m);
\node[model] (r0n4m) at (12.9,-5.6) {M};
\node[action] (r0n4a) at (13.9,-5.6) {Submit};
\draw[->] (r0n4m)--(r0n4a);
\draw[->] (r0n3a)--(r0n4m);
\node[anchor=west,font=\sffamily\scriptsize] at (2.9,-6.09) {Round 1};
\node[model] (r1n0m) at (3.1,-6.55) {M};
\node[action] (r1n0a) at (4.1,-6.55) {6p};
\draw[->] (r1n0m)--(r1n0a);
\node[guard] (r1n1m) at (5.550000000000001,-6.55) {G};
\node[action] (r1n1a) at (6.550000000000001,-6.55) {nUoKN};
\draw[->] (r1n1m)--(r1n1a);
\draw[->] (r1n0a)--(r1n1m);
\node[guard] (r1n2m) at (8.0,-6.55) {G};
\node[action] (r1n2a) at (9.0,-6.55) {ute};
\draw[->] (r1n2m)--(r1n2a);
\draw[->] (r1n1a)--(r1n2m);
\node[guard] (r1n3m) at (10.450000000000001,-6.55) {G};
\node[action] (r1n3a) at (11.450000000000001,-6.55) {Q7};
\draw[->] (r1n3m)--(r1n3a);
\draw[->] (r1n2a)--(r1n3m);
\node[model] (r1n4m) at (12.9,-6.55) {M};
\node[action] (r1n4a) at (13.9,-6.55) {Submit};
\draw[->] (r1n4m)--(r1n4a);
\draw[->] (r1n3a)--(r1n4m);
\node[anchor=west,font=\sffamily\scriptsize] at (2.9,-7.04) {Depth, round 2};
\node[model] (r2n0m) at (3.1,-7.5) {M};
\node[action] (r2n0a) at (4.1,-7.5) {6p};
\draw[->] (r2n0m)--(r2n0a);
\node[guard] (r2n1m) at (5.550000000000001,-7.5) {G};
\node[action] (r2n1a) at (6.550000000000001,-7.5) {nUoKN};
\draw[->] (r2n1m)--(r2n1a);
\draw[->] (r2n0a)--(r2n1m);
\node[guard] (r2n2m) at (8.0,-7.5) {G};
\node[action] (r2n2a) at (9.0,-7.5) {ute};
\draw[->] (r2n2m)--(r2n2a);
\draw[->] (r2n1a)--(r2n2m);
\node[guard] (r2n3m) at (10.450000000000001,-7.5) {G};
\node[action] (r2n3a) at (11.450000000000001,-7.5) {Q7};
\draw[->] (r2n3m)--(r2n3a);
\draw[->] (r2n2a)--(r2n3m);
\node[guard] (r2n4m) at (12.9,-7.5) {G};
\node[action] (r2n4a) at (13.9,-7.5) {Submit};
\draw[->] (r2n4m)--(r2n4a);
\draw[->] (r2n3a)--(r2n4m);
\node[anchor=west,font=\sffamily\scriptsize] at (2.9,-7.989999999999999) {Breadth, round 2};
\node[model] (r3n0m) at (3.1,-8.45) {M};
\node[action] (r3n0a) at (4.1,-8.45) {6p};
\draw[->] (r3n0m)--(r3n0a);
\node[guard] (r3n1m) at (5.550000000000001,-8.45) {G};
\node[action] (r3n1a) at (6.550000000000001,-8.45) {nUoKN};
\draw[->] (r3n1m)--(r3n1a);
\draw[->] (r3n0a)--(r3n1m);
\node[guard] (r3n2m) at (8.0,-8.45) {G};
\node[action] (r3n2a) at (9.0,-8.45) {ute};
\draw[->] (r3n2m)--(r3n2a);
\draw[->] (r3n1a)--(r3n2m);
\node[guard] (r3n3m) at (10.450000000000001,-8.45) {G};
\node[action] (r3n3a) at (11.450000000000001,-8.45) {Q7};
\draw[->] (r3n3m)--(r3n3a);
\draw[->] (r3n2a)--(r3n3m);
\node[guard] (r3n4m) at (12.9,-8.45) {G};
\node[action] (r3n4a) at (13.9,-8.45) {Submit};
\draw[->] (r3n4m)--(r3n4a);
\draw[->] (r3n3a)--(r3n4m);
\draw[teal!75!black,line width=.8pt,rounded corners=2pt] (5.1,-6.88) rectangle (12.05,-6.20);
\draw[teal!75!black,line width=.8pt,rounded corners=2pt] (5.1,-7.83) rectangle (14.5,-7.15);
\draw[orange!85!black,line width=.8pt,rounded corners=2pt] (12.48,-7.83) rectangle (14.5,-7.15);
\node[anchor=west,font=\sffamily\scriptsize,text=black!70] at (2.9,-9.0) {M: model callback\qquad G: visual guard and predecessor receipt};
\end{tikzpicture}}
\caption{Depth and breadth search with six candidates each. (a) Shared first round, then descendant versus initial-graph proposals; teal/solid marks accepted edits. (b) Recorded checkbox actions, model calls and certified substitutions. Boxes mark the guarded execution span; orange highlights the newly reused Submit boundary. Protocol: Section~\ref{hv:app:rsi-graph-case}.}
\label{hv:fig:rsi-graph-case}
\end{figure*}

\begin{table}[t]
\centering\small
\setlength{\tabcolsep}{4pt}
\begin{tabular}{lrrrr}
\toprule
& & \multicolumn{2}{c}{Tokens/task (k)} & \\
Method & Success (\%) & Execution & Search & Calls \\
\midrule
Depth & 100.00 & 4.66 & 5.36 & 32 \\
Breadth & 100.00 & 4.66 & 5.36 & 32 \\
\bottomrule
\end{tabular}
\caption{Depth/breadth search with six candidates each. Execution covers 32 training tasks; Search covers all assessments and proposer calls, with the shared prefix split equally over 192 candidate--task pairs per schedule.}
\label{hv:tab:rsi-graph-case}
\end{table}

\paragraph{First acceptance: retain decisions across feedback.}
The first batch proposes single-use execution, persistent reuse with a limit of 3, and revised native instructions. Complete training assessments select proposal 1b, reducing execution cost from 11.11 to 5.05 Tokens/task (k). In the fixed checkbox example, the root clicks \texttt{6p}; four authored nodes then click \texttt{nUoKN}, \texttt{ute}, \texttt{Q7} and Submit, in that order. Each node depends on the preceding native receipt and has its own screenshot region. Persistent execution certifies the first three future actions. The reuse limit returns Submit to a model callback, so the five-action task uses two model calls and three substitutions.

\paragraph{Second acceptance: reuse the final decision.}
Depth proposes three children of the accepted harness: reuse limits of 4 and 2, and a refinement at limit 3 that requires a nonempty continuation. It accepts D2a, the limit-4 child. The selected harness costs 4.66 Tokens/task (k), a further 7.77\% reduction from the first acceptance on the same 32 training tasks. In the example, the authored root, four action nodes, predecessor edges, literal arguments and visual regions are identical across the two accepted harnesses. The accepted change is the execution horizon: Submit now follows the receipt for \texttt{Q7} through its visual guard. This removes the second model callback and its separately authored Submit graph. The task finishes with one model call and four certified substitutions. Figure~\ref{hv:fig:rsi-graph-case}(b) highlights this newly reused boundary.

\paragraph{Comparison with further proposals from the initial harness.}
Breadth spends its second three-candidate batch on children of the initial harness, proposing persistent reuse limits of 4, 2 and 3. It also selects limit 4 and reaches the same one-call execution path. Both schedules solve all 32 training tasks; their final costs both round to 4.66 Tokens/task (k). Recursive refinement therefore produces a concrete additional improvement after the first acceptance, while this short-workflow comparison yields the same final structure for both schedules. Table~\ref{hv:tab:rsi-graph-case} reports all proposal and candidate-assessment costs separately from selected-harness execution. The comparison establishes the benefit of the second accepted edit within the Depth lineage; a general advantage over Breadth requires a larger difference under the same search budget.

\subsection{LongForm-UI: controlled long-operation-chain benchmark}
\label{hv:app:longform}
LongForm-UI is a custom native-browser benchmark that isolates persistent reuse of visible literal actions. It contains 20 semantic form schemas and eight deterministic instances per schema. Four schemas supply 32 training tasks, four supply 32 validation tasks and twelve supply 96 test tasks. Entire schemas remain disjoint. Structural stratification assigns training schema counts of two/one/one to forms with 8/10/12 fields, validation counts of one/two/one, and test counts of four/four/four. All memberships and seeds are fixed before actor execution. Each instance requires at least one native type operation per field and a native Submit, or 9, 11 or 13 operations. No outcome determines inclusion.

Each required literal is visibly printed beside its blank field. Native Submit succeeds only when every field exactly matches its visible target; premature submission, incorrect values and incomplete episodes count as unsuccessful. The released MiniWoB core transports rewards, while the form generator and exact conjunction evaluator are custom.

The common browser viewport is 320 by 440 pixels, rendered at scale three. Every arm uses the full 960-by-1320-pixel PNG, public action history and the same pixel-only geometry hints. There is no DOM, accessibility, HTML, expected-value or evaluator input to the actor or guard. The Qwen3.5-9B, prompts, constrained schema, native mouse/keyboard, 15-action limit, 180-second allowance, response capacity, seed and routing are fixed within each comparison. An atomic type action focuses and replaces one field. Graphs in the five-arm persistence study contain at most eight future actions. All action arguments are frozen at authorship; exact patch guards and native receipts determine later eligibility. The two graph conditions allow either one future action per graph or repeated execution of its stored actions. The common parent differs from the OCR-backed MiniWoB++ training study above. The geometry-based harness is transferred from MiniWoB++: all five method modules, the authoring prompt, graph schema, coordinate compiler and reuse logic remain unchanged; only the browser viewport and native task entry point change. Section~\ref{hv:sec:cross-benchmark} analyzes this transfer using the same test executions. The target training and validation stages evaluate a single fixed source configuration without parameter updates.

Training evaluates the single-use and Full pair. The prespecified capability gate requires at least 24 Full successes out of 32, no paired success loss, positive complete saving and real second-and-later native reuse. Protected validation selects the one training-frozen candidate with the same paired quality/cost/reuse conditions. Selection freezes before the complete five-arm test, which retains every outcome. Thirty-percent complete saving is the substantial-saving objective. No protected result supplies design changes.

The training comparison solves 3 and 27 of 32 tasks for single-use and Full, respectively. Complete costs are 17.79 and 5.14 Tokens/task (k), saving 71.11\%.

The validation comparison solves 2 and 24 of 32 tasks for single-use and Full, respectively. Complete costs are 19.98 and 4.91 Tokens/task (k), saving 75.42\%.

On all 96 test tasks, Full costs 5.01 versus 18.90 Tokens/task (k), saving 73.49\%, at 78.13\% versus 8.33\% native success. Full executes 869 certified omissions, including 698 beyond the first reuse; maximum observed reuse depth is 8. The paired schema-cluster 95\% saving interval is [70.04, 77.34]\%, using 10,000 resamples of the twelve complete test schema groups with seed 20261019. Every resampled schema retains all eight instances. This interval conditions on the controlled schema population.

\begin{table}[t]\centering\small\caption{LongForm-UI by required chain length: four schema groups and 32 test tasks per row and arm.}\begin{tabular}{rrrrr}\toprule Required actions & \multicolumn{2}{c}{Success (\%)} & \multicolumn{2}{c}{Tokens/task (k)} \\ & Single & Full & Single & Full \\\midrule
9 & 18.75 & 100.00 & 16.08 & 2.74 \\
11 & 0.00 & 84.38 & 19.12 & 5.83 \\
13 & 6.25 & 50.00 & 21.51 & 6.46 \\
\bottomrule\end{tabular}\end{table}

Token totals include all actor input and output, authorship, control, retained callbacks, fallback and unsuccessful executions; every task remains in the denominator. Raw provider usage, native screenshots and actions, exact field rewards, authored graphs and every omission certificate are independently reconstructed. Paired arms begin with matching inputs and images, then decode their own authored actions. Savings use complete measured arm totals, and verification runs entirely on the recorded data.

For the controlled harness comparison in Table~\ref{hv:tab:longform-harnesses}, the separately training-selected HarnessMM-V configuration authors up to twelve ordered literal successors, covering a fully visible form in one graph. Figure~\ref{hv:fig:quality} retains the original eight-successor persistence configuration. The author-time compiler assigns sequential dependencies and fixed pixel patches; the actor supplies every literal action. The compiler and guard operate on the supplied screenshots and public receipts; the hidden evaluator scores task completion separately. The selected configuration solves all 32 training, 32 validation and 96 test tasks, with 957 certified test callback omissions. Initial test screenshots and the native exact-field scorer match across harnesses.

All twenty semantic schemas share one static two-column layout, with every target value visible before execution. The test population comprises twelve schemas with eight instances each. Both HarnessMM-V and DeepSeek Harness complete all 96 form-filling tasks under this shared layout and exact-field evaluator.

Table~\ref{hv:tab:longform-harnesses} uses one complete selected-pipeline cost scope for every method: test execution, the entire selected training and validation phases, optimizer/native learning rollouts and allocated setup. Each physical request is charged once, and the denominator is all 96 test tasks. HarnessMM-V's setup allocation includes training and validation exactly once. Test execution alone costs 2.35 Tokens/task (k); its complete selected-pipeline cost is 3.97. The formal benchmark panels use their stated execution and method-learning/setup accounting.

\begin{table}[t]
\centering\small
\caption{LongForm-UI harnesses on 96 test tasks; costs include the complete selected pipeline.}
\label{hv:tab:longform-harnesses}
\begin{tabular}{lrr}
\toprule Method & Exact successes & Tokens/task (k) \\
\midrule
baseline harness & 15/96 & 51.97 \\
SkillOpt & 14/96 & 106.67 \\
AutoSaddler & 13/96 & 113.47 \\
SoL-Pi & 93/96 & 245.61 \\
Prompt Compaction & 21/96 & 51.54 \\
DeepSeek Harness & 96/96 & 3.93 \\
HarnessMM-V & 96/96 & 3.97 \\
\bottomrule
\end{tabular}
\end{table}

The native DeepSeek Harness configuration uses Qwen3.5-9B in the official multimodal loop and an explicit command codec. The model supplies visible fills and final Submit as ordered tuples. Typed fill, click and sequence tools preserve all five native operations. A concise sequence description, compact sequence schema and omission of the generic SDK preamble reduce repeated context. The schema advertises an array of command arrays; the original decoder still checks each opcode, argument type, arity, coordinate and 15-action limit. The policy retains visibility checks, uncertainty handling and the complete-task token-efficiency objective. The decoder normalizes exact fill/F/click aliases and JSON-encoded command arrays. If complete inner tuples lack only the outer closing bracket, it appends that bracket, inferring no action, target or literal. Original primitive checks remain. Two consecutive rejections without native progress end the turn. Each executed primitive retains its native receipt and full-resolution screenshot; public history and geometry remain exactly reconstructable. The configuration freezes before selection and testing, preserving the actor, seed, tasks, 15-action allowance and exact-field scorer. It reaches 96/96 exact successes. Test execution costs 2.34 Tokens/task (k); selected training, validation and allocated setup bring the complete cost to 3.93.

RHO and HarnessPro each complete eight optimization rounds on the same 32 training forms, followed by 32 validation forms and all 96 test forms. RHO retains its released optimization core. HarnessPro uses the configured visual reproduction with tiered optimization and a generated-test DSL. Both retain the native actions, screenshots, scorer and episode limits. Table~\ref{hv:tab:longform-rho-execution} compares test execution alone; training, optimization and validation are accounted for separately from these costs.

\begin{table}[t]
\centering\small
\caption{Qwen3.5-9B on the same 96 LongForm-UI test tasks. Costs include every test-time actor call and exclude learning and verification.}
\label{hv:tab:longform-rho-execution}
\begin{tabular}{lrr}
\toprule Method & Success (\%) & Tokens/task (k) \\
\midrule
Baseline harness & 15.63 & 31.28 \\
RHO & 51.04 & 31.44 \\
HarnessPro & 44.79 & 33.03 \\
HarnessMM-V & \textbf{100.00} & \textbf{2.35} \\
\bottomrule
\end{tabular}
\end{table}

\paragraph{Progressive forms.}
A distinct controlled workload reveals fields over three stages. Later target values become visible only after native Next clicks; deterministic shuffling changes field positions and switches between one and two columns. Its twenty semantic groups retain the predetermined 32 training, 32 validation and 96 test instances, with disjoint groups and new task identities. All three arms share the same full-resolution screenshots, native scorer, common actor and sampling, 24-action allowance and 240-second limit. This workload and the selected configurations freeze before protected evaluation.

Table~\ref{hv:tab:longform-progressive} reports every test outcome and complete selected-pipeline input-plus-output cost, including training, validation and allocated setup. The unchanged HarnessMM-V configuration authors visible literal successors within the current stage; new stage observations require fresh grounding. HarnessMM-V achieves 96/96 successes by grounding each newly revealed stage and reusing the visible actions within it. The static and progressive form studies test these execution mechanisms alongside the formal visual benchmarks.

\begin{table}[t]
\centering\small
\caption{LongForm-UI Progressive on 96 test tasks per method, including selected training, validation, testing and setup costs.}
\label{hv:tab:longform-progressive}
\begin{tabular}{lrr}
\toprule Method & Exact successes & Tokens/task (k) \\
\midrule
baseline harness & 84/96 & 49.94 \\
DeepSeek Harness & 96/96 & 188.76 \\
HarnessMM-V & 96/96 & 10.85 \\
\bottomrule
\end{tabular}
\end{table}

\subsection{GPT-5.6 Sol on LongForm-UI}
The same 96 test forms and full-resolution screenshots are evaluated with GPT-5.6 Sol high as the actor. The selected harnesses and prompts are frozen before testing; the native exact-field scorer, memberships, action limits and task seeds remain unchanged. Table~\ref{hv:tab:gpt-sol-longform-test} includes every test-time actor call. Training and shared calibration are reported in Table~\ref{hv:tab:gpt-sol-training}.
\begin{table}[t]
\centering\small
\caption{GPT-5.6 Sol high on 96 LongForm-UI test tasks, including all test-time actor input/output costs.}
\label{hv:tab:gpt-sol-longform-test}
\begin{tabular}{lrr}
\toprule
Method & Success (\%) & Tokens/task (k) \\
\midrule
Baseline harness & 86.46 & 63.37 \\
Prompt Compaction & 59.38 & 71.82 \\
HarnessMM-V & \textbf{100.00} & \textbf{10.30} \\
\bottomrule
\end{tabular}
\end{table}

\subsection{Extended visual mechanism measurements}
\label{hv:app:visual-measurements}
\subsubsection{Execution-certificate verification}
The independent audit recomputes reference and observed ROI signatures from actual frames, reconstructs the authored graph, matches literal arguments to native receipts, checks predecessor order, and verifies omitted callbacks against the complete physical ledger. Each certificate identifies its graph instance and native receipt; requests from separately authored graphs are counted individually. Authorship and fallback remain charged.

A certificate verifies native execution and omission of the registered callback; native task scoring measures success. The retained native trajectory example links actual screenshot anchors and receipts to a fixed Search click. Section~\ref{hv:sec:visual-contract} specifies the visual sufficiency condition used for reuse.

\subsubsection{Coverage, persistence, and fallback}
We count authored graphs, eligible boundaries, substitutions and paid fallbacks over every task. Reuse depth counts executed nodes within each actual graph instance.

The earlier fixed-coordinate training comparison demonstrates three continuations within one authored graph. The complete prospective persistence comparisons in Appendix~\ref{hv:app:persistence-train} measure how repeated graph execution reduces cost under shared screenshot inputs.

Rejected ROIs request fresh slow actions. Target correction uses fresh model reasoning with its full cost included.

\subsubsection{Model calls, tokens and verifier computation}
The following measurements describe the earlier fixed-coordinate MiniWoB++ comparison, retained separately from the common-geometry comparison in Figure~\ref{hv:fig:quality}. The retained-callback control and full HarnessMM-V reach 40.63\% and 40.63\% success. They make 464 and 452 slow calls; the difference is 12, matching the 12 full-method certificates. Tokens/task (k) decrease from 3.89 to 3.81, a 2.18\% reduction. This control isolates actual callback removal within the declared graph-execution route. Its retained decision is deliberately charged before executing the graph action.

Full HarnessMM-V authors 89 valid graphs across 96 tasks and executes 12 substitutions, or 0.13 per task. Its 45 paid fallback decisions include 7 malformed-author outputs. Author-time target verification and runtime pixel checking together take 0.14 seconds across the full population and make no model requests. Mean episode duration is 3.55 seconds, versus 4.22 for the baseline harness. These durations include service scheduling, model calls and actual action sequences.

Image requests decrease from 534 to 452 at the shared 480 by 630 pixel resolution. Provider usage measures aggregate multimodal prompt tokens without a separate image-token component. Mean model response latency is 0.55 seconds for Full and 0.56 for the baseline harness. Screenshot counts and model latency complement the provider-measured input and output costs.

\subsubsection{Visual perturbation and verifier coverage}
We replay the 23 certified training boundaries with predetermined pixel interventions, without dispatching any action to a perturbed environment. The unchanged bound ROI passes at all 23 boundaries. Inverting the ROI or covering it with black causes all 23 checks to reject. A three-pixel horizontal image translation passes three checks. Whole-image equality passes only three of the original 23 observations, compared with 23 for the local ROI rule. These retained-frame replays measure the response of the verifier to each intervention. Native task-success scores and callback substitutions are measured separately during execution.

\subsubsection{MiMo visual-actor training results}
A separately frozen MiMo-V2.6-Pro variant completes all five controls on 24 training tasks. The baseline harness and full HarnessMM-V success are 66.67\% and 100.00\%, with 1.64 versus 1.14 Tokens/task (k), saving 30.32\%. Appendix~\ref{hv:app:api} gives the complete controls, omission counts and image-token protocol.

\subsubsection{ScreenSpot-Pro training and validation}
\label{hv:app:pro-training-validation}
The configured screenshot-grounding adapter uses frozen Qwen3.5-9B weights and PP-OCRv5 server detection and recognition from PaddleOCR 3.3.2~\cite{hv:cui2025paddleocr30technicalreport}. OCR reads overlapping tiles of the original pixels. The actor compiles each instruction into up to four likely visible labels, then selects from at most 12 supplied OCR label identifiers. Both configurations require recognition confidence of at least 0.90 and an explicit decision that the selected text itself is the clickable target. The text guard also requires that text to occur once in the screenshot. Label-ID binding accepts the one supplied identifier selected by the actor when its decision is unambiguous. The compiler submits the box center in original-image coordinates; other targets use the native full-image prediction. These configurations evaluate single-point grounding.

Exact RGB copies form one image group before task assignment. The custom partition retains all 1,581 tasks, assigning 327 to training, 322 to validation and 932 to testing, with each image group confined to one partition. The table covers complete training and validation. Each configuration's query, selection and binding rules are frozen from training before its protected validation evaluation. Native point accuracy is the fraction of predictions inside the annotated target rectangle.

Table~\ref{hv:tab:pro-training-validation} includes native training verification for both methods, every training query and label-selection call, and all validation, invalid-output and native-fallback calls. Shared physical calls are charged once within each method. The denominator is all 322 validation tasks. This is the selected pipeline's complete training and validation cost; independent exploratory methods retain separate costs under the fixed study scope. OCR inference and compute time are recorded separately from language-model tokens. Within validation execution, label-ID binding uses 4.70 versus 6.10 Tokens/task (k), saving 23.00\%; its complete stage saving is 8.90\%.

\begin{table}[t]
\caption{ScreenSpot-Pro with Qwen3.5-9B. Costs include selected training and validation, normalized by 322 validation tasks; Text/ID identify label-binding rules.}
\label{hv:tab:pro-training-validation}
\centering\footnotesize
\setlength{\tabcolsep}{3pt}
\begin{tabular}{@{}lrrr@{}}
\toprule
Method & \shortstack[r]{Training\\(\%)} & \shortstack[r]{Validation\\(\%)} & \shortstack[r]{Tokens/task\\(k)} \\
\midrule
Baseline harness & 41.59 & 51.55 & 12.39 \\
HarnessMM-V (text) & \textbf{42.20} & \textbf{53.11} & 11.49 \\
HarnessMM-V (ID) & 41.90 & 52.17 & \textbf{11.29} \\
\bottomrule
\end{tabular}
\end{table}

\subsubsection{ScreenSpot-Pro full evaluation}
\label{hv:app:pro-full-test}
The grounding adapter keeps Qwen3.5-9B, original screenshots and native point-in-box scoring. A training-derived grammar extracts the object name after removing action verbs, articles, interface suffixes and location context. An exact OCR match supplies its box center at confidence 0.98, requiring uniqueness among labels at confidence 0.90. Menu names instead require one occurrence at confidence 0.90 supported by at least three similarly sized, aligned OCR rows above and below. Adding or inserting a named node uses its object name. Other requests retain the unique-label selector and native full-image fallback. Floor maps predicted coordinates to integer screenshot pixels. No annotation is used at runtime.

Identical RGB screenshots stay in one partition: 327 training, 322 validation and 932 test tasks. Training fixes the method before validation and testing. Validation accuracy is 53.11\%, versus 51.55\% for the baseline. On every test task, accuracy rises from 47.53\% to 47.96\% and complete cost falls by 20.87\% (Table~\ref{hv:tab:pro-full-test}). Menu alignment handles 15 test instructions. Integer-pixel decoding contributes four correct predictions to the earlier literal-grounding variant.

Complete pipeline costs include native training calibration, authorship, optimization diagnostics, candidate selection, validation and test execution, including control, invalid-output and fallback calls. Each physical request is charged once within each method, and both costs use all 932 test tasks as the denominator. OCR and rule computation are recorded separately from language-model tokens. Independent exploratory studies retain separate costs under the fixed study scope.

\begin{table}[t]
\centering\small
\caption{Qwen3.5-9B on 932 ScreenSpot-Pro test tasks, including selected training, optimization, validation and testing costs.}
\label{hv:tab:pro-full-test}
\begin{tabular}{@{}lrr@{}}
\toprule
Method & Accuracy (\%) & Tokens/task (k) \\
\midrule
Baseline harness & 47.53 & 10.06 \\
Unique-label adapter & 47.53 & 8.14 \\
Literal grounding & 47.53 & 7.98 \\
Integer-point grounding & 47.96 & 7.98 \\
HarnessMM-V & \textbf{47.96} & \textbf{7.96} \\
\bottomrule
\end{tabular}
\end{table}

\subsubsection{GPT callback ablation}
\label{hv:app:gpt-callback}
The callback ablation keeps graph authorship and native graph execution while calling the slow model before every reusable action. GPT-5.6 Sol high executes all 32 MiniWoB++ training tasks in each condition. All four conditions succeed on every task (Table~\ref{hv:tab:gpt-callback}). Both graph configurations execute 47 reusable nodes. The ablated configuration makes 47 additional native callbacks; HarnessMM-V certifies their omission and uses 32 physical model calls in total. Sixteen omissions are second or later reuses of an authored graph.

Removing these callbacks reduces training execution cost from 3.06 to 1.30 Tokens/task (k), saving 57.49\%. A paired bootstrap with 10,000 draws over the four original task families gives a descriptive 95\% interval of 48.11--62.80\% for this saving. The interval describes variation within the training families. Costs include all actor input and output within each named arm, including authoring, invalid responses and fallbacks. Harness RSI and protected-stage costs are separate. Historical execution concurrency differs across conditions, so episode time is recorded separately.

\begin{table}[t]
\caption{GPT-5.6 Sol high callback ablation on 32 MiniWoB++ training tasks per condition.}
\label{hv:tab:gpt-callback}
\centering\footnotesize
\setlength{\tabcolsep}{3pt}
\begin{tabular}{@{}lrrr@{}}
\toprule
Method & \shortstack[r]{Success\\(\%)} & \shortstack[r]{Tokens/task\\(k)} & Calls \\
\midrule
Baseline harness & 100.00 & 2.94 & 79 \\
Prompt Compaction & 100.00 & 2.93 & 79 \\
Graph with native callback & 100.00 & 3.06 & 79 \\
HarnessMM-V & 100.00 & 1.30 & 32 \\
\bottomrule
\end{tabular}
\end{table}

\subsubsection{Verifier response across two benchmarks}
\label{hv:app:gpt-guard-interventions}
We evaluate nine predetermined interventions at every certified reuse boundary in the complete training trajectories: 47 with GPT-5.6 Sol on MiniWoB++ and 288 with GPT-6.1 Sol on LongForm-UI. The resulting 3,015 checks use the original verifier and issue no model calls or native actions. All unchanged frames pass. Changing one bound pixel or inverting the bound ROI rejects every check; changing only exterior pixels preserves every acceptance (Table~\ref{hv:tab:gpt-guard-interventions}). Missing predecessor receipts and changed image resolution also reject every check.

These retained-frame checks measure the verifier's pixel and receipt conditions. Translation can preserve a uniform bound patch, as reflected by the accepted translations. Native task success and callback omissions are measured during the original executions.

\begin{table}[t]
\caption{Certified training-boundary checks: GPT-5.6 Sol on MiniWoB++; GPT-6.1 Sol on LongForm-UI.}
\label{hv:tab:gpt-guard-interventions}
\centering\footnotesize
\setlength{\tabcolsep}{3pt}
\begin{tabular}{@{}lrr@{}}
\toprule
Condition & MiniWoB++ & LongForm-UI \\
\midrule
Original frame & 47/47 & 288/288 \\
One ROI pixel changed & 0/47 & 0/288 \\
ROI inverted & 0/47 & 0/288 \\
One exterior pixel changed & 47/47 & 288/288 \\
Exterior inverted & 47/47 & 288/288 \\
Translation (3 px) & 16/47 & 256/288 \\
Translation (8 px) & 8/47 & 256/288 \\
Changed resolution & 0/47 & 0/288 \\
Missing predecessor receipt & 0/47 & 0/288 \\
\bottomrule
\end{tabular}
\end{table}

\section{MiMo visual-actor training results}
\label{hv:app:api}
A separate MiMo-V2.6-Pro variant was frozen using training evidence before the Qwen3.5-9B test opened. On the same 24-task development set, its baseline harness reaches 66.67\% success at 1.64 Tokens/task (k), while full HarnessMM-V reaches 100.00\% at 1.14, saving 30.32\%. Authorship-only and single-use controls both reach 91.67\%, at 2.38 and 1.43 Tokens/task (k). The retained-callback control reaches 100.00\% at 2.08. It makes 60 calls versus 29 for full execution, matching 31 certified omissions. Full and single-use produce 31 and 20 substitutions, respectively. These training comparisons separate graph authorship, persistence and callback removal. Qwen and MiMo results are reported separately.

The API variant adds a Boolean pixel-support check at already authored click points and stores the compiled native command plan in the certificate. Its shared root interface also supports literal drag endpoints. All five arms share that interface and the same budgets. Provider usage identifies the image-token component exactly. The baseline harness and full HarnessMM-V execution issue 56 and 29 image requests; the full guard uses 0.10 seconds and no model calls. The timing summary omits one missing single-use wall/guard measurement; complete actor usage and task outcomes include that episode. Full-method timings cover every episode. This protocol variant uses its own protected validation and test populations.

\section{Verifier computation}
The prototype hashes the whole screenshot in each unresolved-node check in addition to the ROI. With $m$ unresolved nodes, screenshot area $|I|$ and ROI areas $|R_i|$, pixel work costs $O(m|I|+\sum_i|R_i|)$. Its current eligibility list scan can additionally require $O(m^2)$ comparisons; the graph cap keeps $m\leq8$. These bounds follow from the implemented hash checks and list scan. Author graphs contain at most eight future nodes; whole-image hashes identify source frames, and ROI hashes determine eligibility. Full-image verification is a separately identifiable ablation because unrelated visual changes can reject an otherwise stable target.

\section{Native VisualWebArena evaluation}
\label{hv:app:native-vwa}
The population contains all 910 released tasks across shopping, classifieds and Reddit. A task-level custom partition fixes 182 training, 182 validation and 546 test tasks before optimization, with seed 20261013. Each arm executes each task once. The three sets are disjoint; no validation or test answer, trajectory or outcome supplies method revisions. Validation selects only the training-frozen method, requiring no native success loss, at least 20\% complete token saving and actual certified omissions. Testing begins after selection is frozen, and reports the complete selected comparison regardless of its final quality--cost trade-off.

Both arms receive the complete native RGB screenshot, the original task instruction and public action history, together with all supplied reference images. The default viewport is 1280 by 2048; tasks with registered native viewport sizes retain those original dimensions. Actor images are neither cropped nor resized by the harness. The baseline harness generates a fresh action at each decision boundary. HarnessMM-V uses losslessly encoded complete public history and an author that fixes a root and at most two ordered, distinct visible click/type successors. Other original root actions remain available. When no stored node passes the checks, the next slow call can author a new graph.

AutoSaddler uses its official two-round optimization engine on the training partition, with GPT-5.6 Sol high as optimizer and the common Qwen3.5-9B native visual actor. The selected instruction skill is frozen before all 182 validation tasks and all 546 test tasks. Its complete cost includes every native learning rollout, optimizer request, selected validation call, calibration allocation and test execution. It reaches 81/546 native successes at 156.06 Tokens/task (k).

Prompt Compaction retains its fixed history-summary policy with the common native actor, full current and reference images, native viewport, tools and scoring. Its complete selected validation and test costs include every summary request and native semantic-judge call. On all 546 test tasks it reaches 75 native successes at 95.17 Tokens/task (k).

Before the root executes, a deterministic pixel compiler binds each fixed future point to the smallest visible control footprint, or a fixed local patch of radius 24 pixels when no control boundary is found. The compiler derives these regions from the model-authored points. Coordinates and text remain fixed after authorship. The guard checks predecessor receipts, image dimensions, the bound public context and the same fixed region. It accepts exact retained pixels or the frozen achromatic edge-and-literal-ink signature for clicks. A changed scene, unsupported target, held input state or failed appearance check returns to the visual actor; the next model call resolves any new target or argument. This web configuration differs from the single-author, exact-ROI MiniWoB++ controls while retaining literal closure and callback certification.

The native interface keeps all 15 action types, supplied-image upload, normalized-to-original-pixel decoding and a maximum of 30 native actions. Both conditions share the same fixed routing, native login and database reset contract. Every episode starts from its independently reset website state; the same task pair uses the same admitted actor route. Generation uses temperature zero, seed 20261013 and the same 2048-token output allowance. Cumulative token accounting has no termination ceiling. Native scoring retains released deterministic and image metrics, including the BLIP2-Flan-T5-XL captioning function; reported test semantic judgments use the same blind GPT-5.6 Sol high configuration across all methods, with the native prompts and label rules. The training-derived harnesses remain frozen. Native scores remain separate from guard acceptance.

Tokens/task (k) sums all physical actor and semantic-judge input and output tokens in the displayed arm, including graph authors, paid parser fallbacks, controls and unsuccessful tasks, then divides by all 546 test tasks and by 1000. Native evaluator computation and visual-verifier CPU work are separate from this token measure. All semantic grading calls are charged to their method, including the common semantic evaluation. Every counted omission has a successful literal native receipt and unchanged slow-call counters, with no physical request bearing its registered callback identifier. The independent audit checks all original requests, usage, images, graph instances, route bindings and native scoring receipts across all 1092 test executions. Complete raw measurements and exact reproduction identities remain in the external evidence package.

\begin{figure*}[t]
\centering
\includegraphics[width=.43\textwidth,height=.69\textheight,keepaspectratio]{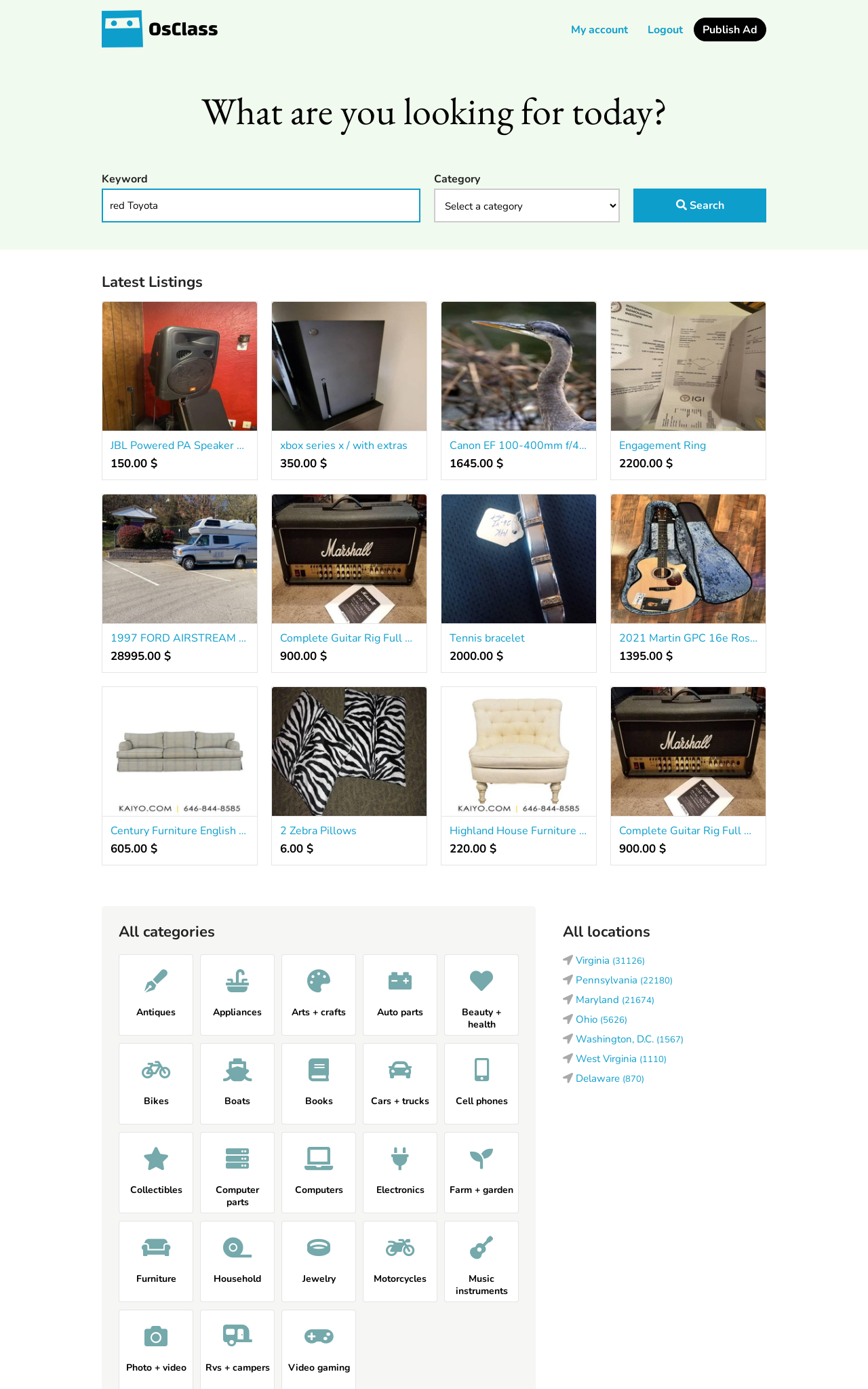}\hfill
\includegraphics[width=.43\textwidth,height=.69\textheight,keepaspectratio]{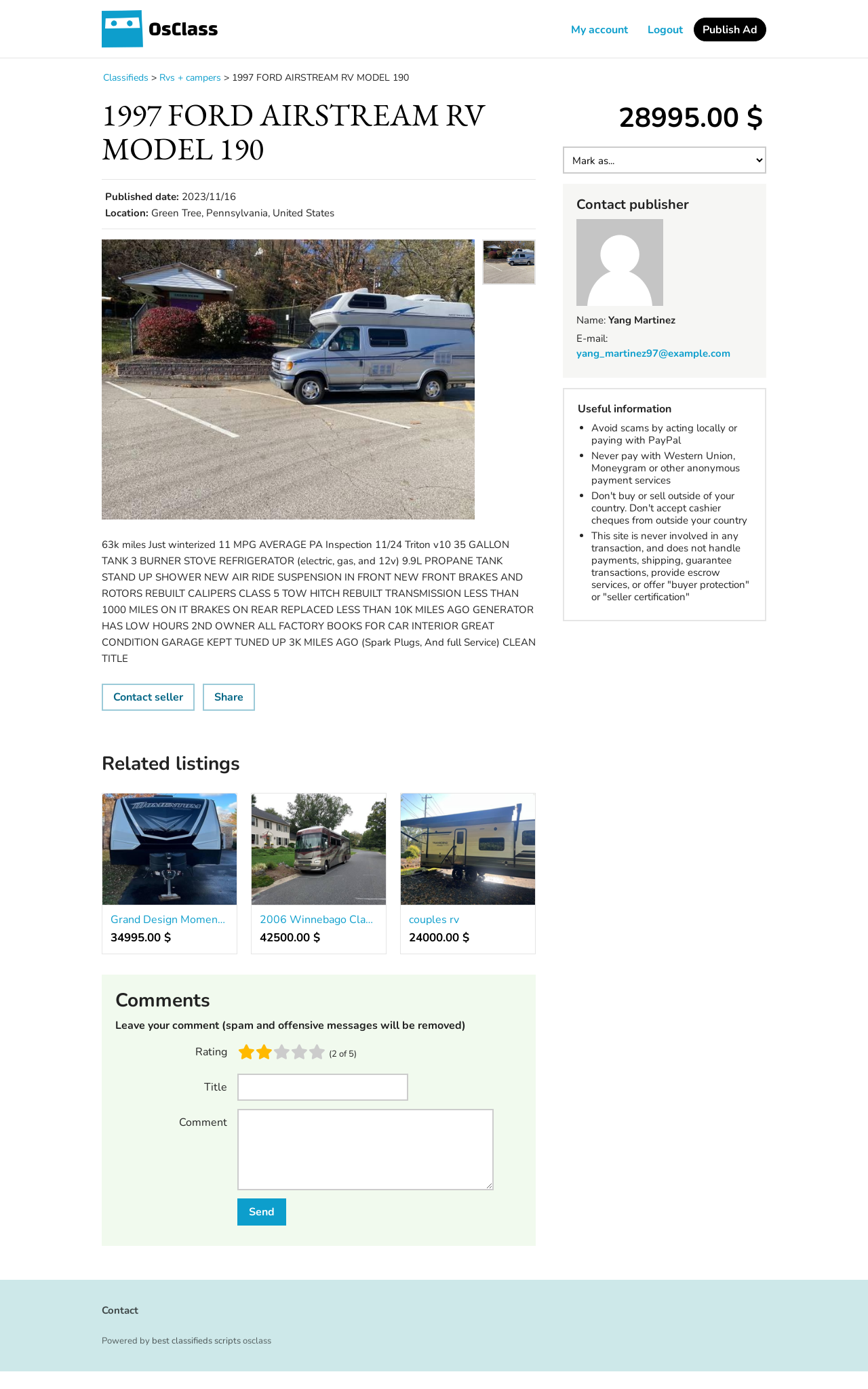}
\caption{VisualWebArena screenshots before two certified actions, with fixed author-bound regions and unchanged literal arguments.}
\label{hv:fig:native-vwa-observations}
\end{figure*}

\section{Visual RSI comparisons on single-image grounding}
The ScreenSpot-v2 and ScreenSpot-Pro comparisons~\cite{hv:seeclick,hv:screenspotpro} retain fixed task-level partitions: the former has 246 training, 252 validation and 774 test tasks, and the latter has 316 training, 316 validation and 949 test tasks. Every method receives the full original screenshot and instruction, uses the common Qwen3.5-9B actor, and submits one point for the native box scorer. Official SkillOpt and AutoSaddler learn instruction skills from training feedback; validation evaluates the frozen candidate before testing. DeepSeek Harness uses its official multimodal runtime. Prompt Compaction has no history to summarize and records zero summary triggers; its original native actor and input are identical to baseline harness. For every method, complete selected-pipeline costs include training and validation verification, optimizer-model calls, native learning rollouts, calibration, all test calls, controls, fallbacks and unsuccessful executions. Each complete cost is divided by the full test population and by 1000. HarnessMM-V binds the authored fractional point to a singleton graph root and submits that exact point once, with no future node, GUI execution or callback omission. These comparisons evaluate single-image point grounding; the interactive benchmarks evaluate multi-step callback reuse.

\paragraph{DeepSeek Harness on VisualWebArena.}
The native multimodal harness uses the same Qwen3.5-9B backbone, complete screenshots, native task actions, 30-action allowance and GPT-5.6 Sol high semantic scorer. Its frozen strategy is assessed on 182 validation tasks before the full 546-task test population. It solves 29 test tasks (5.31\%) at 494.53 Tokens/task (k). This cost includes every measured test actor and semantic-scoring call, including unsuccessful executions and interrupted attempts. Optimization, training and validation are outside this execution-cost comparison.

\subsection{Additional experimental settings}
\label{hv:app:experiment-settings}
SkillOpt~\cite{hv:yang2026skillopt} and AutoSaddler~\cite{hv:park2026autosaddler} use their official optimizers with GPT-5.6 Sol high proposers and Qwen3.5-9B actors. RHO~\cite{hv:pan2026rho} retains its released label-free optimizer with the same actor and proposer. SoL-Pi and DeepSeek Harness retain their official multimodal loops and native tools. Prompt Compaction summarizes public history while retaining the current screenshot and actions. Appendix~\ref{hv:app:miniwob-protocol} details the visual runtimes; Appendix~\ref{hv:app:rsi-common-start} describes optimizer adapters and candidate selection.

The Qwen3.5-4B HumanEval DeepSeek comparison has an exact McNemar $p=0.0161$ for the paired success difference. The MBPP-sanitized DeepSeek comparison records 472 certified bypasses.

The ALFWorld instruction ablation uses history length two, a 50-step limit and seeds 9401--9403, covering 2,133 tasks. Neutral text matches the instruction token count. The four-condition MBPP-sanitized ablation uses one attempt per task and retains graph authorship when EDG execution is disabled.

\subsection{GPT-5.6 Sol training and transfer measurements}
\label{hv:app:gpt-training-transfer}
On all 32 training tasks per benchmark, HarnessMM-V solves every task while saving 58.46\% on MiniWoB++ and 84.15\% on LongForm-UI. Training costs include episode calls and model/vision calibration allocated once across the six benchmark--method groups; optimization and selection are separate. Transfer costs include every target test call, with source optimization and target training/validation accounted for separately.

\begin{table}[t]
\centering
\caption{GPT-5.6 Sol high: 32 training tasks per benchmark and 96 target test tasks per transfer direction. Costs include all episode calls and shared training calibration; optimization/selection and target training/validation are separate.}
\label{hv:tab:gpt-sol-training}
\small
\setlength{\tabcolsep}{4pt}
\begin{tabular}{lrr}
\toprule
Method & Success (\%) & Tokens/task (k) \\
\midrule
\multicolumn{3}{l}{\textit{MiniWoB++ training}} \\
Baseline harness & 100.00 & 10.74 \\
Prompt Compaction & 100.00 & 10.74 \\
HarnessMM-V & 100.00 & \textbf{4.46} \\
\midrule
\multicolumn{3}{l}{\textit{LongForm-UI training}} \\
Baseline harness & 96.88 & 58.39 \\
Prompt Compaction & 96.88 & 62.68 \\
HarnessMM-V & 100.00 & \textbf{9.25} \\
\midrule
\multicolumn{3}{l}{\textit{HarnessMM-V transfer: source $\rightarrow$ target}} \\
MiniWoB++ $\rightarrow$ LongForm-UI & 98.96 & 6.08 \\
LongForm-UI $\rightarrow$ MiniWoB++ & 86.46 & 2.13 \\
\bottomrule
\end{tabular}
\end{table}

\subsection{Transfer configurations and uncertainty}
\label{hv:app:transfer-settings}
Qwen3.5-9B transfers from MiniWoB++ to LongForm-UI; GPT-5.6 Sol high evaluates both directions. The Qwen source fixes eight future nodes and uses geometry-based inputs, differing from the OCR-equipped main MiniWoB++ configuration. Visible fill-and-submit dependencies support reuse; guard rejection or graph exhaustion returns to fresh reasoning. Both conditions share the token-efficiency objective. The Qwen LongForm-UI comparison uses the same 96 target test tasks as Figure~\ref{hv:fig:quality}.

For GPT-5.6 Sol transfer to LongForm-UI, the paired schema-bootstrap 95\% interval for token saving is 81.61--84.76\%. The reverse transfer uses 161 model calls with 114 certified omissions, including 53 later graph uses. Every fallback call is included in target execution cost.

\section{HarnessMM-V coding-harness experiments}
\label{hv:app:reho}
\begin{table*}[t]
\centering\small
\setlength{\tabcolsep}{5pt}
\begin{tabular}{lllrrrrr}
\toprule
& & & & \multicolumn{2}{c}{Success (\%)} & \multicolumn{2}{c}{Tokens/task (k)} \\
Benchmark & Backbone & Harness & Tasks & Native & +HarnessMM-V & Native & +HarnessMM-V \\
\midrule
HumanEval & Qwen3.5-4B & Codex harness & 98 & 90.82 & 90.82 & 66.72 & 47.02 \\
HumanEval & Qwen3.5-4B & DeepSeek Harness & 98 & 71.43 & 85.71 & 96.28 & 11.62 \\
HumanEval & MiMo-V2.6-Pro & DeepSeek Harness & 98 & 96.94 & 97.96 & 11.62 & 4.40 \\
MBPP-sanitized & Qwen3.5-4B & DeepSeek Harness & 257 & 90.27 & 91.83 & 53.98 & 10.61 \\
\bottomrule\end{tabular}
\caption{Coding-harness comparisons on complete test populations, including all authoring, control and fallback costs. Visual results: Figure~\ref{hv:fig:quality}.}
\label{hv:tab:reho-background}
\end{table*}

\begin{table*}[t]
\centering
\small
\begin{tabular}{lllrrrr}
\toprule
Benchmark & Actor model & Harness & \multicolumn{2}{c}{Success (\%)} & \multicolumn{2}{c}{Tokens/task (k)} \\
& & & Native & +HarnessMM-V & Native & +HarnessMM-V \\
\midrule
HumanEval & Qwen3.5-4B & Codex harness & 90.82 & 90.82 & 66.72 & 47.02 \\
HumanEval & Qwen3.5-4B & DeepSeek Harness & 71.43 & 85.71 & 96.28 & 11.62 \\
MBPP-sanitized & Qwen3.5-4B & Codex harness & 91.83 & 90.66 & 129.55 & 49.22 \\
MBPP-sanitized & Qwen3.5-4B & DeepSeek Harness & 90.27 & 91.83 & 53.98 & 10.61 \\
HumanEval & MiMo-V2.6-Pro & Codex harness & 93.88 & 94.90 & 62.60 & 10.09 \\
HumanEval & MiMo-V2.6-Pro & DeepSeek Harness & 96.94 & 97.96 & 11.62 & 4.40 \\
MBPP-sanitized & MiMo-V2.6-Pro & Codex harness & 96.50 & 97.28 & 47.56 & 11.00 \\
MBPP-sanitized & MiMo-V2.6-Pro & DeepSeek Harness & 97.67 & 98.05 & 6.20 & 5.11 \\
\midrule
\multicolumn{7}{l}{Development: the same 32 GAIA2-mini scenarios per arm} \\
GAIA2-mini & DeepSeek-V4.1-Flash & Codex harness & 37.50 & 40.62 & 907.06 & 825.99 \\
GAIA2-mini & DeepSeek-V4.1-Flash & DeepSeek Harness & 40.62 & 46.88 & 891.75 & 750.53 \\
\midrule
\multicolumn{7}{l}{Test: the same 480 MiniWoB++ tasks per arm} \\
MiniWoB++ & GPT-5.6 Sol high & Codex harness & 87.50 & 88.13 & 31.25 & 29.59 \\
MiniWoB++ & GPT-5.6 Sol high & DeepSeek Harness & 98.33 & 98.54 & 16.31 & 15.45 \\
\bottomrule
\end{tabular}
\caption{Complete harness comparisons by actor and benchmark. Coding and MiniWoB++ rows use full test populations; GAIA2-mini uses 32 development scenarios. Costs include all execution calls; optimization/selection are separate.}
\label{hv:tab:api-backbone-values}
\end{table*}

\paragraph{MiniWoB++ protocol.}
Each harness pair uses the same 480 test tasks after selection on 160 training and 160 validation tasks. Both methods receive the token-efficiency objective. The DeepSeek configuration preserves the native action schema and conversation, encodes each public observation as a standalone state, and retains every original image selected by the runtime. Every test action is issued through the native model loop. Costs include all actor calls within each completed scored execution; development and selection are separate.

The HarnessMM-V results in Table~\ref{hv:tab:reho-background} evaluate decision reuse within existing coding harnesses. HarnessMM-V adds fixed visual anchors, image-bound eligibility checks and certificates that bind observed screenshots to native GUI actions. Each coding comparison uses its stated benchmark, actor, harness and selected implementation. The visual method is developed independently from visual training observations.

\paragraph{Populations and scoring.}
HumanEval uses its 164-task release with a custom partition of 33 training, 33 validation and 98 test tasks, assignment seed 20261006. MBPP-sanitized uses its 427-task release with 85 training, 85 validation and 257 test tasks, seed 20261005. Training and validation counts are rounded deterministically from one fifth of each complete release, and the remaining tasks form the test set. Each compared arm executes every test task once. Training develops the candidate, protected validation selects it, and the method and source are frozen before testing. HumanEval actors receive the specification, signature and docstring examples; reference solutions and private verifier tests stay outside their context. MBPP actors receive the common released public assertions; reference solutions remain excluded. Native functional tests determine single-attempt success. The common actor and verifier deadlines are 300 and 30 seconds.

\paragraph{Cost scope and integrated comparisons.}
All costs use complete actor input plus output divided by all test tasks in each comparison and 1,000. Shared author and control requests, fast decisions, finalizers, paid fallback and unsuccessful executions are charged. Training and selection costs are reported separately. Qwen3.5-4B remains fixed within each comparison that names this actor; MiMo-V2.6-Pro remains fixed within its HumanEval DeepSeek Harness pair. Codex harness and DeepSeek Harness identify the execution runtimes; both harnesses in each pair use the stated backbone. Savings are computed from measured complete arm totals. The MiniWoB++ pair uses all 480 predetermined test tasks per arm with native screenshot-based scoring; costs include every model call within each complete scored execution.

\paragraph{Persistence and callback ablation.}
Table~\ref{hv:tab:reho-ablation} gives the separately frozen four-condition HarnessMM-V study using DeepSeek Harness and Qwen3.5-4B on the same 257 MBPP-sanitized test tasks. All conditions retain the common literal-bank authoring interface, prompts, task limits and native scorer. The full method retains the graph across actions and removes eligible slow callbacks. Single-use execution discards a graph after its first substitution. The retained-callback condition preserves the registered slow call; disabling EDG prevents graph execution.
\begin{table}[t]
\centering\footnotesize
\setlength{\tabcolsep}{3pt}
\begin{tabular}{lrrr}
\toprule
Condition & Success (\%) & Tokens/task (k) & CCS \\
\midrule
EDG execution disabled & 95.33 & 16.19 & 0 \\
Single-use EDG & 93.00 & 18.03 & 212 \\
Callback retained & 93.39 & 16.37 & 0 \\
Full HarnessMM-V & 93.77 & 13.17 & 407 \\
\bottomrule\end{tabular}
\caption{Four-condition MBPP-sanitized ablation on 257 test tasks with DeepSeek Harness and Qwen3.5-4B. CCS counts certified callback substitutions.}
\label{hv:tab:reho-ablation}
\end{table}

Full HarnessMM-V saves 26.91\% against single-use graphs, with a paired adjusted interval of $[4.01,45.57]\%$, and 19.52\% against retained callbacks, with $[-2.07,38.68]\%$. It saves 18.62\% against disabled EDG execution. The respective success-difference intervals are $[-3.50,5.06]$, $[-2.72,3.50]$ and $[-5.45,2.33]$ percentage points. Joint task resampling uses 10,000 bootstrap draws and seed 20261009; the three predeclared contrasts use 98.33\% adjusted intervals. These intervals condition on the fixed population and one execution per task. Persistence's cost interval excludes zero; callback and disabled-execution cost intervals include zero. Certificates establish native execution and callback omission, while task quality is determined separately by the benchmark scorer.

\section{Experimental coverage and cost scope}
\label{hv:app:reho-complete}
Each visual and coding comparison states its benchmark, model, intervention, task population, evaluation phase and included costs. The HarnessMM-V label identifies each study's selected configuration. Complete actor execution costs include authorship, control, finalization, fallback and unsuccessful executions across the entire declared population. Selected-pipeline costs additionally include the measured learning and selection stages where specified. Full RSI research costs include every evaluated proposal and historical research expenditure within the registered scope. Execution savings compare episode costs. Net RSI savings compare the same workload after including all declared optimization and research costs.

The external reproduction record retains source-frozen candidate histories, all original outcomes, physical request usage and the complete experimental program. Main-text and supplementary comparisons include every scheduled task and failed execution.

\end{document}